\pdfoutput=1
\newif\ifreview 
\newif\ifarxiv 
\newif\ifcamera \newcommand{\cameraready}{\cameratrue}
\newif\ifrebuttal 
\cameraready %

\documentclass[10pt,twocolumn,letterpaper]{article}
\def\confName{CVPR}
\def\confYear{2027}
\def\paperTitle{ANON: Asymmetric and Narrated Outward-expansion Network for Photo Composition Repair}
\newcommand{\anon}{ANON}
\AtBeginDocument{\setlength{\abovedisplayskip}{4pt plus 2pt minus 2pt}\setlength{\belowdisplayskip}{4pt plus 2pt minus 2pt}\setlength{\abovedisplayshortskip}{2pt plus 1pt}\setlength{\belowdisplayshortskip}{2pt plus 1pt}}
\usepackage[T1,OT1]{fontenc}
\usepackage{listings}
\newcommand{\anonnum}{\anon$_{\mathrm{num}}$} %
\newcommand{\anonlong}{Asymmetric and Narrated Outward-expansion Network}
\def\authorBlock{Kuan-Lin Chen$^{1}$ \quad Yuan-Kang Lee$^{2}$ \quad Chia-Che Chang$^{2}$ \quad Chih-Heng Chang$^{1}$ \quad Yu-Lun Liu$^{3}$\\ $^{1}$National Taiwan University \quad $^{2}$MediaTek Inc. \quad $^{3}$National Yang Ming Chiao Tung University}

\ifreview \usepackage[review]{cvpr} \fi
\ifarxiv \usepackage[pagenumbers]{cvpr} \fi
\ifrebuttal \usepackage[rebuttal]{cvpr} \fi
\ifcamera \usepackage[pagenumbers]{cvpr} \fi

\usepackage{graphicx}
\usepackage{amsmath,amssymb}
\usepackage{booktabs}
\usepackage{times}
\usepackage{microtype}
\usepackage{caption}
\usepackage{subcaption}
\usepackage{float}
\usepackage{placeins}
\usepackage{xcolor,colortbl}
\usepackage{multirow}
\usepackage{tabularx}
\newcommand{\papertablestyle}{\centering\footnotesize\setlength{\tabcolsep}{3pt}}
\newcommand{\fitpapertable}[1]{\resizebox{\ifdim\width>\linewidth\linewidth\else\width\fi}{!}{#1}}
\usepackage{enumitem}
\usepackage{xspace}
\usepackage{xstring}
\usepackage{url}
\usepackage{pgfkeys}          %
\usepackage[hang,flushmargin]{footmisc}

\newcommand{\vlm}{Qwen3.5-4B\xspace}
\newcommand{\miou}{mIoU\xspace}

\pgfkeys{
  /num/InternVL3-2B-hf_icl9@test300_seed0/miou/.initial=0.610,
  /num/InternVL3-2B-hf_icl9@test300_seed0/lo/.initial=0.599,
  /num/InternVL3-2B-hf_icl9@test300_seed0/hi/.initial=0.621,
  /num/InternVL3-2B-hf_icl9@test300_seed0/dir/.initial=0.237,
  /num/InternVL3-2B-hf_icl9@test300_seed0/parse/.initial=1.000,
  /num/InternVL3-2B-hf_icl9@test300_seed0/n/.initial=300,
  /num/InternVL3-2B-hf_icl9@test300_seed0/mae/.initial=15.07,
  /num/InternVL3-2B-hf_icl9@test300_seed0/f1/.initial=0.003,
  /num/InternVL3-2B-hf_icl9@test300_seed0/f1_5/.initial=0.050,
  /num/InternVL3-2B-hf_icl9@test300_seed0/f1_15/.initial=0.004,
  /num/InternVL3-2B-hf_icl9@test300_seed0/f1_20/.initial=0.005,
  /num/InternVL3-2B-hf_icl9@test300_seed0/p10/.initial=0.500,
  /num/InternVL3-2B-hf_icl9@test300_seed0/r10/.initial=0.002,
  /num/InternVL3-2B-hf_icl9@test_full/miou/.initial=0.606,
  /num/InternVL3-2B-hf_icl9@test_full/lo/.initial=0.603,
  /num/InternVL3-2B-hf_icl9@test_full/hi/.initial=0.609,
  /num/InternVL3-2B-hf_icl9@test_full/dir/.initial=0.255,
  /num/InternVL3-2B-hf_icl9@test_full/parse/.initial=1.000,
  /num/InternVL3-2B-hf_icl9@test_full/n/.initial=3807,
  /num/InternVL3-2B-hf_zs@test300_seed0/miou/.initial=0.613,
  /num/InternVL3-2B-hf_zs@test300_seed0/lo/.initial=0.602,
  /num/InternVL3-2B-hf_zs@test300_seed0/hi/.initial=0.624,
  /num/InternVL3-2B-hf_zs@test300_seed0/dir/.initial=0.237,
  /num/InternVL3-2B-hf_zs@test300_seed0/parse/.initial=1.000,
  /num/InternVL3-2B-hf_zs@test300_seed0/n/.initial=300,
  /num/InternVL3-2B-hf_zs@test300_seed0/mae/.initial=15.22,
  /num/InternVL3-2B-hf_zs@test300_seed0/f1/.initial=0.293,
  /num/InternVL3-2B-hf_zs@test300_seed0/f1_5/.initial=0.324,
  /num/InternVL3-2B-hf_zs@test300_seed0/f1_15/.initial=0.266,
  /num/InternVL3-2B-hf_zs@test300_seed0/f1_20/.initial=0.010,
  /num/InternVL3-2B-hf_zs@test300_seed0/p10/.initial=0.540,
  /num/InternVL3-2B-hf_zs@test300_seed0/r10/.initial=0.201,
  /num/InternVL3-2B-hf_zs@test_full/miou/.initial=0.611,
  /num/InternVL3-2B-hf_zs@test_full/lo/.initial=0.608,
  /num/InternVL3-2B-hf_zs@test_full/hi/.initial=0.614,
  /num/InternVL3-2B-hf_zs@test_full/dir/.initial=0.256,
  /num/InternVL3-2B-hf_zs@test_full/parse/.initial=1.000,
  /num/InternVL3-2B-hf_zs@test_full/n/.initial=3807,
  /num/abl_en_15k_s8090@test300_seed0/miou/.initial=0.788,
  /num/abl_en_15k_s8090@test300_seed0/lo/.initial=0.772,
  /num/abl_en_15k_s8090@test300_seed0/hi/.initial=0.805,
  /num/abl_en_15k_s8090@test300_seed0/dir/.initial=0.573,
  /num/abl_en_15k_s8090@test300_seed0/parse/.initial=1.000,
  /num/abl_en_15k_s8090@test300_seed0/n/.initial=300,
  /num/abl_en_15k_s8090@test300_seed0/mae/.initial=8.73,
  /num/abl_en_15k_s8090@test300_seed0/f1/.initial=0.746,
  /num/abl_en_15k_s8090@test300_seed0/f1_5/.initial=0.759,
  /num/abl_en_15k_s8090@test300_seed0/f1_15/.initial=0.714,
  /num/abl_en_15k_s8090@test300_seed0/f1_20/.initial=0.706,
  /num/abl_en_15k_s8090@test300_seed0/p10/.initial=0.752,
  /num/abl_en_15k_s8090@test300_seed0/r10/.initial=0.740,
  /num/abl_en_15k_s8090@ep1@val300_seed0/miou/.initial=0.754,
  /num/abl_en_15k_s8090@ep1@val300_seed0/lo/.initial=0.738,
  /num/abl_en_15k_s8090@ep1@val300_seed0/hi/.initial=0.771,
  /num/abl_en_15k_s8090@ep1@val300_seed0/dir/.initial=0.520,
  /num/abl_en_15k_s8090@ep1@val300_seed0/parse/.initial=1.000,
  /num/abl_en_15k_s8090@ep1@val300_seed0/n/.initial=300,
  /num/abl_en_15k_s8090@ep2@val300_seed0/miou/.initial=0.781,
  /num/abl_en_15k_s8090@ep2@val300_seed0/lo/.initial=0.765,
  /num/abl_en_15k_s8090@ep2@val300_seed0/hi/.initial=0.797,
  /num/abl_en_15k_s8090@ep2@val300_seed0/dir/.initial=0.593,
  /num/abl_en_15k_s8090@ep2@val300_seed0/parse/.initial=1.000,
  /num/abl_en_15k_s8090@ep2@val300_seed0/n/.initial=300,
  /num/abl_en_15k_s8090@ep3@val300_seed0/miou/.initial=0.778,
  /num/abl_en_15k_s8090@ep3@val300_seed0/lo/.initial=0.761,
  /num/abl_en_15k_s8090@ep3@val300_seed0/hi/.initial=0.793,
  /num/abl_en_15k_s8090@ep3@val300_seed0/dir/.initial=0.573,
  /num/abl_en_15k_s8090@ep3@val300_seed0/parse/.initial=1.000,
  /num/abl_en_15k_s8090@ep3@val300_seed0/n/.initial=300,
  /num/abl_en_15k_s8090@ep4@val300_seed0/miou/.initial=0.793,
  /num/abl_en_15k_s8090@ep4@val300_seed0/lo/.initial=0.777,
  /num/abl_en_15k_s8090@ep4@val300_seed0/hi/.initial=0.808,
  /num/abl_en_15k_s8090@ep4@val300_seed0/dir/.initial=0.613,
  /num/abl_en_15k_s8090@ep4@val300_seed0/parse/.initial=1.000,
  /num/abl_en_15k_s8090@ep4@val300_seed0/n/.initial=300,
  /num/abl_en_15k_s8090@ep5@val300_seed0/miou/.initial=0.794,
  /num/abl_en_15k_s8090@ep5@val300_seed0/lo/.initial=0.779,
  /num/abl_en_15k_s8090@ep5@val300_seed0/hi/.initial=0.810,
  /num/abl_en_15k_s8090@ep5@val300_seed0/dir/.initial=0.610,
  /num/abl_en_15k_s8090@ep5@val300_seed0/parse/.initial=1.000,
  /num/abl_en_15k_s8090@ep5@val300_seed0/n/.initial=300,
  /num/abl_en_5k_s8090@test300_seed0/miou/.initial=0.786,
  /num/abl_en_5k_s8090@test300_seed0/lo/.initial=0.770,
  /num/abl_en_5k_s8090@test300_seed0/hi/.initial=0.802,
  /num/abl_en_5k_s8090@test300_seed0/dir/.initial=0.553,
  /num/abl_en_5k_s8090@test300_seed0/parse/.initial=1.000,
  /num/abl_en_5k_s8090@test300_seed0/n/.initial=300,
  /num/abl_en_5k_s8090@test300_seed0/mae/.initial=8.78,
  /num/abl_en_5k_s8090@test300_seed0/f1/.initial=0.731,
  /num/abl_en_5k_s8090@test300_seed0/f1_5/.initial=0.758,
  /num/abl_en_5k_s8090@test300_seed0/f1_15/.initial=0.704,
  /num/abl_en_5k_s8090@test300_seed0/f1_20/.initial=0.696,
  /num/abl_en_5k_s8090@test300_seed0/p10/.initial=0.736,
  /num/abl_en_5k_s8090@test300_seed0/r10/.initial=0.727,
  /num/abl_en_5k_s8090@ep1@val300_seed0/miou/.initial=0.757,
  /num/abl_en_5k_s8090@ep1@val300_seed0/lo/.initial=0.741,
  /num/abl_en_5k_s8090@ep1@val300_seed0/hi/.initial=0.775,
  /num/abl_en_5k_s8090@ep1@val300_seed0/dir/.initial=0.567,
  /num/abl_en_5k_s8090@ep1@val300_seed0/parse/.initial=1.000,
  /num/abl_en_5k_s8090@ep1@val300_seed0/n/.initial=300,
  /num/abl_en_5k_s8090@ep2@val300_seed0/miou/.initial=0.776,
  /num/abl_en_5k_s8090@ep2@val300_seed0/lo/.initial=0.759,
  /num/abl_en_5k_s8090@ep2@val300_seed0/hi/.initial=0.791,
  /num/abl_en_5k_s8090@ep2@val300_seed0/dir/.initial=0.583,
  /num/abl_en_5k_s8090@ep2@val300_seed0/parse/.initial=1.000,
  /num/abl_en_5k_s8090@ep2@val300_seed0/n/.initial=300,
  /num/abl_en_5k_s8090@ep3@val300_seed0/miou/.initial=0.785,
  /num/abl_en_5k_s8090@ep3@val300_seed0/lo/.initial=0.771,
  /num/abl_en_5k_s8090@ep3@val300_seed0/hi/.initial=0.800,
  /num/abl_en_5k_s8090@ep3@val300_seed0/dir/.initial=0.573,
  /num/abl_en_5k_s8090@ep3@val300_seed0/parse/.initial=1.000,
  /num/abl_en_5k_s8090@ep3@val300_seed0/n/.initial=300,
  /num/abl_en_5k_s8090@ep4@val300_seed0/miou/.initial=0.786,
  /num/abl_en_5k_s8090@ep4@val300_seed0/lo/.initial=0.770,
  /num/abl_en_5k_s8090@ep4@val300_seed0/hi/.initial=0.801,
  /num/abl_en_5k_s8090@ep4@val300_seed0/dir/.initial=0.527,
  /num/abl_en_5k_s8090@ep4@val300_seed0/parse/.initial=1.000,
  /num/abl_en_5k_s8090@ep4@val300_seed0/n/.initial=300,
  /num/abl_en_5k_s8090@ep5@val300_seed0/miou/.initial=0.782,
  /num/abl_en_5k_s8090@ep5@val300_seed0/lo/.initial=0.765,
  /num/abl_en_5k_s8090@ep5@val300_seed0/hi/.initial=0.797,
  /num/abl_en_5k_s8090@ep5@val300_seed0/dir/.initial=0.533,
  /num/abl_en_5k_s8090@ep5@val300_seed0/parse/.initial=1.000,
  /num/abl_en_5k_s8090@ep5@val300_seed0/n/.initial=300,
  /num/abl_en_r1to0@test300_seed0/miou/.initial=0.740,
  /num/abl_en_r1to0@test300_seed0/lo/.initial=0.724,
  /num/abl_en_r1to0@test300_seed0/hi/.initial=0.758,
  /num/abl_en_r1to0@test300_seed0/dir/.initial=0.517,
  /num/abl_en_r1to0@test300_seed0/parse/.initial=1.000,
  /num/abl_en_r1to0@test300_seed0/n/.initial=300,
  /num/abl_en_r1to0@test300_seed0/mae/.initial=11.03,
  /num/abl_en_r1to0@test300_seed0/f1/.initial=0.706,
  /num/abl_en_r1to0@test300_seed0/f1_5/.initial=0.725,
  /num/abl_en_r1to0@test300_seed0/f1_15/.initial=0.656,
  /num/abl_en_r1to0@test300_seed0/f1_20/.initial=0.650,
  /num/abl_en_r1to0@test300_seed0/p10/.initial=0.716,
  /num/abl_en_r1to0@test300_seed0/r10/.initial=0.697,
  /num/abl_en_r1to0@test_full/miou/.initial=0.745,
  /num/abl_en_r1to0@test_full/lo/.initial=0.740,
  /num/abl_en_r1to0@test_full/hi/.initial=0.749,
  /num/abl_en_r1to0@test_full/dir/.initial=0.536,
  /num/abl_en_r1to0@test_full/parse/.initial=1.000,
  /num/abl_en_r1to0@test_full/n/.initial=3807,
  /num/abl_en_r1to0@ep1@val300_seed0/miou/.initial=0.697,
  /num/abl_en_r1to0@ep1@val300_seed0/lo/.initial=0.679,
  /num/abl_en_r1to0@ep1@val300_seed0/hi/.initial=0.716,
  /num/abl_en_r1to0@ep1@val300_seed0/dir/.initial=0.507,
  /num/abl_en_r1to0@ep1@val300_seed0/parse/.initial=1.000,
  /num/abl_en_r1to0@ep1@val300_seed0/n/.initial=300,
  /num/abl_en_r1to0@ep2@val300_seed0/miou/.initial=0.732,
  /num/abl_en_r1to0@ep2@val300_seed0/lo/.initial=0.714,
  /num/abl_en_r1to0@ep2@val300_seed0/hi/.initial=0.749,
  /num/abl_en_r1to0@ep2@val300_seed0/dir/.initial=0.547,
  /num/abl_en_r1to0@ep2@val300_seed0/parse/.initial=1.000,
  /num/abl_en_r1to0@ep2@val300_seed0/n/.initial=300,
  /num/abl_en_r1to0@ep3@val300_seed0/miou/.initial=0.746,
  /num/abl_en_r1to0@ep3@val300_seed0/lo/.initial=0.728,
  /num/abl_en_r1to0@ep3@val300_seed0/hi/.initial=0.763,
  /num/abl_en_r1to0@ep3@val300_seed0/dir/.initial=0.517,
  /num/abl_en_r1to0@ep3@val300_seed0/parse/.initial=1.000,
  /num/abl_en_r1to0@ep3@val300_seed0/n/.initial=300,
  /num/abl_en_r1to0@ep4@val300_seed0/miou/.initial=0.754,
  /num/abl_en_r1to0@ep4@val300_seed0/lo/.initial=0.737,
  /num/abl_en_r1to0@ep4@val300_seed0/hi/.initial=0.771,
  /num/abl_en_r1to0@ep4@val300_seed0/dir/.initial=0.543,
  /num/abl_en_r1to0@ep4@val300_seed0/parse/.initial=1.000,
  /num/abl_en_r1to0@ep4@val300_seed0/n/.initial=300,
  /num/abl_en_r1to0@ep5@val300_seed0/miou/.initial=0.754,
  /num/abl_en_r1to0@ep5@val300_seed0/lo/.initial=0.737,
  /num/abl_en_r1to0@ep5@val300_seed0/hi/.initial=0.769,
  /num/abl_en_r1to0@ep5@val300_seed0/dir/.initial=0.540,
  /num/abl_en_r1to0@ep5@val300_seed0/parse/.initial=1.000,
  /num/abl_en_r1to0@ep5@val300_seed0/n/.initial=300,
  /num/abl_en_r1to0_s8090@test300_seed0/miou/.initial=0.745,
  /num/abl_en_r1to0_s8090@test300_seed0/lo/.initial=0.729,
  /num/abl_en_r1to0_s8090@test300_seed0/hi/.initial=0.761,
  /num/abl_en_r1to0_s8090@test300_seed0/dir/.initial=0.510,
  /num/abl_en_r1to0_s8090@test300_seed0/parse/.initial=1.000,
  /num/abl_en_r1to0_s8090@test300_seed0/n/.initial=300,
  /num/abl_en_r1to0_s8090@test300_seed0/mae/.initial=10.75,
  /num/abl_en_r1to0_s8090@test300_seed0/f1/.initial=0.688,
  /num/abl_en_r1to0_s8090@test300_seed0/f1_5/.initial=0.729,
  /num/abl_en_r1to0_s8090@test300_seed0/f1_15/.initial=0.648,
  /num/abl_en_r1to0_s8090@test300_seed0/f1_20/.initial=0.630,
  /num/abl_en_r1to0_s8090@test300_seed0/p10/.initial=0.683,
  /num/abl_en_r1to0_s8090@test300_seed0/r10/.initial=0.694,
  /num/abl_en_r1to0_s8090@test_full/miou/.initial=0.741,
  /num/abl_en_r1to0_s8090@test_full/lo/.initial=0.737,
  /num/abl_en_r1to0_s8090@test_full/hi/.initial=0.746,
  /num/abl_en_r1to0_s8090@test_full/dir/.initial=0.498,
  /num/abl_en_r1to0_s8090@test_full/parse/.initial=0.999,
  /num/abl_en_r1to0_s8090@test_full/n/.initial=3807,
  /num/abl_en_r1to0_s8090@ep1@val300_seed0/miou/.initial=0.752,
  /num/abl_en_r1to0_s8090@ep1@val300_seed0/lo/.initial=0.734,
  /num/abl_en_r1to0_s8090@ep1@val300_seed0/hi/.initial=0.769,
  /num/abl_en_r1to0_s8090@ep1@val300_seed0/dir/.initial=0.533,
  /num/abl_en_r1to0_s8090@ep1@val300_seed0/parse/.initial=1.000,
  /num/abl_en_r1to0_s8090@ep1@val300_seed0/n/.initial=300,
  /num/abl_en_r1to0_s8090@ep2@val300_seed0/miou/.initial=0.757,
  /num/abl_en_r1to0_s8090@ep2@val300_seed0/lo/.initial=0.741,
  /num/abl_en_r1to0_s8090@ep2@val300_seed0/hi/.initial=0.772,
  /num/abl_en_r1to0_s8090@ep2@val300_seed0/dir/.initial=0.540,
  /num/abl_en_r1to0_s8090@ep2@val300_seed0/parse/.initial=1.000,
  /num/abl_en_r1to0_s8090@ep2@val300_seed0/n/.initial=300,
  /num/abl_en_r1to0_s8090@ep3@val300_seed0/miou/.initial=0.758,
  /num/abl_en_r1to0_s8090@ep3@val300_seed0/lo/.initial=0.742,
  /num/abl_en_r1to0_s8090@ep3@val300_seed0/hi/.initial=0.775,
  /num/abl_en_r1to0_s8090@ep3@val300_seed0/dir/.initial=0.530,
  /num/abl_en_r1to0_s8090@ep3@val300_seed0/parse/.initial=1.000,
  /num/abl_en_r1to0_s8090@ep3@val300_seed0/n/.initial=300,
  /num/abl_en_r1to0_s8090@ep4@val300_seed0/miou/.initial=0.756,
  /num/abl_en_r1to0_s8090@ep4@val300_seed0/lo/.initial=0.740,
  /num/abl_en_r1to0_s8090@ep4@val300_seed0/hi/.initial=0.772,
  /num/abl_en_r1to0_s8090@ep4@val300_seed0/dir/.initial=0.577,
  /num/abl_en_r1to0_s8090@ep4@val300_seed0/parse/.initial=1.000,
  /num/abl_en_r1to0_s8090@ep4@val300_seed0/n/.initial=300,
  /num/abl_en_r1to0_s8090@ep5@val300_seed0/miou/.initial=0.755,
  /num/abl_en_r1to0_s8090@ep5@val300_seed0/lo/.initial=0.738,
  /num/abl_en_r1to0_s8090@ep5@val300_seed0/hi/.initial=0.772,
  /num/abl_en_r1to0_s8090@ep5@val300_seed0/dir/.initial=0.540,
  /num/abl_en_r1to0_s8090@ep5@val300_seed0/parse/.initial=1.000,
  /num/abl_en_r1to0_s8090@ep5@val300_seed0/n/.initial=300,
  /num/abl_en_r1to1_s8090@test300_seed0/miou/.initial=0.762,
  /num/abl_en_r1to1_s8090@test300_seed0/lo/.initial=0.746,
  /num/abl_en_r1to1_s8090@test300_seed0/hi/.initial=0.779,
  /num/abl_en_r1to1_s8090@test300_seed0/dir/.initial=0.507,
  /num/abl_en_r1to1_s8090@test300_seed0/parse/.initial=1.000,
  /num/abl_en_r1to1_s8090@test300_seed0/n/.initial=300,
  /num/abl_en_r1to1_s8090@test300_seed0/mae/.initial=9.92,
  /num/abl_en_r1to1_s8090@test300_seed0/f1/.initial=0.705,
  /num/abl_en_r1to1_s8090@test300_seed0/f1_5/.initial=0.745,
  /num/abl_en_r1to1_s8090@test300_seed0/f1_15/.initial=0.669,
  /num/abl_en_r1to1_s8090@test300_seed0/f1_20/.initial=0.658,
  /num/abl_en_r1to1_s8090@test300_seed0/p10/.initial=0.703,
  /num/abl_en_r1to1_s8090@test300_seed0/r10/.initial=0.708,
  /num/abl_en_r1to1_s8090@test_full/miou/.initial=0.760,
  /num/abl_en_r1to1_s8090@test_full/lo/.initial=0.755,
  /num/abl_en_r1to1_s8090@test_full/hi/.initial=0.765,
  /num/abl_en_r1to1_s8090@test_full/dir/.initial=0.525,
  /num/abl_en_r1to1_s8090@test_full/parse/.initial=1.000,
  /num/abl_en_r1to1_s8090@test_full/n/.initial=3807,
  /num/abl_en_r1to1_s8090@ep1@val300_seed0/miou/.initial=0.762,
  /num/abl_en_r1to1_s8090@ep1@val300_seed0/lo/.initial=0.746,
  /num/abl_en_r1to1_s8090@ep1@val300_seed0/hi/.initial=0.778,
  /num/abl_en_r1to1_s8090@ep1@val300_seed0/dir/.initial=0.550,
  /num/abl_en_r1to1_s8090@ep1@val300_seed0/parse/.initial=1.000,
  /num/abl_en_r1to1_s8090@ep1@val300_seed0/n/.initial=300,
  /num/abl_en_r1to1_s8090@ep2@val300_seed0/miou/.initial=0.777,
  /num/abl_en_r1to1_s8090@ep2@val300_seed0/lo/.initial=0.762,
  /num/abl_en_r1to1_s8090@ep2@val300_seed0/hi/.initial=0.793,
  /num/abl_en_r1to1_s8090@ep2@val300_seed0/dir/.initial=0.580,
  /num/abl_en_r1to1_s8090@ep2@val300_seed0/parse/.initial=1.000,
  /num/abl_en_r1to1_s8090@ep2@val300_seed0/n/.initial=300,
  /num/abl_en_r1to1_s8090@ep3@val300_seed0/miou/.initial=0.774,
  /num/abl_en_r1to1_s8090@ep3@val300_seed0/lo/.initial=0.758,
  /num/abl_en_r1to1_s8090@ep3@val300_seed0/hi/.initial=0.789,
  /num/abl_en_r1to1_s8090@ep3@val300_seed0/dir/.initial=0.560,
  /num/abl_en_r1to1_s8090@ep3@val300_seed0/parse/.initial=1.000,
  /num/abl_en_r1to1_s8090@ep3@val300_seed0/n/.initial=300,
  /num/abl_en_r1to1_s8090@ep4@val300_seed0/miou/.initial=0.775,
  /num/abl_en_r1to1_s8090@ep4@val300_seed0/lo/.initial=0.760,
  /num/abl_en_r1to1_s8090@ep4@val300_seed0/hi/.initial=0.790,
  /num/abl_en_r1to1_s8090@ep4@val300_seed0/dir/.initial=0.563,
  /num/abl_en_r1to1_s8090@ep4@val300_seed0/parse/.initial=1.000,
  /num/abl_en_r1to1_s8090@ep4@val300_seed0/n/.initial=300,
  /num/abl_en_r1to1_s8090@ep5@val300_seed0/miou/.initial=0.773,
  /num/abl_en_r1to1_s8090@ep5@val300_seed0/lo/.initial=0.759,
  /num/abl_en_r1to1_s8090@ep5@val300_seed0/hi/.initial=0.788,
  /num/abl_en_r1to1_s8090@ep5@val300_seed0/dir/.initial=0.557,
  /num/abl_en_r1to1_s8090@ep5@val300_seed0/parse/.initial=1.000,
  /num/abl_en_r1to1_s8090@ep5@val300_seed0/n/.initial=300,
  /num/abl_en_r1to2_s8090@test300_seed0/miou/.initial=0.773,
  /num/abl_en_r1to2_s8090@test300_seed0/lo/.initial=0.758,
  /num/abl_en_r1to2_s8090@test300_seed0/hi/.initial=0.788,
  /num/abl_en_r1to2_s8090@test300_seed0/dir/.initial=0.563,
  /num/abl_en_r1to2_s8090@test300_seed0/parse/.initial=1.000,
  /num/abl_en_r1to2_s8090@test300_seed0/n/.initial=300,
  /num/abl_en_r1to2_s8090@test300_seed0/mae/.initial=9.36,
  /num/abl_en_r1to2_s8090@test300_seed0/f1/.initial=0.725,
  /num/abl_en_r1to2_s8090@test300_seed0/f1_5/.initial=0.776,
  /num/abl_en_r1to2_s8090@test300_seed0/f1_15/.initial=0.684,
  /num/abl_en_r1to2_s8090@test300_seed0/f1_20/.initial=0.672,
  /num/abl_en_r1to2_s8090@test300_seed0/p10/.initial=0.697,
  /num/abl_en_r1to2_s8090@test300_seed0/r10/.initial=0.756,
  /num/abl_en_r1to2_s8090@ep1@val300_seed0/miou/.initial=0.750,
  /num/abl_en_r1to2_s8090@ep1@val300_seed0/lo/.initial=0.732,
  /num/abl_en_r1to2_s8090@ep1@val300_seed0/hi/.initial=0.768,
  /num/abl_en_r1to2_s8090@ep1@val300_seed0/dir/.initial=0.513,
  /num/abl_en_r1to2_s8090@ep1@val300_seed0/parse/.initial=1.000,
  /num/abl_en_r1to2_s8090@ep1@val300_seed0/n/.initial=300,
  /num/abl_en_r1to2_s8090@ep2@val300_seed0/miou/.initial=0.762,
  /num/abl_en_r1to2_s8090@ep2@val300_seed0/lo/.initial=0.746,
  /num/abl_en_r1to2_s8090@ep2@val300_seed0/hi/.initial=0.778,
  /num/abl_en_r1to2_s8090@ep2@val300_seed0/dir/.initial=0.547,
  /num/abl_en_r1to2_s8090@ep2@val300_seed0/parse/.initial=1.000,
  /num/abl_en_r1to2_s8090@ep2@val300_seed0/n/.initial=300,
  /num/abl_en_r1to2_s8090@ep3@val300_seed0/miou/.initial=0.773,
  /num/abl_en_r1to2_s8090@ep3@val300_seed0/lo/.initial=0.759,
  /num/abl_en_r1to2_s8090@ep3@val300_seed0/hi/.initial=0.788,
  /num/abl_en_r1to2_s8090@ep3@val300_seed0/dir/.initial=0.520,
  /num/abl_en_r1to2_s8090@ep3@val300_seed0/parse/.initial=1.000,
  /num/abl_en_r1to2_s8090@ep3@val300_seed0/n/.initial=300,
  /num/abl_en_r1to2_s8090@ep4@val300_seed0/miou/.initial=0.782,
  /num/abl_en_r1to2_s8090@ep4@val300_seed0/lo/.initial=0.767,
  /num/abl_en_r1to2_s8090@ep4@val300_seed0/hi/.initial=0.796,
  /num/abl_en_r1to2_s8090@ep4@val300_seed0/dir/.initial=0.553,
  /num/abl_en_r1to2_s8090@ep4@val300_seed0/parse/.initial=1.000,
  /num/abl_en_r1to2_s8090@ep4@val300_seed0/n/.initial=300,
  /num/abl_en_r1to2_s8090@ep5@val300_seed0/miou/.initial=0.781,
  /num/abl_en_r1to2_s8090@ep5@val300_seed0/lo/.initial=0.766,
  /num/abl_en_r1to2_s8090@ep5@val300_seed0/hi/.initial=0.796,
  /num/abl_en_r1to2_s8090@ep5@val300_seed0/dir/.initial=0.560,
  /num/abl_en_r1to2_s8090@ep5@val300_seed0/parse/.initial=1.000,
  /num/abl_en_r1to2_s8090@ep5@val300_seed0/n/.initial=300,
  /num/abl_en_scale10k_s8090@test300_seed0/miou/.initial=0.791,
  /num/abl_en_scale10k_s8090@test300_seed0/lo/.initial=0.774,
  /num/abl_en_scale10k_s8090@test300_seed0/hi/.initial=0.809,
  /num/abl_en_scale10k_s8090@test300_seed0/dir/.initial=0.580,
  /num/abl_en_scale10k_s8090@test300_seed0/parse/.initial=1.000,
  /num/abl_en_scale10k_s8090@test300_seed0/n/.initial=300,
  /num/abl_en_scale10k_s8090@test300_seed0/mae/.initial=8.70,
  /num/abl_en_scale10k_s8090@test300_seed0/f1/.initial=0.754,
  /num/abl_en_scale10k_s8090@test300_seed0/f1_5/.initial=0.768,
  /num/abl_en_scale10k_s8090@test300_seed0/f1_15/.initial=0.718,
  /num/abl_en_scale10k_s8090@test300_seed0/f1_20/.initial=0.724,
  /num/abl_en_scale10k_s8090@test300_seed0/p10/.initial=0.769,
  /num/abl_en_scale10k_s8090@test300_seed0/r10/.initial=0.740,
  /num/abl_en_scale10k_s8090@ep1@val300_seed0/miou/.initial=0.743,
  /num/abl_en_scale10k_s8090@ep1@val300_seed0/lo/.initial=0.725,
  /num/abl_en_scale10k_s8090@ep1@val300_seed0/hi/.initial=0.761,
  /num/abl_en_scale10k_s8090@ep1@val300_seed0/dir/.initial=0.533,
  /num/abl_en_scale10k_s8090@ep1@val300_seed0/parse/.initial=1.000,
  /num/abl_en_scale10k_s8090@ep1@val300_seed0/n/.initial=300,
  /num/abl_en_scale10k_s8090@ep2@val300_seed0/miou/.initial=0.786,
  /num/abl_en_scale10k_s8090@ep2@val300_seed0/lo/.initial=0.770,
  /num/abl_en_scale10k_s8090@ep2@val300_seed0/hi/.initial=0.803,
  /num/abl_en_scale10k_s8090@ep2@val300_seed0/dir/.initial=0.597,
  /num/abl_en_scale10k_s8090@ep2@val300_seed0/parse/.initial=1.000,
  /num/abl_en_scale10k_s8090@ep2@val300_seed0/n/.initial=300,
  /num/abl_en_scale10k_s8090@ep3@val300_seed0/miou/.initial=0.787,
  /num/abl_en_scale10k_s8090@ep3@val300_seed0/lo/.initial=0.770,
  /num/abl_en_scale10k_s8090@ep3@val300_seed0/hi/.initial=0.802,
  /num/abl_en_scale10k_s8090@ep3@val300_seed0/dir/.initial=0.587,
  /num/abl_en_scale10k_s8090@ep3@val300_seed0/parse/.initial=1.000,
  /num/abl_en_scale10k_s8090@ep3@val300_seed0/n/.initial=300,
  /num/abl_en_scale10k_s8090@ep4@val300_seed0/miou/.initial=0.795,
  /num/abl_en_scale10k_s8090@ep4@val300_seed0/lo/.initial=0.778,
  /num/abl_en_scale10k_s8090@ep4@val300_seed0/hi/.initial=0.811,
  /num/abl_en_scale10k_s8090@ep4@val300_seed0/dir/.initial=0.587,
  /num/abl_en_scale10k_s8090@ep4@val300_seed0/parse/.initial=1.000,
  /num/abl_en_scale10k_s8090@ep4@val300_seed0/n/.initial=300,
  /num/abl_en_scale10k_s8090@ep5@val300_seed0/miou/.initial=0.795,
  /num/abl_en_scale10k_s8090@ep5@val300_seed0/lo/.initial=0.778,
  /num/abl_en_scale10k_s8090@ep5@val300_seed0/hi/.initial=0.811,
  /num/abl_en_scale10k_s8090@ep5@val300_seed0/dir/.initial=0.593,
  /num/abl_en_scale10k_s8090@ep5@val300_seed0/parse/.initial=1.000,
  /num/abl_en_scale10k_s8090@ep5@val300_seed0/n/.initial=300,
  /num/abl_en_scalefull_s8090@test300_seed0/miou/.initial=0.790,
  /num/abl_en_scalefull_s8090@test300_seed0/lo/.initial=0.771,
  /num/abl_en_scalefull_s8090@test300_seed0/hi/.initial=0.807,
  /num/abl_en_scalefull_s8090@test300_seed0/dir/.initial=0.590,
  /num/abl_en_scalefull_s8090@test300_seed0/parse/.initial=1.000,
  /num/abl_en_scalefull_s8090@test300_seed0/n/.initial=300,
  /num/abl_en_scalefull_s8090@test300_seed0/mae/.initial=8.80,
  /num/abl_en_scalefull_s8090@test300_seed0/f1/.initial=0.751,
  /num/abl_en_scalefull_s8090@test300_seed0/f1_5/.initial=0.760,
  /num/abl_en_scalefull_s8090@test300_seed0/f1_15/.initial=0.706,
  /num/abl_en_scalefull_s8090@test300_seed0/f1_20/.initial=0.700,
  /num/abl_en_scalefull_s8090@test300_seed0/p10/.initial=0.765,
  /num/abl_en_scalefull_s8090@test300_seed0/r10/.initial=0.738,
  /num/abl_en_scalefull_s8090@ep1@val300_seed0/miou/.initial=0.766,
  /num/abl_en_scalefull_s8090@ep1@val300_seed0/lo/.initial=0.749,
  /num/abl_en_scalefull_s8090@ep1@val300_seed0/hi/.initial=0.783,
  /num/abl_en_scalefull_s8090@ep1@val300_seed0/dir/.initial=0.573,
  /num/abl_en_scalefull_s8090@ep1@val300_seed0/parse/.initial=1.000,
  /num/abl_en_scalefull_s8090@ep1@val300_seed0/n/.initial=300,
  /num/abl_en_scalefull_s8090@ep2@val300_seed0/miou/.initial=0.787,
  /num/abl_en_scalefull_s8090@ep2@val300_seed0/lo/.initial=0.772,
  /num/abl_en_scalefull_s8090@ep2@val300_seed0/hi/.initial=0.802,
  /num/abl_en_scalefull_s8090@ep2@val300_seed0/dir/.initial=0.603,
  /num/abl_en_scalefull_s8090@ep2@val300_seed0/parse/.initial=1.000,
  /num/abl_en_scalefull_s8090@ep2@val300_seed0/n/.initial=300,
  /num/abl_en_scalefull_s8090@ep3@val300_seed0/miou/.initial=0.784,
  /num/abl_en_scalefull_s8090@ep3@val300_seed0/lo/.initial=0.767,
  /num/abl_en_scalefull_s8090@ep3@val300_seed0/hi/.initial=0.799,
  /num/abl_en_scalefull_s8090@ep3@val300_seed0/dir/.initial=0.557,
  /num/abl_en_scalefull_s8090@ep3@val300_seed0/parse/.initial=1.000,
  /num/abl_en_scalefull_s8090@ep3@val300_seed0/n/.initial=300,
  /num/abl_en_scalefull_s8090@ep4@val300_seed0/miou/.initial=0.790,
  /num/abl_en_scalefull_s8090@ep4@val300_seed0/lo/.initial=0.774,
  /num/abl_en_scalefull_s8090@ep4@val300_seed0/hi/.initial=0.806,
  /num/abl_en_scalefull_s8090@ep4@val300_seed0/dir/.initial=0.580,
  /num/abl_en_scalefull_s8090@ep4@val300_seed0/parse/.initial=1.000,
  /num/abl_en_scalefull_s8090@ep4@val300_seed0/n/.initial=300,
  /num/abl_en_scalefull_s8090@ep5@val300_seed0/miou/.initial=0.792,
  /num/abl_en_scalefull_s8090@ep5@val300_seed0/lo/.initial=0.776,
  /num/abl_en_scalefull_s8090@ep5@val300_seed0/hi/.initial=0.807,
  /num/abl_en_scalefull_s8090@ep5@val300_seed0/dir/.initial=0.603,
  /num/abl_en_scalefull_s8090@ep5@val300_seed0/parse/.initial=1.000,
  /num/abl_en_scalefull_s8090@ep5@val300_seed0/n/.initial=300,
  /num/abl_final_r1to0@test300_seed0/miou/.initial=0.782,
  /num/abl_final_r1to0@test300_seed0/lo/.initial=0.765,
  /num/abl_final_r1to0@test300_seed0/hi/.initial=0.799,
  /num/abl_final_r1to0@test300_seed0/dir/.initial=0.540,
  /num/abl_final_r1to0@test300_seed0/parse/.initial=1.000,
  /num/abl_final_r1to0@test300_seed0/n/.initial=300,
  /num/abl_final_r1to0@test300_seed0/mae/.initial=9.00,
  /num/abl_final_r1to0@test300_seed0/f1/.initial=0.737,
  /num/abl_final_r1to0@test300_seed0/f1_5/.initial=0.764,
  /num/abl_final_r1to0@test300_seed0/f1_15/.initial=0.701,
  /num/abl_final_r1to0@test300_seed0/f1_20/.initial=0.687,
  /num/abl_final_r1to0@test300_seed0/p10/.initial=0.738,
  /num/abl_final_r1to0@test300_seed0/r10/.initial=0.736,
  /num/abl_final_r1to0@s1875@val300_seed0/miou/.initial=0.767,
  /num/abl_final_r1to0@s1875@val300_seed0/lo/.initial=0.752,
  /num/abl_final_r1to0@s1875@val300_seed0/hi/.initial=0.784,
  /num/abl_final_r1to0@s1875@val300_seed0/dir/.initial=0.580,
  /num/abl_final_r1to0@s1875@val300_seed0/parse/.initial=1.000,
  /num/abl_final_r1to0@s1875@val300_seed0/n/.initial=300,
  /num/abl_final_r1to0@s3750@val300_seed0/miou/.initial=0.766,
  /num/abl_final_r1to0@s3750@val300_seed0/lo/.initial=0.750,
  /num/abl_final_r1to0@s3750@val300_seed0/hi/.initial=0.782,
  /num/abl_final_r1to0@s3750@val300_seed0/dir/.initial=0.547,
  /num/abl_final_r1to0@s3750@val300_seed0/parse/.initial=1.000,
  /num/abl_final_r1to0@s3750@val300_seed0/n/.initial=300,
  /num/abl_final_r1to0@s5625@val300_seed0/miou/.initial=0.780,
  /num/abl_final_r1to0@s5625@val300_seed0/lo/.initial=0.764,
  /num/abl_final_r1to0@s5625@val300_seed0/hi/.initial=0.796,
  /num/abl_final_r1to0@s5625@val300_seed0/dir/.initial=0.593,
  /num/abl_final_r1to0@s5625@val300_seed0/parse/.initial=1.000,
  /num/abl_final_r1to0@s5625@val300_seed0/n/.initial=300,
  /num/abl_final_r1to0@s7500@val300_seed0/miou/.initial=0.781,
  /num/abl_final_r1to0@s7500@val300_seed0/lo/.initial=0.765,
  /num/abl_final_r1to0@s7500@val300_seed0/hi/.initial=0.797,
  /num/abl_final_r1to0@s7500@val300_seed0/dir/.initial=0.567,
  /num/abl_final_r1to0@s7500@val300_seed0/parse/.initial=1.000,
  /num/abl_final_r1to0@s7500@val300_seed0/n/.initial=300,
  /num/abl_final_r1to0@s9375@val300_seed0/miou/.initial=0.779,
  /num/abl_final_r1to0@s9375@val300_seed0/lo/.initial=0.763,
  /num/abl_final_r1to0@s9375@val300_seed0/hi/.initial=0.795,
  /num/abl_final_r1to0@s9375@val300_seed0/dir/.initial=0.530,
  /num/abl_final_r1to0@s9375@val300_seed0/parse/.initial=1.000,
  /num/abl_final_r1to0@s9375@val300_seed0/n/.initial=300,
  /num/abl_final_r1to2@test300_seed0/miou/.initial=0.792,
  /num/abl_final_r1to2@test300_seed0/lo/.initial=0.775,
  /num/abl_final_r1to2@test300_seed0/hi/.initial=0.808,
  /num/abl_final_r1to2@test300_seed0/dir/.initial=0.590,
  /num/abl_final_r1to2@test300_seed0/parse/.initial=1.000,
  /num/abl_final_r1to2@test300_seed0/n/.initial=300,
  /num/abl_final_r1to2@test300_seed0/mae/.initial=8.52,
  /num/abl_final_r1to2@test300_seed0/f1/.initial=0.747,
  /num/abl_final_r1to2@test300_seed0/f1_5/.initial=0.770,
  /num/abl_final_r1to2@test300_seed0/f1_15/.initial=0.715,
  /num/abl_final_r1to2@test300_seed0/f1_20/.initial=0.711,
  /num/abl_final_r1to2@test300_seed0/p10/.initial=0.755,
  /num/abl_final_r1to2@test300_seed0/r10/.initial=0.740,
  /num/abl_final_r1to2@s1875@val300_seed0/miou/.initial=0.759,
  /num/abl_final_r1to2@s1875@val300_seed0/lo/.initial=0.743,
  /num/abl_final_r1to2@s1875@val300_seed0/hi/.initial=0.776,
  /num/abl_final_r1to2@s1875@val300_seed0/dir/.initial=0.547,
  /num/abl_final_r1to2@s1875@val300_seed0/parse/.initial=1.000,
  /num/abl_final_r1to2@s1875@val300_seed0/n/.initial=300,
  /num/abl_final_r1to2@s3750@val300_seed0/miou/.initial=0.785,
  /num/abl_final_r1to2@s3750@val300_seed0/lo/.initial=0.769,
  /num/abl_final_r1to2@s3750@val300_seed0/hi/.initial=0.800,
  /num/abl_final_r1to2@s3750@val300_seed0/dir/.initial=0.577,
  /num/abl_final_r1to2@s3750@val300_seed0/parse/.initial=1.000,
  /num/abl_final_r1to2@s3750@val300_seed0/n/.initial=300,
  /num/abl_final_r1to2@s5625@val300_seed0/miou/.initial=0.795,
  /num/abl_final_r1to2@s5625@val300_seed0/lo/.initial=0.780,
  /num/abl_final_r1to2@s5625@val300_seed0/hi/.initial=0.810,
  /num/abl_final_r1to2@s5625@val300_seed0/dir/.initial=0.617,
  /num/abl_final_r1to2@s5625@val300_seed0/parse/.initial=1.000,
  /num/abl_final_r1to2@s5625@val300_seed0/n/.initial=300,
  /num/abl_final_r1to2@s7500@val300_seed0/miou/.initial=0.798,
  /num/abl_final_r1to2@s7500@val300_seed0/lo/.initial=0.783,
  /num/abl_final_r1to2@s7500@val300_seed0/hi/.initial=0.813,
  /num/abl_final_r1to2@s7500@val300_seed0/dir/.initial=0.620,
  /num/abl_final_r1to2@s7500@val300_seed0/parse/.initial=1.000,
  /num/abl_final_r1to2@s7500@val300_seed0/n/.initial=300,
  /num/abl_final_r1to2@s9375@val300_seed0/miou/.initial=0.796,
  /num/abl_final_r1to2@s9375@val300_seed0/lo/.initial=0.781,
  /num/abl_final_r1to2@s9375@val300_seed0/hi/.initial=0.811,
  /num/abl_final_r1to2@s9375@val300_seed0/dir/.initial=0.597,
  /num/abl_final_r1to2@s9375@val300_seed0/parse/.initial=1.000,
  /num/abl_final_r1to2@s9375@val300_seed0/n/.initial=300,
  /num/abl_final_rs4780@test300_seed0/miou/.initial=0.784,
  /num/abl_final_rs4780@test300_seed0/lo/.initial=0.769,
  /num/abl_final_rs4780@test300_seed0/hi/.initial=0.800,
  /num/abl_final_rs4780@test300_seed0/dir/.initial=0.570,
  /num/abl_final_rs4780@test300_seed0/parse/.initial=1.000,
  /num/abl_final_rs4780@test300_seed0/n/.initial=300,
  /num/abl_final_rs4780@test300_seed0/mae/.initial=8.88,
  /num/abl_final_rs4780@test300_seed0/f1/.initial=0.748,
  /num/abl_final_rs4780@test300_seed0/f1_5/.initial=0.768,
  /num/abl_final_rs4780@test300_seed0/f1_15/.initial=0.689,
  /num/abl_final_rs4780@test300_seed0/f1_20/.initial=0.688,
  /num/abl_final_rs4780@test300_seed0/p10/.initial=0.734,
  /num/abl_final_rs4780@test300_seed0/r10/.initial=0.763,
  /num/abl_final_rs4780@s1875@val300_seed0/miou/.initial=0.771,
  /num/abl_final_rs4780@s1875@val300_seed0/lo/.initial=0.755,
  /num/abl_final_rs4780@s1875@val300_seed0/hi/.initial=0.787,
  /num/abl_final_rs4780@s1875@val300_seed0/dir/.initial=0.570,
  /num/abl_final_rs4780@s1875@val300_seed0/parse/.initial=1.000,
  /num/abl_final_rs4780@s1875@val300_seed0/n/.initial=300,
  /num/abl_final_rs4780@s3750@val300_seed0/miou/.initial=0.772,
  /num/abl_final_rs4780@s3750@val300_seed0/lo/.initial=0.755,
  /num/abl_final_rs4780@s3750@val300_seed0/hi/.initial=0.788,
  /num/abl_final_rs4780@s3750@val300_seed0/dir/.initial=0.533,
  /num/abl_final_rs4780@s3750@val300_seed0/parse/.initial=1.000,
  /num/abl_final_rs4780@s3750@val300_seed0/n/.initial=300,
  /num/abl_final_rs4780@s5625@val300_seed0/miou/.initial=0.784,
  /num/abl_final_rs4780@s5625@val300_seed0/lo/.initial=0.769,
  /num/abl_final_rs4780@s5625@val300_seed0/hi/.initial=0.800,
  /num/abl_final_rs4780@s5625@val300_seed0/dir/.initial=0.577,
  /num/abl_final_rs4780@s5625@val300_seed0/parse/.initial=1.000,
  /num/abl_final_rs4780@s5625@val300_seed0/n/.initial=300,
  /num/abl_final_rs4780@s7500@val300_seed0/miou/.initial=0.788,
  /num/abl_final_rs4780@s7500@val300_seed0/lo/.initial=0.773,
  /num/abl_final_rs4780@s7500@val300_seed0/hi/.initial=0.804,
  /num/abl_final_rs4780@s7500@val300_seed0/dir/.initial=0.577,
  /num/abl_final_rs4780@s7500@val300_seed0/parse/.initial=1.000,
  /num/abl_final_rs4780@s7500@val300_seed0/n/.initial=300,
  /num/abl_final_rs4780@s9375@val300_seed0/miou/.initial=0.791,
  /num/abl_final_rs4780@s9375@val300_seed0/lo/.initial=0.776,
  /num/abl_final_rs4780@s9375@val300_seed0/hi/.initial=0.805,
  /num/abl_final_rs4780@s9375@val300_seed0/dir/.initial=0.607,
  /num/abl_final_rs4780@s9375@val300_seed0/parse/.initial=1.000,
  /num/abl_final_rs4780@s9375@val300_seed0/n/.initial=300,
  /num/abl_reason5b3k@ep1@val300_seed0/miou/.initial=0.718,
  /num/abl_reason5b3k@ep1@val300_seed0/lo/.initial=0.701,
  /num/abl_reason5b3k@ep1@val300_seed0/hi/.initial=0.735,
  /num/abl_reason5b3k@ep1@val300_seed0/dir/.initial=0.460,
  /num/abl_reason5b3k@ep1@val300_seed0/parse/.initial=1.000,
  /num/abl_reason5b3k@ep1@val300_seed0/n/.initial=300,
  /num/abl_reason5b3k_mix@ep1@val300_seed0/miou/.initial=0.709,
  /num/abl_reason5b3k_mix@ep1@val300_seed0/lo/.initial=0.692,
  /num/abl_reason5b3k_mix@ep1@val300_seed0/hi/.initial=0.726,
  /num/abl_reason5b3k_mix@ep1@val300_seed0/dir/.initial=0.477,
  /num/abl_reason5b3k_mix@ep1@val300_seed0/parse/.initial=1.000,
  /num/abl_reason5b3k_mix@ep1@val300_seed0/n/.initial=300,
  /num/abl_reason5b3k_r1to0p5@ep1@val300_seed0/miou/.initial=0.731,
  /num/abl_reason5b3k_r1to0p5@ep1@val300_seed0/lo/.initial=0.714,
  /num/abl_reason5b3k_r1to0p5@ep1@val300_seed0/hi/.initial=0.749,
  /num/abl_reason5b3k_r1to0p5@ep1@val300_seed0/dir/.initial=0.510,
  /num/abl_reason5b3k_r1to0p5@ep1@val300_seed0/parse/.initial=1.000,
  /num/abl_reason5b3k_r1to0p5@ep1@val300_seed0/n/.initial=300,
  /num/abl_reason5b3k_r1to2@ep1@val300_seed0/miou/.initial=0.751,
  /num/abl_reason5b3k_r1to2@ep1@val300_seed0/lo/.initial=0.734,
  /num/abl_reason5b3k_r1to2@ep1@val300_seed0/hi/.initial=0.768,
  /num/abl_reason5b3k_r1to2@ep1@val300_seed0/dir/.initial=0.530,
  /num/abl_reason5b3k_r1to2@ep1@val300_seed0/parse/.initial=1.000,
  /num/abl_reason5b3k_r1to2@ep1@val300_seed0/n/.initial=300,
  /num/align_4b_reason2@test300_seed0/miou/.initial=0.641,
  /num/align_4b_reason2@test300_seed0/lo/.initial=0.624,
  /num/align_4b_reason2@test300_seed0/hi/.initial=0.658,
  /num/align_4b_reason2@test300_seed0/dir/.initial=0.270,
  /num/align_4b_reason2@test300_seed0/parse/.initial=1.000,
  /num/align_4b_reason2@test300_seed0/n/.initial=300,
  /num/align_direct_ep8@test300_seed0/miou/.initial=0.821,
  /num/align_direct_ep8@test300_seed0/lo/.initial=0.806,
  /num/align_direct_ep8@test300_seed0/hi/.initial=0.836,
  /num/align_direct_ep8@test300_seed0/dir/.initial=0.610,
  /num/align_direct_ep8@test300_seed0/parse/.initial=1.000,
  /num/align_direct_ep8@test300_seed0/n/.initial=300,
  /num/align_e36_ep4@test300_seed0/miou/.initial=0.782,
  /num/align_e36_ep4@test300_seed0/lo/.initial=0.765,
  /num/align_e36_ep4@test300_seed0/hi/.initial=0.799,
  /num/align_e36_ep4@test300_seed0/dir/.initial=0.553,
  /num/align_e36_ep4@test300_seed0/parse/.initial=1.000,
  /num/align_e36_ep4@test300_seed0/n/.initial=300,
  /num/axis2d15k@test300_seed0/miou/.initial=0.717,
  /num/axis2d15k@test300_seed0/lo/.initial=0.706,
  /num/axis2d15k@test300_seed0/hi/.initial=0.729,
  /num/axis2d15k@test300_seed0/dir/.initial=0.237,
  /num/axis2d15k@test300_seed0/parse/.initial=1.000,
  /num/axis2d15k@test300_seed0/n/.initial=300,
  /num/axis2d15k@test300_seed0/mae/.initial=12.09,
  /num/axis2d15k@test300_seed0/f1/.initial=0.650,
  /num/axis2d15k@test300_seed0/f1_5/.initial=0.777,
  /num/axis2d15k@test300_seed0/f1_15/.initial=0.577,
  /num/axis2d15k@test300_seed0/f1_20/.initial=0.085,
  /num/axis2d15k@test300_seed0/p10/.initial=0.552,
  /num/axis2d15k@test300_seed0/r10/.initial=0.791,
  /num/axis2d15k@ep1@val300_seed0/miou/.initial=0.706,
  /num/axis2d15k@ep1@val300_seed0/lo/.initial=0.696,
  /num/axis2d15k@ep1@val300_seed0/hi/.initial=0.715,
  /num/axis2d15k@ep1@val300_seed0/dir/.initial=0.307,
  /num/axis2d15k@ep1@val300_seed0/parse/.initial=1.000,
  /num/axis2d15k@ep1@val300_seed0/n/.initial=300,
  /num/axis2d15k@ep10@val300_seed0/miou/.initial=0.706,
  /num/axis2d15k@ep10@val300_seed0/lo/.initial=0.695,
  /num/axis2d15k@ep10@val300_seed0/hi/.initial=0.717,
  /num/axis2d15k@ep10@val300_seed0/dir/.initial=0.303,
  /num/axis2d15k@ep10@val300_seed0/parse/.initial=0.997,
  /num/axis2d15k@ep10@val300_seed0/n/.initial=300,
  /num/axis2d15k@ep2@val300_seed0/miou/.initial=0.702,
  /num/axis2d15k@ep2@val300_seed0/lo/.initial=0.693,
  /num/axis2d15k@ep2@val300_seed0/hi/.initial=0.711,
  /num/axis2d15k@ep2@val300_seed0/dir/.initial=0.307,
  /num/axis2d15k@ep2@val300_seed0/parse/.initial=1.000,
  /num/axis2d15k@ep2@val300_seed0/n/.initial=300,
  /num/axis2d15k@ep3@val300_seed0/miou/.initial=0.710,
  /num/axis2d15k@ep3@val300_seed0/lo/.initial=0.700,
  /num/axis2d15k@ep3@val300_seed0/hi/.initial=0.721,
  /num/axis2d15k@ep3@val300_seed0/dir/.initial=0.307,
  /num/axis2d15k@ep3@val300_seed0/parse/.initial=1.000,
  /num/axis2d15k@ep3@val300_seed0/n/.initial=300,
  /num/axis2d15k@ep4@val300_seed0/miou/.initial=0.709,
  /num/axis2d15k@ep4@val300_seed0/lo/.initial=0.699,
  /num/axis2d15k@ep4@val300_seed0/hi/.initial=0.719,
  /num/axis2d15k@ep4@val300_seed0/dir/.initial=0.307,
  /num/axis2d15k@ep4@val300_seed0/parse/.initial=1.000,
  /num/axis2d15k@ep4@val300_seed0/n/.initial=300,
  /num/axis2d15k@ep5@val300_seed0/miou/.initial=0.709,
  /num/axis2d15k@ep5@val300_seed0/lo/.initial=0.699,
  /num/axis2d15k@ep5@val300_seed0/hi/.initial=0.719,
  /num/axis2d15k@ep5@val300_seed0/dir/.initial=0.307,
  /num/axis2d15k@ep5@val300_seed0/parse/.initial=1.000,
  /num/axis2d15k@ep5@val300_seed0/n/.initial=300,
  /num/axis2d15k@ep6@val300_seed0/miou/.initial=0.707,
  /num/axis2d15k@ep6@val300_seed0/lo/.initial=0.697,
  /num/axis2d15k@ep6@val300_seed0/hi/.initial=0.717,
  /num/axis2d15k@ep6@val300_seed0/dir/.initial=0.307,
  /num/axis2d15k@ep6@val300_seed0/parse/.initial=1.000,
  /num/axis2d15k@ep6@val300_seed0/n/.initial=300,
  /num/axis2d15k@ep7@val300_seed0/miou/.initial=0.706,
  /num/axis2d15k@ep7@val300_seed0/lo/.initial=0.694,
  /num/axis2d15k@ep7@val300_seed0/hi/.initial=0.717,
  /num/axis2d15k@ep7@val300_seed0/dir/.initial=0.303,
  /num/axis2d15k@ep7@val300_seed0/parse/.initial=0.997,
  /num/axis2d15k@ep7@val300_seed0/n/.initial=300,
  /num/axis2d15k@ep8@val300_seed0/miou/.initial=0.708,
  /num/axis2d15k@ep8@val300_seed0/lo/.initial=0.698,
  /num/axis2d15k@ep8@val300_seed0/hi/.initial=0.718,
  /num/axis2d15k@ep8@val300_seed0/dir/.initial=0.307,
  /num/axis2d15k@ep8@val300_seed0/parse/.initial=1.000,
  /num/axis2d15k@ep8@val300_seed0/n/.initial=300,
  /num/axis2d15k@ep9@val300_seed0/miou/.initial=0.706,
  /num/axis2d15k@ep9@val300_seed0/lo/.initial=0.695,
  /num/axis2d15k@ep9@val300_seed0/hi/.initial=0.717,
  /num/axis2d15k@ep9@val300_seed0/dir/.initial=0.303,
  /num/axis2d15k@ep9@val300_seed0/parse/.initial=0.997,
  /num/axis2d15k@ep9@val300_seed0/n/.initial=300,
  /num/claude-haiku_icl9@test300_seed0/miou/.initial=0.688,
  /num/claude-haiku_icl9@test300_seed0/lo/.initial=0.676,
  /num/claude-haiku_icl9@test300_seed0/hi/.initial=0.701,
  /num/claude-haiku_icl9@test300_seed0/dir/.initial=0.343,
  /num/claude-haiku_icl9@test300_seed0/parse/.initial=1.000,
  /num/claude-haiku_icl9@test300_seed0/n/.initial=300,
  /num/claude-haiku_icl9@test300_seed0/mae/.initial=12.48,
  /num/claude-haiku_icl9@test300_seed0/f1/.initial=0.435,
  /num/claude-haiku_icl9@test300_seed0/f1_5/.initial=0.658,
  /num/claude-haiku_icl9@test300_seed0/f1_15/.initial=0.248,
  /num/claude-haiku_icl9@test300_seed0/f1_20/.initial=0.159,
  /num/claude-haiku_icl9@test300_seed0/p10/.initial=0.614,
  /num/claude-haiku_icl9@test300_seed0/r10/.initial=0.336,
  /num/claude-haiku_icl9txt@test300_seed0/miou/.initial=0.691,
  /num/claude-haiku_icl9txt@test300_seed0/lo/.initial=0.678,
  /num/claude-haiku_icl9txt@test300_seed0/hi/.initial=0.704,
  /num/claude-haiku_icl9txt@test300_seed0/dir/.initial=0.397,
  /num/claude-haiku_icl9txt@test300_seed0/parse/.initial=1.000,
  /num/claude-haiku_icl9txt@test300_seed0/n/.initial=300,
  /num/claude-haiku_icl9txt@test300_seed0/mae/.initial=12.25,
  /num/claude-haiku_icl9txt@test300_seed0/f1/.initial=0.389,
  /num/claude-haiku_icl9txt@test300_seed0/f1_5/.initial=0.655,
  /num/claude-haiku_icl9txt@test300_seed0/f1_15/.initial=0.223,
  /num/claude-haiku_icl9txt@test300_seed0/f1_20/.initial=0.144,
  /num/claude-haiku_icl9txt@test300_seed0/p10/.initial=0.663,
  /num/claude-haiku_icl9txt@test300_seed0/r10/.initial=0.275,
  /num/claude-haiku_zs@test300_seed0/miou/.initial=0.690,
  /num/claude-haiku_zs@test300_seed0/lo/.initial=0.678,
  /num/claude-haiku_zs@test300_seed0/hi/.initial=0.702,
  /num/claude-haiku_zs@test300_seed0/dir/.initial=0.407,
  /num/claude-haiku_zs@test300_seed0/parse/.initial=1.000,
  /num/claude-haiku_zs@test300_seed0/n/.initial=300,
  /num/claude-haiku_zs@test300_seed0/mae/.initial=12.14,
  /num/claude-haiku_zs@test300_seed0/f1/.initial=0.435,
  /num/claude-haiku_zs@test300_seed0/f1_5/.initial=0.622,
  /num/claude-haiku_zs@test300_seed0/f1_15/.initial=0.325,
  /num/claude-haiku_zs@test300_seed0/f1_20/.initial=0.204,
  /num/claude-haiku_zs@test300_seed0/p10/.initial=0.758,
  /num/claude-haiku_zs@test300_seed0/r10/.initial=0.305,
  /num/claude-opus_icl9@test300_seed0/miou/.initial=0.748,
  /num/claude-opus_icl9@test300_seed0/lo/.initial=0.737,
  /num/claude-opus_icl9@test300_seed0/hi/.initial=0.760,
  /num/claude-opus_icl9@test300_seed0/dir/.initial=0.523,
  /num/claude-opus_icl9@test300_seed0/parse/.initial=1.000,
  /num/claude-opus_icl9@test300_seed0/n/.initial=300,
  /num/claude-opus_icl9@test300_seed0/mae/.initial=9.96,
  /num/claude-opus_icl9@test300_seed0/f1/.initial=0.686,
  /num/claude-opus_icl9@test300_seed0/f1_5/.initial=0.774,
  /num/claude-opus_icl9@test300_seed0/f1_15/.initial=0.585,
  /num/claude-opus_icl9@test300_seed0/f1_20/.initial=0.468,
  /num/claude-opus_icl9@test300_seed0/p10/.initial=0.727,
  /num/claude-opus_icl9@test300_seed0/r10/.initial=0.649,
  /num/claude-opus_icl9_reason@test300_seed0/miou/.initial=0.751,
  /num/claude-opus_icl9_reason@test300_seed0/lo/.initial=0.739,
  /num/claude-opus_icl9_reason@test300_seed0/hi/.initial=0.762,
  /num/claude-opus_icl9_reason@test300_seed0/dir/.initial=0.523,
  /num/claude-opus_icl9_reason@test300_seed0/parse/.initial=1.000,
  /num/claude-opus_icl9_reason@test300_seed0/n/.initial=300,
  /num/claude-opus_icl9_set1@test300_seed0/miou/.initial=0.747,
  /num/claude-opus_icl9_set1@test300_seed0/lo/.initial=0.735,
  /num/claude-opus_icl9_set1@test300_seed0/hi/.initial=0.758,
  /num/claude-opus_icl9_set1@test300_seed0/dir/.initial=0.550,
  /num/claude-opus_icl9_set1@test300_seed0/parse/.initial=1.000,
  /num/claude-opus_icl9_set1@test300_seed0/n/.initial=300,
  /num/claude-opus_icl9_set2@test300_seed0/miou/.initial=0.747,
  /num/claude-opus_icl9_set2@test300_seed0/lo/.initial=0.735,
  /num/claude-opus_icl9_set2@test300_seed0/hi/.initial=0.759,
  /num/claude-opus_icl9_set2@test300_seed0/dir/.initial=0.510,
  /num/claude-opus_icl9_set2@test300_seed0/parse/.initial=1.000,
  /num/claude-opus_icl9_set2@test300_seed0/n/.initial=300,
  /num/claude-opus_icl9txt@test300_seed0/miou/.initial=0.742,
  /num/claude-opus_icl9txt@test300_seed0/lo/.initial=0.730,
  /num/claude-opus_icl9txt@test300_seed0/hi/.initial=0.753,
  /num/claude-opus_icl9txt@test300_seed0/dir/.initial=0.543,
  /num/claude-opus_icl9txt@test300_seed0/parse/.initial=1.000,
  /num/claude-opus_icl9txt@test300_seed0/n/.initial=300,
  /num/claude-opus_icl9txt@test300_seed0/mae/.initial=10.23,
  /num/claude-opus_icl9txt@test300_seed0/f1/.initial=0.666,
  /num/claude-opus_icl9txt@test300_seed0/f1_5/.initial=0.770,
  /num/claude-opus_icl9txt@test300_seed0/f1_15/.initial=0.546,
  /num/claude-opus_icl9txt@test300_seed0/f1_20/.initial=0.434,
  /num/claude-opus_icl9txt@test300_seed0/p10/.initial=0.741,
  /num/claude-opus_icl9txt@test300_seed0/r10/.initial=0.605,
  /num/claude-opus_zs@test300_seed0/miou/.initial=0.724,
  /num/claude-opus_zs@test300_seed0/lo/.initial=0.712,
  /num/claude-opus_zs@test300_seed0/hi/.initial=0.736,
  /num/claude-opus_zs@test300_seed0/dir/.initial=0.520,
  /num/claude-opus_zs@test300_seed0/parse/.initial=1.000,
  /num/claude-opus_zs@test300_seed0/n/.initial=300,
  /num/claude-opus_zs@test300_seed0/mae/.initial=10.77,
  /num/claude-opus_zs@test300_seed0/f1/.initial=0.585,
  /num/claude-opus_zs@test300_seed0/f1_5/.initial=0.708,
  /num/claude-opus_zs@test300_seed0/f1_15/.initial=0.464,
  /num/claude-opus_zs@test300_seed0/f1_20/.initial=0.375,
  /num/claude-opus_zs@test300_seed0/p10/.initial=0.771,
  /num/claude-opus_zs@test300_seed0/r10/.initial=0.471,
  /num/claude-sonnet_icl9@test300_seed0/miou/.initial=0.730,
  /num/claude-sonnet_icl9@test300_seed0/lo/.initial=0.718,
  /num/claude-sonnet_icl9@test300_seed0/hi/.initial=0.742,
  /num/claude-sonnet_icl9@test300_seed0/dir/.initial=0.433,
  /num/claude-sonnet_icl9@test300_seed0/parse/.initial=1.000,
  /num/claude-sonnet_icl9@test300_seed0/n/.initial=300,
  /num/claude-sonnet_icl9@test300_seed0/mae/.initial=10.90,
  /num/claude-sonnet_icl9@test300_seed0/f1/.initial=0.645,
  /num/claude-sonnet_icl9@test300_seed0/f1_5/.initial=0.764,
  /num/claude-sonnet_icl9@test300_seed0/f1_15/.initial=0.467,
  /num/claude-sonnet_icl9@test300_seed0/f1_20/.initial=0.410,
  /num/claude-sonnet_icl9@test300_seed0/p10/.initial=0.679,
  /num/claude-sonnet_icl9@test300_seed0/r10/.initial=0.614,
  /num/claude-sonnet_icl9txt@test300_seed0/miou/.initial=0.728,
  /num/claude-sonnet_icl9txt@test300_seed0/lo/.initial=0.717,
  /num/claude-sonnet_icl9txt@test300_seed0/hi/.initial=0.740,
  /num/claude-sonnet_icl9txt@test300_seed0/dir/.initial=0.430,
  /num/claude-sonnet_icl9txt@test300_seed0/parse/.initial=1.000,
  /num/claude-sonnet_icl9txt@test300_seed0/n/.initial=300,
  /num/claude-sonnet_icl9txt@test300_seed0/mae/.initial=10.65,
  /num/claude-sonnet_icl9txt@test300_seed0/f1/.initial=0.583,
  /num/claude-sonnet_icl9txt@test300_seed0/f1_5/.initial=0.752,
  /num/claude-sonnet_icl9txt@test300_seed0/f1_15/.initial=0.422,
  /num/claude-sonnet_icl9txt@test300_seed0/f1_20/.initial=0.362,
  /num/claude-sonnet_icl9txt@test300_seed0/p10/.initial=0.718,
  /num/claude-sonnet_icl9txt@test300_seed0/r10/.initial=0.491,
  /num/claude-sonnet_zs@test300_seed0/miou/.initial=0.727,
  /num/claude-sonnet_zs@test300_seed0/lo/.initial=0.715,
  /num/claude-sonnet_zs@test300_seed0/hi/.initial=0.738,
  /num/claude-sonnet_zs@test300_seed0/dir/.initial=0.447,
  /num/claude-sonnet_zs@test300_seed0/parse/.initial=1.000,
  /num/claude-sonnet_zs@test300_seed0/n/.initial=300,
  /num/claude-sonnet_zs@test300_seed0/mae/.initial=10.83,
  /num/claude-sonnet_zs@test300_seed0/f1/.initial=0.605,
  /num/claude-sonnet_zs@test300_seed0/f1_5/.initial=0.741,
  /num/claude-sonnet_zs@test300_seed0/f1_15/.initial=0.446,
  /num/claude-sonnet_zs@test300_seed0/f1_20/.initial=0.396,
  /num/claude-sonnet_zs@test300_seed0/p10/.initial=0.705,
  /num/claude-sonnet_zs@test300_seed0/r10/.initial=0.529,
  /num/clip_mlp_s0@test300_seed0/miou/.initial=0.712,
  /num/clip_mlp_s0@test300_seed0/lo/.initial=0.701,
  /num/clip_mlp_s0@test300_seed0/hi/.initial=0.722,
  /num/clip_mlp_s0@test300_seed0/dir/.initial=0.380,
  /num/clip_mlp_s0@test300_seed0/parse/.initial=1.000,
  /num/clip_mlp_s0@test300_seed0/n/.initial=300,
  /num/clip_mlp_s0@test300_seed0/mae/.initial=12.02,
  /num/clip_mlp_s0@test300_seed0/f1/.initial=0.663,
  /num/clip_mlp_s0@test300_seed0/f1_5/.initial=0.776,
  /num/clip_mlp_s0@test300_seed0/f1_15/.initial=0.519,
  /num/clip_mlp_s0@test300_seed0/f1_20/.initial=0.386,
  /num/clip_mlp_s0@test300_seed0/p10/.initial=0.555,
  /num/clip_mlp_s0@test300_seed0/r10/.initial=0.824,
  /num/clip_mlp_s1@test300_seed0/miou/.initial=0.716,
  /num/clip_mlp_s1@test300_seed0/lo/.initial=0.706,
  /num/clip_mlp_s1@test300_seed0/hi/.initial=0.727,
  /num/clip_mlp_s1@test300_seed0/dir/.initial=0.373,
  /num/clip_mlp_s1@test300_seed0/parse/.initial=1.000,
  /num/clip_mlp_s1@test300_seed0/n/.initial=300,
  /num/clip_mlp_s1@test300_seed0/mae/.initial=11.71,
  /num/clip_mlp_s1@test300_seed0/f1/.initial=0.664,
  /num/clip_mlp_s1@test300_seed0/f1_5/.initial=0.776,
  /num/clip_mlp_s1@test300_seed0/f1_15/.initial=0.539,
  /num/clip_mlp_s1@test300_seed0/f1_20/.initial=0.356,
  /num/clip_mlp_s1@test300_seed0/p10/.initial=0.548,
  /num/clip_mlp_s1@test300_seed0/r10/.initial=0.843,
  /num/clip_mlp_s2@test300_seed0/miou/.initial=0.710,
  /num/clip_mlp_s2@test300_seed0/lo/.initial=0.700,
  /num/clip_mlp_s2@test300_seed0/hi/.initial=0.720,
  /num/clip_mlp_s2@test300_seed0/dir/.initial=0.317,
  /num/clip_mlp_s2@test300_seed0/parse/.initial=1.000,
  /num/clip_mlp_s2@test300_seed0/n/.initial=300,
  /num/clip_mlp_s2@test300_seed0/mae/.initial=12.06,
  /num/clip_mlp_s2@test300_seed0/f1/.initial=0.668,
  /num/clip_mlp_s2@test300_seed0/f1_5/.initial=0.770,
  /num/clip_mlp_s2@test300_seed0/f1_15/.initial=0.521,
  /num/clip_mlp_s2@test300_seed0/f1_20/.initial=0.369,
  /num/clip_mlp_s2@test300_seed0/p10/.initial=0.559,
  /num/clip_mlp_s2@test300_seed0/r10/.initial=0.830,
  /num/cnn_s0@test300_seed0/miou/.initial=0.712,
  /num/cnn_s0@test300_seed0/lo/.initial=0.700,
  /num/cnn_s0@test300_seed0/hi/.initial=0.723,
  /num/cnn_s0@test300_seed0/dir/.initial=0.400,
  /num/cnn_s0@test300_seed0/parse/.initial=1.000,
  /num/cnn_s0@test300_seed0/n/.initial=300,
  /num/cnn_s0@test300_seed0/mae/.initial=11.95,
  /num/cnn_s0@test300_seed0/f1/.initial=0.660,
  /num/cnn_s0@test300_seed0/f1_5/.initial=0.771,
  /num/cnn_s0@test300_seed0/f1_15/.initial=0.515,
  /num/cnn_s0@test300_seed0/f1_20/.initial=0.321,
  /num/cnn_s0@test300_seed0/p10/.initial=0.545,
  /num/cnn_s0@test300_seed0/r10/.initial=0.837,
  /num/cnn_s1@test300_seed0/miou/.initial=0.709,
  /num/cnn_s1@test300_seed0/lo/.initial=0.697,
  /num/cnn_s1@test300_seed0/hi/.initial=0.720,
  /num/cnn_s1@test300_seed0/dir/.initial=0.350,
  /num/cnn_s1@test300_seed0/parse/.initial=1.000,
  /num/cnn_s1@test300_seed0/n/.initial=300,
  /num/cnn_s1@test300_seed0/mae/.initial=12.12,
  /num/cnn_s1@test300_seed0/f1/.initial=0.657,
  /num/cnn_s1@test300_seed0/f1_5/.initial=0.770,
  /num/cnn_s1@test300_seed0/f1_15/.initial=0.534,
  /num/cnn_s1@test300_seed0/f1_20/.initial=0.374,
  /num/cnn_s1@test300_seed0/p10/.initial=0.542,
  /num/cnn_s1@test300_seed0/r10/.initial=0.832,
  /num/cnn_s2@test300_seed0/miou/.initial=0.708,
  /num/cnn_s2@test300_seed0/lo/.initial=0.698,
  /num/cnn_s2@test300_seed0/hi/.initial=0.719,
  /num/cnn_s2@test300_seed0/dir/.initial=0.400,
  /num/cnn_s2@test300_seed0/parse/.initial=1.000,
  /num/cnn_s2@test300_seed0/n/.initial=300,
  /num/cnn_s2@test300_seed0/mae/.initial=12.16,
  /num/cnn_s2@test300_seed0/f1/.initial=0.668,
  /num/cnn_s2@test300_seed0/f1_5/.initial=0.771,
  /num/cnn_s2@test300_seed0/f1_15/.initial=0.536,
  /num/cnn_s2@test300_seed0/f1_20/.initial=0.332,
  /num/cnn_s2@test300_seed0/p10/.initial=0.529,
  /num/cnn_s2@test300_seed0/r10/.initial=0.906,
  /num/constant@test300_seed0/miou/.initial=0.691,
  /num/constant@test300_seed0/lo/.initial=0.682,
  /num/constant@test300_seed0/hi/.initial=0.701,
  /num/constant@test300_seed0/dir/.initial=0.323,
  /num/constant@test300_seed0/parse/.initial=1.000,
  /num/constant@test300_seed0/n/.initial=300,
  /num/constant@test300_seed0/mae/.initial=13.11,
  /num/constant@test300_seed0/f1/.initial=0.672,
  /num/constant@test300_seed0/f1_5/.initial=0.771,
  /num/constant@test300_seed0/f1_15/.initial=0.566,
  /num/constant@test300_seed0/f1_20/.initial=0.000,
  /num/constant@test300_seed0/p10/.initial=0.506,
  /num/constant@test300_seed0/r10/.initial=1.000,
  /num/constant@test_full/miou/.initial=0.692,
  /num/constant@test_full/lo/.initial=0.690,
  /num/constant@test_full/hi/.initial=0.695,
  /num/constant@test_full/dir/.initial=0.295,
  /num/constant@test_full/parse/.initial=1.000,
  /num/constant@test_full/n/.initial=3807,
  /num/dino_cls_s0@test300_seed0/miou/.initial=0.722,
  /num/dino_cls_s0@test300_seed0/lo/.initial=0.708,
  /num/dino_cls_s0@test300_seed0/hi/.initial=0.735,
  /num/dino_cls_s0@test300_seed0/dir/.initial=0.437,
  /num/dino_cls_s0@test300_seed0/parse/.initial=1.000,
  /num/dino_cls_s0@test300_seed0/n/.initial=300,
  /num/dino_cls_s0@test300_seed0/mae/.initial=11.65,
  /num/dino_cls_s0@test300_seed0/f1/.initial=0.660,
  /num/dino_cls_s0@test300_seed0/f1_5/.initial=0.761,
  /num/dino_cls_s0@test300_seed0/f1_15/.initial=0.603,
  /num/dino_cls_s0@test300_seed0/f1_20/.initial=0.523,
  /num/dino_cls_s0@test300_seed0/p10/.initial=0.622,
  /num/dino_cls_s0@test300_seed0/r10/.initial=0.703,
  /num/dino_cls_s1@test300_seed0/miou/.initial=0.727,
  /num/dino_cls_s1@test300_seed0/lo/.initial=0.713,
  /num/dino_cls_s1@test300_seed0/hi/.initial=0.740,
  /num/dino_cls_s1@test300_seed0/dir/.initial=0.417,
  /num/dino_cls_s1@test300_seed0/parse/.initial=1.000,
  /num/dino_cls_s1@test300_seed0/n/.initial=300,
  /num/dino_cls_s1@test300_seed0/mae/.initial=11.34,
  /num/dino_cls_s1@test300_seed0/f1/.initial=0.668,
  /num/dino_cls_s1@test300_seed0/f1_5/.initial=0.762,
  /num/dino_cls_s1@test300_seed0/f1_15/.initial=0.595,
  /num/dino_cls_s1@test300_seed0/f1_20/.initial=0.549,
  /num/dino_cls_s1@test300_seed0/p10/.initial=0.621,
  /num/dino_cls_s1@test300_seed0/r10/.initial=0.722,
  /num/dino_cls_s2@test300_seed0/miou/.initial=0.726,
  /num/dino_cls_s2@test300_seed0/lo/.initial=0.713,
  /num/dino_cls_s2@test300_seed0/hi/.initial=0.739,
  /num/dino_cls_s2@test300_seed0/dir/.initial=0.420,
  /num/dino_cls_s2@test300_seed0/parse/.initial=1.000,
  /num/dino_cls_s2@test300_seed0/n/.initial=300,
  /num/dino_cls_s2@test300_seed0/mae/.initial=11.40,
  /num/dino_cls_s2@test300_seed0/f1/.initial=0.676,
  /num/dino_cls_s2@test300_seed0/f1_5/.initial=0.763,
  /num/dino_cls_s2@test300_seed0/f1_15/.initial=0.615,
  /num/dino_cls_s2@test300_seed0/f1_20/.initial=0.554,
  /num/dino_cls_s2@test300_seed0/p10/.initial=0.618,
  /num/dino_cls_s2@test300_seed0/r10/.initial=0.746,
  /num/dino_grid3_s0@test300_seed0/miou/.initial=0.781,
  /num/dino_grid3_s0@test300_seed0/lo/.initial=0.769,
  /num/dino_grid3_s0@test300_seed0/hi/.initial=0.793,
  /num/dino_grid3_s0@test300_seed0/dir/.initial=0.630,
  /num/dino_grid3_s0@test300_seed0/parse/.initial=1.000,
  /num/dino_grid3_s0@test300_seed0/n/.initial=300,
  /num/dino_grid3_s0@test300_seed0/mae/.initial=8.76,
  /num/dino_grid3_s0@test300_seed0/f1/.initial=0.772,
  /num/dino_grid3_s0@test300_seed0/f1_5/.initial=0.820,
  /num/dino_grid3_s0@test300_seed0/f1_15/.initial=0.702,
  /num/dino_grid3_s0@test300_seed0/f1_20/.initial=0.691,
  /num/dino_grid3_s0@test300_seed0/p10/.initial=0.715,
  /num/dino_grid3_s0@test300_seed0/r10/.initial=0.839,
  /num/dino_grid3_s1@test300_seed0/miou/.initial=0.787,
  /num/dino_grid3_s1@test300_seed0/lo/.initial=0.775,
  /num/dino_grid3_s1@test300_seed0/hi/.initial=0.799,
  /num/dino_grid3_s1@test300_seed0/dir/.initial=0.640,
  /num/dino_grid3_s1@test300_seed0/parse/.initial=1.000,
  /num/dino_grid3_s1@test300_seed0/n/.initial=300,
  /num/dino_grid3_s1@test300_seed0/mae/.initial=8.49,
  /num/dino_grid3_s1@test300_seed0/f1/.initial=0.767,
  /num/dino_grid3_s1@test300_seed0/f1_5/.initial=0.820,
  /num/dino_grid3_s1@test300_seed0/f1_15/.initial=0.716,
  /num/dino_grid3_s1@test300_seed0/f1_20/.initial=0.687,
  /num/dino_grid3_s1@test300_seed0/p10/.initial=0.717,
  /num/dino_grid3_s1@test300_seed0/r10/.initial=0.825,
  /num/dino_grid3_s2@test300_seed0/miou/.initial=0.781,
  /num/dino_grid3_s2@test300_seed0/lo/.initial=0.768,
  /num/dino_grid3_s2@test300_seed0/hi/.initial=0.793,
  /num/dino_grid3_s2@test300_seed0/dir/.initial=0.597,
  /num/dino_grid3_s2@test300_seed0/parse/.initial=1.000,
  /num/dino_grid3_s2@test300_seed0/n/.initial=300,
  /num/dino_grid3_s2@test300_seed0/mae/.initial=8.80,
  /num/dino_grid3_s2@test300_seed0/f1/.initial=0.765,
  /num/dino_grid3_s2@test300_seed0/f1_5/.initial=0.820,
  /num/dino_grid3_s2@test300_seed0/f1_15/.initial=0.710,
  /num/dino_grid3_s2@test300_seed0/f1_20/.initial=0.682,
  /num/dino_grid3_s2@test300_seed0/p10/.initial=0.723,
  /num/dino_grid3_s2@test300_seed0/r10/.initial=0.812,
  /num/dino_mean_s0@test300_seed0/miou/.initial=0.774,
  /num/dino_mean_s0@test300_seed0/lo/.initial=0.761,
  /num/dino_mean_s0@test300_seed0/hi/.initial=0.786,
  /num/dino_mean_s0@test300_seed0/dir/.initial=0.543,
  /num/dino_mean_s0@test300_seed0/parse/.initial=1.000,
  /num/dino_mean_s0@test300_seed0/n/.initial=300,
  /num/dino_mean_s0@test300_seed0/mae/.initial=9.06,
  /num/dino_mean_s0@test300_seed0/f1/.initial=0.747,
  /num/dino_mean_s0@test300_seed0/f1_5/.initial=0.808,
  /num/dino_mean_s0@test300_seed0/f1_15/.initial=0.686,
  /num/dino_mean_s0@test300_seed0/f1_20/.initial=0.655,
  /num/dino_mean_s0@test300_seed0/p10/.initial=0.697,
  /num/dino_mean_s0@test300_seed0/r10/.initial=0.806,
  /num/dino_mean_s1@test300_seed0/miou/.initial=0.774,
  /num/dino_mean_s1@test300_seed0/lo/.initial=0.761,
  /num/dino_mean_s1@test300_seed0/hi/.initial=0.787,
  /num/dino_mean_s1@test300_seed0/dir/.initial=0.587,
  /num/dino_mean_s1@test300_seed0/parse/.initial=1.000,
  /num/dino_mean_s1@test300_seed0/n/.initial=300,
  /num/dino_mean_s1@test300_seed0/mae/.initial=9.14,
  /num/dino_mean_s1@test300_seed0/f1/.initial=0.751,
  /num/dino_mean_s1@test300_seed0/f1_5/.initial=0.805,
  /num/dino_mean_s1@test300_seed0/f1_15/.initial=0.687,
  /num/dino_mean_s1@test300_seed0/f1_20/.initial=0.675,
  /num/dino_mean_s1@test300_seed0/p10/.initial=0.709,
  /num/dino_mean_s1@test300_seed0/r10/.initial=0.799,
  /num/dino_mean_s2@test300_seed0/miou/.initial=0.772,
  /num/dino_mean_s2@test300_seed0/lo/.initial=0.760,
  /num/dino_mean_s2@test300_seed0/hi/.initial=0.784,
  /num/dino_mean_s2@test300_seed0/dir/.initial=0.547,
  /num/dino_mean_s2@test300_seed0/parse/.initial=1.000,
  /num/dino_mean_s2@test300_seed0/n/.initial=300,
  /num/dino_mean_s2@test300_seed0/mae/.initial=9.17,
  /num/dino_mean_s2@test300_seed0/f1/.initial=0.758,
  /num/dino_mean_s2@test300_seed0/f1_5/.initial=0.809,
  /num/dino_mean_s2@test300_seed0/f1_15/.initial=0.710,
  /num/dino_mean_s2@test300_seed0/f1_20/.initial=0.672,
  /num/dino_mean_s2@test300_seed0/p10/.initial=0.696,
  /num/dino_mean_s2@test300_seed0/r10/.initial=0.832,
  /num/dino_mlp_s0@test300_seed0/miou/.initial=0.726,
  /num/dino_mlp_s0@test300_seed0/lo/.initial=0.715,
  /num/dino_mlp_s0@test300_seed0/hi/.initial=0.738,
  /num/dino_mlp_s0@test300_seed0/dir/.initial=0.397,
  /num/dino_mlp_s0@test300_seed0/parse/.initial=1.000,
  /num/dino_mlp_s0@test300_seed0/n/.initial=300,
  /num/dino_mlp_s0@test300_seed0/mae/.initial=11.31,
  /num/dino_mlp_s0@test300_seed0/f1/.initial=0.678,
  /num/dino_mlp_s0@test300_seed0/f1_5/.initial=0.774,
  /num/dino_mlp_s0@test300_seed0/f1_15/.initial=0.580,
  /num/dino_mlp_s0@test300_seed0/f1_20/.initial=0.480,
  /num/dino_mlp_s0@test300_seed0/p10/.initial=0.603,
  /num/dino_mlp_s0@test300_seed0/r10/.initial=0.774,
  /num/dino_mlp_s1@test300_seed0/miou/.initial=0.721,
  /num/dino_mlp_s1@test300_seed0/lo/.initial=0.709,
  /num/dino_mlp_s1@test300_seed0/hi/.initial=0.732,
  /num/dino_mlp_s1@test300_seed0/dir/.initial=0.437,
  /num/dino_mlp_s1@test300_seed0/parse/.initial=1.000,
  /num/dino_mlp_s1@test300_seed0/n/.initial=300,
  /num/dino_mlp_s1@test300_seed0/mae/.initial=11.62,
  /num/dino_mlp_s1@test300_seed0/f1/.initial=0.678,
  /num/dino_mlp_s1@test300_seed0/f1_5/.initial=0.774,
  /num/dino_mlp_s1@test300_seed0/f1_15/.initial=0.573,
  /num/dino_mlp_s1@test300_seed0/f1_20/.initial=0.503,
  /num/dino_mlp_s1@test300_seed0/p10/.initial=0.606,
  /num/dino_mlp_s1@test300_seed0/r10/.initial=0.769,
  /num/dino_mlp_s2@test300_seed0/miou/.initial=0.723,
  /num/dino_mlp_s2@test300_seed0/lo/.initial=0.712,
  /num/dino_mlp_s2@test300_seed0/hi/.initial=0.733,
  /num/dino_mlp_s2@test300_seed0/dir/.initial=0.440,
  /num/dino_mlp_s2@test300_seed0/parse/.initial=1.000,
  /num/dino_mlp_s2@test300_seed0/n/.initial=300,
  /num/dino_mlp_s2@test300_seed0/mae/.initial=11.49,
  /num/dino_mlp_s2@test300_seed0/f1/.initial=0.673,
  /num/dino_mlp_s2@test300_seed0/f1_5/.initial=0.773,
  /num/dino_mlp_s2@test300_seed0/f1_15/.initial=0.601,
  /num/dino_mlp_s2@test300_seed0/f1_20/.initial=0.515,
  /num/dino_mlp_s2@test300_seed0/p10/.initial=0.590,
  /num/dino_mlp_s2@test300_seed0/r10/.initial=0.783,
  /num/direct15k@test300_seed0/miou/.initial=0.789,
  /num/direct15k@test300_seed0/lo/.initial=0.775,
  /num/direct15k@test300_seed0/hi/.initial=0.804,
  /num/direct15k@test300_seed0/dir/.initial=0.587,
  /num/direct15k@test300_seed0/parse/.initial=1.000,
  /num/direct15k@test300_seed0/n/.initial=300,
  /num/direct15k@test300_seed0/mae/.initial=8.58,
  /num/direct15k@test300_seed0/f1/.initial=0.748,
  /num/direct15k@test300_seed0/f1_5/.initial=0.797,
  /num/direct15k@test300_seed0/f1_15/.initial=0.707,
  /num/direct15k@test300_seed0/f1_20/.initial=0.687,
  /num/direct15k@test300_seed0/p10/.initial=0.731,
  /num/direct15k@test300_seed0/r10/.initial=0.766,
  /num/direct15k@ep1@val300_seed0/miou/.initial=0.771,
  /num/direct15k@ep1@val300_seed0/lo/.initial=0.754,
  /num/direct15k@ep1@val300_seed0/hi/.initial=0.787,
  /num/direct15k@ep1@val300_seed0/dir/.initial=0.577,
  /num/direct15k@ep1@val300_seed0/parse/.initial=1.000,
  /num/direct15k@ep1@val300_seed0/n/.initial=300,
  /num/direct15k@ep10@val300_seed0/miou/.initial=0.807,
  /num/direct15k@ep10@val300_seed0/lo/.initial=0.794,
  /num/direct15k@ep10@val300_seed0/hi/.initial=0.822,
  /num/direct15k@ep10@val300_seed0/dir/.initial=0.607,
  /num/direct15k@ep10@val300_seed0/parse/.initial=1.000,
  /num/direct15k@ep10@val300_seed0/n/.initial=300,
  /num/direct15k@ep2@val300_seed0/miou/.initial=0.789,
  /num/direct15k@ep2@val300_seed0/lo/.initial=0.773,
  /num/direct15k@ep2@val300_seed0/hi/.initial=0.805,
  /num/direct15k@ep2@val300_seed0/dir/.initial=0.617,
  /num/direct15k@ep2@val300_seed0/parse/.initial=1.000,
  /num/direct15k@ep2@val300_seed0/n/.initial=300,
  /num/direct15k@ep3@val300_seed0/miou/.initial=0.798,
  /num/direct15k@ep3@val300_seed0/lo/.initial=0.783,
  /num/direct15k@ep3@val300_seed0/hi/.initial=0.813,
  /num/direct15k@ep3@val300_seed0/dir/.initial=0.597,
  /num/direct15k@ep3@val300_seed0/parse/.initial=1.000,
  /num/direct15k@ep3@val300_seed0/n/.initial=300,
  /num/direct15k@ep4@val300_seed0/miou/.initial=0.798,
  /num/direct15k@ep4@val300_seed0/lo/.initial=0.783,
  /num/direct15k@ep4@val300_seed0/hi/.initial=0.814,
  /num/direct15k@ep4@val300_seed0/dir/.initial=0.583,
  /num/direct15k@ep4@val300_seed0/parse/.initial=1.000,
  /num/direct15k@ep4@val300_seed0/n/.initial=300,
  /num/direct15k@ep5@val300_seed0/miou/.initial=0.809,
  /num/direct15k@ep5@val300_seed0/lo/.initial=0.795,
  /num/direct15k@ep5@val300_seed0/hi/.initial=0.824,
  /num/direct15k@ep5@val300_seed0/dir/.initial=0.603,
  /num/direct15k@ep5@val300_seed0/parse/.initial=1.000,
  /num/direct15k@ep5@val300_seed0/n/.initial=300,
  /num/direct15k@ep6@val300_seed0/miou/.initial=0.805,
  /num/direct15k@ep6@val300_seed0/lo/.initial=0.791,
  /num/direct15k@ep6@val300_seed0/hi/.initial=0.819,
  /num/direct15k@ep6@val300_seed0/dir/.initial=0.597,
  /num/direct15k@ep6@val300_seed0/parse/.initial=1.000,
  /num/direct15k@ep6@val300_seed0/n/.initial=300,
  /num/direct15k@ep7@val300_seed0/miou/.initial=0.805,
  /num/direct15k@ep7@val300_seed0/lo/.initial=0.791,
  /num/direct15k@ep7@val300_seed0/hi/.initial=0.819,
  /num/direct15k@ep7@val300_seed0/dir/.initial=0.607,
  /num/direct15k@ep7@val300_seed0/parse/.initial=1.000,
  /num/direct15k@ep7@val300_seed0/n/.initial=300,
  /num/direct15k@ep8@val300_seed0/miou/.initial=0.805,
  /num/direct15k@ep8@val300_seed0/lo/.initial=0.792,
  /num/direct15k@ep8@val300_seed0/hi/.initial=0.820,
  /num/direct15k@ep8@val300_seed0/dir/.initial=0.600,
  /num/direct15k@ep8@val300_seed0/parse/.initial=1.000,
  /num/direct15k@ep8@val300_seed0/n/.initial=300,
  /num/direct15k@ep9@val300_seed0/miou/.initial=0.807,
  /num/direct15k@ep9@val300_seed0/lo/.initial=0.794,
  /num/direct15k@ep9@val300_seed0/hi/.initial=0.822,
  /num/direct15k@ep9@val300_seed0/dir/.initial=0.613,
  /num/direct15k@ep9@val300_seed0/parse/.initial=1.000,
  /num/direct15k@ep9@val300_seed0/n/.initial=300,
  /num/direct15k_balanced@test300_seed0/miou/.initial=0.716,
  /num/direct15k_balanced@test300_seed0/lo/.initial=0.706,
  /num/direct15k_balanced@test300_seed0/hi/.initial=0.728,
  /num/direct15k_balanced@test300_seed0/dir/.initial=0.237,
  /num/direct15k_balanced@test300_seed0/parse/.initial=1.000,
  /num/direct15k_balanced@test300_seed0/n/.initial=300,
  /num/direct15k_balanced@test300_seed0/mae/.initial=12.04,
  /num/direct15k_balanced@test300_seed0/f1/.initial=0.669,
  /num/direct15k_balanced@test300_seed0/f1_5/.initial=0.780,
  /num/direct15k_balanced@test300_seed0/f1_15/.initial=0.570,
  /num/direct15k_balanced@test300_seed0/f1_20/.initial=0.258,
  /num/direct15k_balanced@test300_seed0/p10/.initial=0.548,
  /num/direct15k_balanced@test300_seed0/r10/.initial=0.858,
  /num/direct15k_bin30@test300_seed0/miou/.initial=0.731,
  /num/direct15k_bin30@test300_seed0/lo/.initial=0.717,
  /num/direct15k_bin30@test300_seed0/hi/.initial=0.745,
  /num/direct15k_bin30@test300_seed0/dir/.initial=0.553,
  /num/direct15k_bin30@test300_seed0/parse/.initial=1.000,
  /num/direct15k_bin30@test300_seed0/n/.initial=300,
  /num/direct15k_bin30@test300_seed0/mae/.initial=11.01,
  /num/direct15k_bin30@test300_seed0/f1/.initial=0.739,
  /num/direct15k_bin30@test300_seed0/f1_5/.initial=0.740,
  /num/direct15k_bin30@test300_seed0/f1_15/.initial=0.698,
  /num/direct15k_bin30@test300_seed0/f1_20/.initial=0.653,
  /num/direct15k_bin30@test300_seed0/p10/.initial=0.755,
  /num/direct15k_bin30@test300_seed0/r10/.initial=0.725,
  /num/direct15k_fixmag@test300_seed0/miou/.initial=0.744,
  /num/direct15k_fixmag@test300_seed0/lo/.initial=0.730,
  /num/direct15k_fixmag@test300_seed0/hi/.initial=0.757,
  /num/direct15k_fixmag@test300_seed0/dir/.initial=0.583,
  /num/direct15k_fixmag@test300_seed0/parse/.initial=1.000,
  /num/direct15k_fixmag@test300_seed0/n/.initial=300,
  /num/direct15k_fixmag@test300_seed0/mae/.initial=10.64,
  /num/direct15k_fixmag@test300_seed0/f1/.initial=0.742,
  /num/direct15k_fixmag@test300_seed0/f1_5/.initial=0.790,
  /num/direct15k_fixmag@test300_seed0/f1_15/.initial=0.673,
  /num/direct15k_fixmag@test300_seed0/f1_20/.initial=0.630,
  /num/direct15k_fixmag@test300_seed0/p10/.initial=0.684,
  /num/direct15k_fixmag@test300_seed0/r10/.initial=0.812,
  /num/direct15k_q10@test300_seed0/miou/.initial=0.775,
  /num/direct15k_q10@test300_seed0/lo/.initial=0.761,
  /num/direct15k_q10@test300_seed0/hi/.initial=0.789,
  /num/direct15k_q10@test300_seed0/dir/.initial=0.560,
  /num/direct15k_q10@test300_seed0/parse/.initial=1.000,
  /num/direct15k_q10@test300_seed0/n/.initial=300,
  /num/direct15k_q10@test300_seed0/mae/.initial=9.20,
  /num/direct15k_q10@test300_seed0/f1/.initial=0.739,
  /num/direct15k_q10@test300_seed0/f1_5/.initial=0.797,
  /num/direct15k_q10@test300_seed0/f1_15/.initial=0.698,
  /num/direct15k_q10@test300_seed0/f1_20/.initial=0.670,
  /num/direct15k_q10@test300_seed0/p10/.initial=0.755,
  /num/direct15k_q10@test300_seed0/r10/.initial=0.725,
  /num/direct15k_q25@test300_seed0/miou/.initial=0.736,
  /num/direct15k_q25@test300_seed0/lo/.initial=0.723,
  /num/direct15k_q25@test300_seed0/hi/.initial=0.749,
  /num/direct15k_q25@test300_seed0/dir/.initial=0.563,
  /num/direct15k_q25@test300_seed0/parse/.initial=1.000,
  /num/direct15k_q25@test300_seed0/n/.initial=300,
  /num/direct15k_q25@test300_seed0/mae/.initial=10.92,
  /num/direct15k_q25@test300_seed0/f1/.initial=0.749,
  /num/direct15k_q25@test300_seed0/f1_5/.initial=0.753,
  /num/direct15k_q25@test300_seed0/f1_15/.initial=0.707,
  /num/direct15k_q25@test300_seed0/f1_20/.initial=0.655,
  /num/direct15k_q25@test300_seed0/p10/.initial=0.748,
  /num/direct15k_q25@test300_seed0/r10/.initial=0.751,
  /num/direct15k_seed1@test300_seed0/miou/.initial=0.812,
  /num/direct15k_seed1@test300_seed0/lo/.initial=0.798,
  /num/direct15k_seed1@test300_seed0/hi/.initial=0.827,
  /num/direct15k_seed1@test300_seed0/dir/.initial=0.610,
  /num/direct15k_seed1@test300_seed0/parse/.initial=1.000,
  /num/direct15k_seed1@test300_seed0/n/.initial=300,
  /num/direct15k_seed1@ep1@val300_seed0/miou/.initial=0.775,
  /num/direct15k_seed1@ep1@val300_seed0/lo/.initial=0.758,
  /num/direct15k_seed1@ep1@val300_seed0/hi/.initial=0.793,
  /num/direct15k_seed1@ep1@val300_seed0/dir/.initial=0.533,
  /num/direct15k_seed1@ep1@val300_seed0/parse/.initial=1.000,
  /num/direct15k_seed1@ep1@val300_seed0/n/.initial=300,
  /num/direct15k_seed1@ep10@val300_seed0/miou/.initial=0.816,
  /num/direct15k_seed1@ep10@val300_seed0/lo/.initial=0.803,
  /num/direct15k_seed1@ep10@val300_seed0/hi/.initial=0.831,
  /num/direct15k_seed1@ep10@val300_seed0/dir/.initial=0.623,
  /num/direct15k_seed1@ep10@val300_seed0/parse/.initial=1.000,
  /num/direct15k_seed1@ep10@val300_seed0/n/.initial=300,
  /num/direct15k_seed1@ep2@val300_seed0/miou/.initial=0.796,
  /num/direct15k_seed1@ep2@val300_seed0/lo/.initial=0.780,
  /num/direct15k_seed1@ep2@val300_seed0/hi/.initial=0.812,
  /num/direct15k_seed1@ep2@val300_seed0/dir/.initial=0.640,
  /num/direct15k_seed1@ep2@val300_seed0/parse/.initial=1.000,
  /num/direct15k_seed1@ep2@val300_seed0/n/.initial=300,
  /num/direct15k_seed1@ep3@val300_seed0/miou/.initial=0.803,
  /num/direct15k_seed1@ep3@val300_seed0/lo/.initial=0.787,
  /num/direct15k_seed1@ep3@val300_seed0/hi/.initial=0.820,
  /num/direct15k_seed1@ep3@val300_seed0/dir/.initial=0.617,
  /num/direct15k_seed1@ep3@val300_seed0/parse/.initial=1.000,
  /num/direct15k_seed1@ep3@val300_seed0/n/.initial=300,
  /num/direct15k_seed1@ep4@val300_seed0/miou/.initial=0.805,
  /num/direct15k_seed1@ep4@val300_seed0/lo/.initial=0.789,
  /num/direct15k_seed1@ep4@val300_seed0/hi/.initial=0.821,
  /num/direct15k_seed1@ep4@val300_seed0/dir/.initial=0.620,
  /num/direct15k_seed1@ep4@val300_seed0/parse/.initial=1.000,
  /num/direct15k_seed1@ep4@val300_seed0/n/.initial=300,
  /num/direct15k_seed1@ep5@val300_seed0/miou/.initial=0.806,
  /num/direct15k_seed1@ep5@val300_seed0/lo/.initial=0.792,
  /num/direct15k_seed1@ep5@val300_seed0/hi/.initial=0.820,
  /num/direct15k_seed1@ep5@val300_seed0/dir/.initial=0.610,
  /num/direct15k_seed1@ep5@val300_seed0/parse/.initial=1.000,
  /num/direct15k_seed1@ep5@val300_seed0/n/.initial=300,
  /num/direct15k_seed1@ep6@val300_seed0/miou/.initial=0.814,
  /num/direct15k_seed1@ep6@val300_seed0/lo/.initial=0.800,
  /num/direct15k_seed1@ep6@val300_seed0/hi/.initial=0.827,
  /num/direct15k_seed1@ep6@val300_seed0/dir/.initial=0.623,
  /num/direct15k_seed1@ep6@val300_seed0/parse/.initial=1.000,
  /num/direct15k_seed1@ep6@val300_seed0/n/.initial=300,
  /num/direct15k_seed1@ep7@val300_seed0/miou/.initial=0.813,
  /num/direct15k_seed1@ep7@val300_seed0/lo/.initial=0.799,
  /num/direct15k_seed1@ep7@val300_seed0/hi/.initial=0.827,
  /num/direct15k_seed1@ep7@val300_seed0/dir/.initial=0.623,
  /num/direct15k_seed1@ep7@val300_seed0/parse/.initial=1.000,
  /num/direct15k_seed1@ep7@val300_seed0/n/.initial=300,
  /num/direct15k_seed1@ep8@val300_seed0/miou/.initial=0.817,
  /num/direct15k_seed1@ep8@val300_seed0/lo/.initial=0.804,
  /num/direct15k_seed1@ep8@val300_seed0/hi/.initial=0.831,
  /num/direct15k_seed1@ep8@val300_seed0/dir/.initial=0.613,
  /num/direct15k_seed1@ep8@val300_seed0/parse/.initial=1.000,
  /num/direct15k_seed1@ep8@val300_seed0/n/.initial=300,
  /num/direct15k_seed1@ep9@val300_seed0/miou/.initial=0.815,
  /num/direct15k_seed1@ep9@val300_seed0/lo/.initial=0.802,
  /num/direct15k_seed1@ep9@val300_seed0/hi/.initial=0.830,
  /num/direct15k_seed1@ep9@val300_seed0/dir/.initial=0.613,
  /num/direct15k_seed1@ep9@val300_seed0/parse/.initial=1.000,
  /num/direct15k_seed1@ep9@val300_seed0/n/.initial=300,
  /num/direct15k_seed2@test300_seed0/miou/.initial=0.800,
  /num/direct15k_seed2@test300_seed0/lo/.initial=0.786,
  /num/direct15k_seed2@test300_seed0/hi/.initial=0.816,
  /num/direct15k_seed2@test300_seed0/dir/.initial=0.573,
  /num/direct15k_seed2@test300_seed0/parse/.initial=1.000,
  /num/direct15k_seed2@test300_seed0/n/.initial=300,
  /num/direct15k_seed2@ep1@val300_seed0/miou/.initial=0.769,
  /num/direct15k_seed2@ep1@val300_seed0/lo/.initial=0.753,
  /num/direct15k_seed2@ep1@val300_seed0/hi/.initial=0.785,
  /num/direct15k_seed2@ep1@val300_seed0/dir/.initial=0.527,
  /num/direct15k_seed2@ep1@val300_seed0/parse/.initial=1.000,
  /num/direct15k_seed2@ep1@val300_seed0/n/.initial=300,
  /num/direct15k_seed2@ep10@val300_seed0/miou/.initial=0.812,
  /num/direct15k_seed2@ep10@val300_seed0/lo/.initial=0.797,
  /num/direct15k_seed2@ep10@val300_seed0/hi/.initial=0.826,
  /num/direct15k_seed2@ep10@val300_seed0/dir/.initial=0.633,
  /num/direct15k_seed2@ep10@val300_seed0/parse/.initial=1.000,
  /num/direct15k_seed2@ep10@val300_seed0/n/.initial=300,
  /num/direct15k_seed2@ep2@val300_seed0/miou/.initial=0.787,
  /num/direct15k_seed2@ep2@val300_seed0/lo/.initial=0.772,
  /num/direct15k_seed2@ep2@val300_seed0/hi/.initial=0.803,
  /num/direct15k_seed2@ep2@val300_seed0/dir/.initial=0.577,
  /num/direct15k_seed2@ep2@val300_seed0/parse/.initial=1.000,
  /num/direct15k_seed2@ep2@val300_seed0/n/.initial=300,
  /num/direct15k_seed2@ep3@val300_seed0/miou/.initial=0.791,
  /num/direct15k_seed2@ep3@val300_seed0/lo/.initial=0.776,
  /num/direct15k_seed2@ep3@val300_seed0/hi/.initial=0.807,
  /num/direct15k_seed2@ep3@val300_seed0/dir/.initial=0.607,
  /num/direct15k_seed2@ep3@val300_seed0/parse/.initial=1.000,
  /num/direct15k_seed2@ep3@val300_seed0/n/.initial=300,
  /num/direct15k_seed2@ep4@val300_seed0/miou/.initial=0.810,
  /num/direct15k_seed2@ep4@val300_seed0/lo/.initial=0.795,
  /num/direct15k_seed2@ep4@val300_seed0/hi/.initial=0.824,
  /num/direct15k_seed2@ep4@val300_seed0/dir/.initial=0.627,
  /num/direct15k_seed2@ep4@val300_seed0/parse/.initial=1.000,
  /num/direct15k_seed2@ep4@val300_seed0/n/.initial=300,
  /num/direct15k_seed2@ep5@val300_seed0/miou/.initial=0.811,
  /num/direct15k_seed2@ep5@val300_seed0/lo/.initial=0.797,
  /num/direct15k_seed2@ep5@val300_seed0/hi/.initial=0.826,
  /num/direct15k_seed2@ep5@val300_seed0/dir/.initial=0.650,
  /num/direct15k_seed2@ep5@val300_seed0/parse/.initial=1.000,
  /num/direct15k_seed2@ep5@val300_seed0/n/.initial=300,
  /num/direct15k_seed2@ep6@val300_seed0/miou/.initial=0.814,
  /num/direct15k_seed2@ep6@val300_seed0/lo/.initial=0.800,
  /num/direct15k_seed2@ep6@val300_seed0/hi/.initial=0.828,
  /num/direct15k_seed2@ep6@val300_seed0/dir/.initial=0.640,
  /num/direct15k_seed2@ep6@val300_seed0/parse/.initial=1.000,
  /num/direct15k_seed2@ep6@val300_seed0/n/.initial=300,
  /num/direct15k_seed2@ep7@val300_seed0/miou/.initial=0.815,
  /num/direct15k_seed2@ep7@val300_seed0/lo/.initial=0.802,
  /num/direct15k_seed2@ep7@val300_seed0/hi/.initial=0.829,
  /num/direct15k_seed2@ep7@val300_seed0/dir/.initial=0.637,
  /num/direct15k_seed2@ep7@val300_seed0/parse/.initial=1.000,
  /num/direct15k_seed2@ep7@val300_seed0/n/.initial=300,
  /num/direct15k_seed2@ep8@val300_seed0/miou/.initial=0.813,
  /num/direct15k_seed2@ep8@val300_seed0/lo/.initial=0.800,
  /num/direct15k_seed2@ep8@val300_seed0/hi/.initial=0.827,
  /num/direct15k_seed2@ep8@val300_seed0/dir/.initial=0.610,
  /num/direct15k_seed2@ep8@val300_seed0/parse/.initial=1.000,
  /num/direct15k_seed2@ep8@val300_seed0/n/.initial=300,
  /num/direct15k_seed2@ep9@val300_seed0/miou/.initial=0.812,
  /num/direct15k_seed2@ep9@val300_seed0/lo/.initial=0.798,
  /num/direct15k_seed2@ep9@val300_seed0/hi/.initial=0.826,
  /num/direct15k_seed2@ep9@val300_seed0/dir/.initial=0.623,
  /num/direct15k_seed2@ep9@val300_seed0/parse/.initial=1.000,
  /num/direct15k_seed2@ep9@val300_seed0/n/.initial=300,
  /num/direct15k_symarea@test300_seed0/miou/.initial=0.709,
  /num/direct15k_symarea@test300_seed0/lo/.initial=0.697,
  /num/direct15k_symarea@test300_seed0/hi/.initial=0.721,
  /num/direct15k_symarea@test300_seed0/dir/.initial=0.237,
  /num/direct15k_symarea@test300_seed0/parse/.initial=1.000,
  /num/direct15k_symarea@test300_seed0/n/.initial=300,
  /num/direct15k_symarea@test300_seed0/mae/.initial=12.45,
  /num/direct15k_symarea@test300_seed0/f1/.initial=0.656,
  /num/direct15k_symarea@test300_seed0/f1_5/.initial=0.771,
  /num/direct15k_symarea@test300_seed0/f1_15/.initial=0.558,
  /num/direct15k_symarea@test300_seed0/f1_20/.initial=0.214,
  /num/direct15k_symarea@test300_seed0/p10/.initial=0.530,
  /num/direct15k_symarea@test300_seed0/r10/.initial=0.860,
  /num/direct21k@test300_seed0/miou/.initial=0.821,
  /num/direct21k@test300_seed0/lo/.initial=0.806,
  /num/direct21k@test300_seed0/hi/.initial=0.835,
  /num/direct21k@test300_seed0/dir/.initial=0.610,
  /num/direct21k@test300_seed0/parse/.initial=1.000,
  /num/direct21k@test300_seed0/n/.initial=300,
  /num/direct21k@test300_seed0/mae/.initial=7.21,
  /num/direct21k@test300_seed0/f1/.initial=0.766,
  /num/direct21k@test300_seed0/f1_5/.initial=0.810,
  /num/direct21k@test300_seed0/f1_15/.initial=0.773,
  /num/direct21k@test300_seed0/f1_20/.initial=0.743,
  /num/direct21k@test300_seed0/p10/.initial=0.750,
  /num/direct21k@test300_seed0/r10/.initial=0.783,
  /num/direct21k@test_full/miou/.initial=0.812,
  /num/direct21k@test_full/lo/.initial=0.808,
  /num/direct21k@test_full/hi/.initial=0.816,
  /num/direct21k@test_full/dir/.initial=0.598,
  /num/direct21k@test_full/parse/.initial=1.000,
  /num/direct21k@test_full/n/.initial=3807,
  /num/direct21k@ep1@val300_seed0/miou/.initial=0.793,
  /num/direct21k@ep1@val300_seed0/lo/.initial=0.777,
  /num/direct21k@ep1@val300_seed0/hi/.initial=0.809,
  /num/direct21k@ep1@val300_seed0/dir/.initial=0.600,
  /num/direct21k@ep1@val300_seed0/parse/.initial=1.000,
  /num/direct21k@ep1@val300_seed0/n/.initial=300,
  /num/direct21k@ep10@val300_seed0/miou/.initial=0.821,
  /num/direct21k@ep10@val300_seed0/lo/.initial=0.807,
  /num/direct21k@ep10@val300_seed0/hi/.initial=0.835,
  /num/direct21k@ep10@val300_seed0/dir/.initial=0.637,
  /num/direct21k@ep10@val300_seed0/parse/.initial=1.000,
  /num/direct21k@ep10@val300_seed0/n/.initial=300,
  /num/direct21k@ep2@val300_seed0/miou/.initial=0.800,
  /num/direct21k@ep2@val300_seed0/lo/.initial=0.783,
  /num/direct21k@ep2@val300_seed0/hi/.initial=0.816,
  /num/direct21k@ep2@val300_seed0/dir/.initial=0.600,
  /num/direct21k@ep2@val300_seed0/parse/.initial=1.000,
  /num/direct21k@ep2@val300_seed0/n/.initial=300,
  /num/direct21k@ep3@val300_seed0/miou/.initial=0.793,
  /num/direct21k@ep3@val300_seed0/lo/.initial=0.776,
  /num/direct21k@ep3@val300_seed0/hi/.initial=0.810,
  /num/direct21k@ep3@val300_seed0/dir/.initial=0.600,
  /num/direct21k@ep3@val300_seed0/parse/.initial=1.000,
  /num/direct21k@ep3@val300_seed0/n/.initial=300,
  /num/direct21k@ep4@val300_seed0/miou/.initial=0.804,
  /num/direct21k@ep4@val300_seed0/lo/.initial=0.789,
  /num/direct21k@ep4@val300_seed0/hi/.initial=0.820,
  /num/direct21k@ep4@val300_seed0/dir/.initial=0.613,
  /num/direct21k@ep4@val300_seed0/parse/.initial=1.000,
  /num/direct21k@ep4@val300_seed0/n/.initial=300,
  /num/direct21k@ep5@val300_seed0/miou/.initial=0.825,
  /num/direct21k@ep5@val300_seed0/lo/.initial=0.810,
  /num/direct21k@ep5@val300_seed0/hi/.initial=0.839,
  /num/direct21k@ep5@val300_seed0/dir/.initial=0.647,
  /num/direct21k@ep5@val300_seed0/parse/.initial=1.000,
  /num/direct21k@ep5@val300_seed0/n/.initial=300,
  /num/direct21k@ep6@val300_seed0/miou/.initial=0.818,
  /num/direct21k@ep6@val300_seed0/lo/.initial=0.805,
  /num/direct21k@ep6@val300_seed0/hi/.initial=0.832,
  /num/direct21k@ep6@val300_seed0/dir/.initial=0.643,
  /num/direct21k@ep6@val300_seed0/parse/.initial=1.000,
  /num/direct21k@ep6@val300_seed0/n/.initial=300,
  /num/direct21k@ep7@val300_seed0/miou/.initial=0.825,
  /num/direct21k@ep7@val300_seed0/lo/.initial=0.811,
  /num/direct21k@ep7@val300_seed0/hi/.initial=0.838,
  /num/direct21k@ep7@val300_seed0/dir/.initial=0.640,
  /num/direct21k@ep7@val300_seed0/parse/.initial=1.000,
  /num/direct21k@ep7@val300_seed0/n/.initial=300,
  /num/direct21k@ep8@val300_seed0/miou/.initial=0.825,
  /num/direct21k@ep8@val300_seed0/lo/.initial=0.811,
  /num/direct21k@ep8@val300_seed0/hi/.initial=0.839,
  /num/direct21k@ep8@val300_seed0/dir/.initial=0.650,
  /num/direct21k@ep8@val300_seed0/parse/.initial=1.000,
  /num/direct21k@ep8@val300_seed0/n/.initial=300,
  /num/direct21k@ep9@val300_seed0/miou/.initial=0.823,
  /num/direct21k@ep9@val300_seed0/lo/.initial=0.809,
  /num/direct21k@ep9@val300_seed0/hi/.initial=0.836,
  /num/direct21k@ep9@val300_seed0/dir/.initial=0.637,
  /num/direct21k@ep9@val300_seed0/parse/.initial=1.000,
  /num/direct21k@ep9@val300_seed0/n/.initial=300,
  /num/direct21k_bin30@test300_seed0/miou/.initial=0.750,
  /num/direct21k_bin30@test300_seed0/lo/.initial=0.736,
  /num/direct21k_bin30@test300_seed0/hi/.initial=0.763,
  /num/direct21k_bin30@test300_seed0/dir/.initial=0.543,
  /num/direct21k_bin30@test300_seed0/parse/.initial=1.000,
  /num/direct21k_bin30@test300_seed0/n/.initial=300,
  /num/direct21k_bin30@test300_seed0/mae/.initial=10.00,
  /num/direct21k_bin30@test300_seed0/f1/.initial=0.751,
  /num/direct21k_bin30@test300_seed0/f1_5/.initial=0.700,
  /num/direct21k_bin30@test300_seed0/f1_15/.initial=0.767,
  /num/direct21k_bin30@test300_seed0/f1_20/.initial=0.719,
  /num/direct21k_bin30@test300_seed0/p10/.initial=0.815,
  /num/direct21k_bin30@test300_seed0/r10/.initial=0.697,
  /num/direct21k_bin30@test_full/miou/.initial=0.753,
  /num/direct21k_bin30@test_full/lo/.initial=0.749,
  /num/direct21k_bin30@test_full/hi/.initial=0.757,
  /num/direct21k_bin30@test_full/dir/.initial=0.560,
  /num/direct21k_bin30@test_full/parse/.initial=1.000,
  /num/direct21k_bin30@test_full/n/.initial=3807,
  /num/direct21k_q10@test300_seed0/miou/.initial=0.804,
  /num/direct21k_q10@test300_seed0/lo/.initial=0.789,
  /num/direct21k_q10@test300_seed0/hi/.initial=0.818,
  /num/direct21k_q10@test300_seed0/dir/.initial=0.583,
  /num/direct21k_q10@test300_seed0/parse/.initial=1.000,
  /num/direct21k_q10@test300_seed0/n/.initial=300,
  /num/direct21k_q10@test300_seed0/mae/.initial=7.90,
  /num/direct21k_q10@test300_seed0/f1/.initial=0.751,
  /num/direct21k_q10@test300_seed0/f1_5/.initial=0.810,
  /num/direct21k_q10@test300_seed0/f1_15/.initial=0.767,
  /num/direct21k_q10@test300_seed0/f1_20/.initial=0.701,
  /num/direct21k_q10@test300_seed0/p10/.initial=0.815,
  /num/direct21k_q10@test300_seed0/r10/.initial=0.697,
  /num/direct21k_q10@test_full/miou/.initial=0.796,
  /num/direct21k_q10@test_full/lo/.initial=0.792,
  /num/direct21k_q10@test_full/hi/.initial=0.799,
  /num/direct21k_q10@test_full/dir/.initial=0.582,
  /num/direct21k_q10@test_full/parse/.initial=1.000,
  /num/direct21k_q10@test_full/n/.initial=3807,
  /num/direct21k_q25@test300_seed0/miou/.initial=0.755,
  /num/direct21k_q25@test300_seed0/lo/.initial=0.742,
  /num/direct21k_q25@test300_seed0/hi/.initial=0.767,
  /num/direct21k_q25@test300_seed0/dir/.initial=0.580,
  /num/direct21k_q25@test300_seed0/parse/.initial=1.000,
  /num/direct21k_q25@test300_seed0/n/.initial=300,
  /num/direct21k_q25@test300_seed0/mae/.initial=9.78,
  /num/direct21k_q25@test300_seed0/f1/.initial=0.764,
  /num/direct21k_q25@test300_seed0/f1_5/.initial=0.726,
  /num/direct21k_q25@test300_seed0/f1_15/.initial=0.762,
  /num/direct21k_q25@test300_seed0/f1_20/.initial=0.708,
  /num/direct21k_q25@test300_seed0/p10/.initial=0.800,
  /num/direct21k_q25@test300_seed0/r10/.initial=0.731,
  /num/direct21k_q25@test_full/miou/.initial=0.752,
  /num/direct21k_q25@test_full/lo/.initial=0.749,
  /num/direct21k_q25@test_full/hi/.initial=0.756,
  /num/direct21k_q25@test_full/dir/.initial=0.561,
  /num/direct21k_q25@test_full/parse/.initial=1.000,
  /num/direct21k_q25@test_full/n/.initial=3807,
  /num/direct21k_symarea@test300_seed0/miou/.initial=0.715,
  /num/direct21k_symarea@test300_seed0/lo/.initial=0.703,
  /num/direct21k_symarea@test300_seed0/hi/.initial=0.728,
  /num/direct21k_symarea@test300_seed0/dir/.initial=0.237,
  /num/direct21k_symarea@test300_seed0/parse/.initial=1.000,
  /num/direct21k_symarea@test300_seed0/n/.initial=300,
  /num/direct21k_symarea@test300_seed0/mae/.initial=12.20,
  /num/direct21k_symarea@test300_seed0/f1/.initial=0.670,
  /num/direct21k_symarea@test300_seed0/f1_5/.initial=0.771,
  /num/direct21k_symarea@test300_seed0/f1_15/.initial=0.582,
  /num/direct21k_symarea@test300_seed0/f1_20/.initial=0.226,
  /num/direct21k_symarea@test300_seed0/p10/.initial=0.548,
  /num/direct21k_symarea@test300_seed0/r10/.initial=0.860,
  /num/direct21k_symarea@test_full/miou/.initial=0.712,
  /num/direct21k_symarea@test_full/lo/.initial=0.709,
  /num/direct21k_symarea@test_full/hi/.initial=0.715,
  /num/direct21k_symarea@test_full/dir/.initial=0.256,
  /num/direct21k_symarea@test_full/parse/.initial=1.000,
  /num/direct21k_symarea@test_full/n/.initial=3807,
  /num/direct21k_symmax@test_full/miou/.initial=0.651,
  /num/direct21k_symmax@test_full/lo/.initial=0.647,
  /num/direct21k_symmax@test_full/hi/.initial=0.655,
  /num/direct21k_symmax@test_full/dir/.initial=0.256,
  /num/direct21k_symmax@test_full/parse/.initial=1.000,
  /num/direct21k_symmax@test_full/n/.initial=3807,
  /num/direct21k_symmean@test300_seed0/miou/.initial=0.715,
  /num/direct21k_symmean@test300_seed0/lo/.initial=0.702,
  /num/direct21k_symmean@test300_seed0/hi/.initial=0.728,
  /num/direct21k_symmean@test300_seed0/dir/.initial=0.237,
  /num/direct21k_symmean@test300_seed0/parse/.initial=1.000,
  /num/direct21k_symmean@test300_seed0/n/.initial=300,
  /num/direct21k_symmean@test300_seed0/mae/.initial=12.26,
  /num/direct21k_symmean@test300_seed0/f1/.initial=0.670,
  /num/direct21k_symmean@test300_seed0/f1_5/.initial=0.771,
  /num/direct21k_symmean@test300_seed0/f1_15/.initial=0.582,
  /num/direct21k_symmean@test300_seed0/f1_20/.initial=0.252,
  /num/direct21k_symmean@test300_seed0/p10/.initial=0.548,
  /num/direct21k_symmean@test300_seed0/r10/.initial=0.860,
  /num/direct21k_symmean@test_full/miou/.initial=0.711,
  /num/direct21k_symmean@test_full/lo/.initial=0.708,
  /num/direct21k_symmean@test_full/hi/.initial=0.715,
  /num/direct21k_symmean@test_full/dir/.initial=0.256,
  /num/direct21k_symmean@test_full/parse/.initial=1.000,
  /num/direct21k_symmean@test_full/n/.initial=3807,
  /num/gaic@test300_seed0/miou/.initial=0.401,
  /num/gaic@test300_seed0/lo/.initial=0.390,
  /num/gaic@test300_seed0/hi/.initial=0.412,
  /num/gaic@test300_seed0/dir/.initial=0.360,
  /num/gaic@test300_seed0/parse/.initial=1.000,
  /num/gaic@test300_seed0/n/.initial=300,
  /num/gaic@test300_seed0/mae/.initial=15.71,
  /num/gaic@test300_seed0/f1/.initial=0.000,
  /num/gaic@test300_seed0/f1_5/.initial=0.000,
  /num/gaic@test300_seed0/f1_15/.initial=0.000,
  /num/gaic@test300_seed0/f1_20/.initial=0.000,
  /num/gaic@test300_seed0/p10/.initial=0.000,
  /num/gaic@test300_seed0/r10/.initial=0.000,
  /num/gaic@test_full/miou/.initial=0.397,
  /num/gaic@test_full/lo/.initial=0.394,
  /num/gaic@test_full/hi/.initial=0.400,
  /num/gaic@test_full/dir/.initial=0.336,
  /num/gaic@test_full/parse/.initial=1.000,
  /num/gaic@test_full/n/.initial=3807,
  /num/gpt-4o_icl9@test300_seed0/miou/.initial=0.676,
  /num/gpt-4o_icl9@test300_seed0/lo/.initial=0.663,
  /num/gpt-4o_icl9@test300_seed0/hi/.initial=0.689,
  /num/gpt-4o_icl9@test300_seed0/dir/.initial=0.233,
  /num/gpt-4o_icl9@test300_seed0/parse/.initial=1.000,
  /num/gpt-4o_icl9@test300_seed0/n/.initial=300,
  /num/gpt-4o_icl9@test300_seed0/mae/.initial=13.72,
  /num/gpt-4o_icl9@test300_seed0/f1/.initial=0.474,
  /num/gpt-4o_icl9@test300_seed0/f1_5/.initial=0.656,
  /num/gpt-4o_icl9@test300_seed0/f1_15/.initial=0.322,
  /num/gpt-4o_icl9@test300_seed0/f1_20/.initial=0.221,
  /num/gpt-4o_icl9@test300_seed0/p10/.initial=0.510,
  /num/gpt-4o_icl9@test300_seed0/r10/.initial=0.443,
  /num/gpt-4o_zs@test300_seed0/miou/.initial=0.672,
  /num/gpt-4o_zs@test300_seed0/lo/.initial=0.660,
  /num/gpt-4o_zs@test300_seed0/hi/.initial=0.684,
  /num/gpt-4o_zs@test300_seed0/dir/.initial=0.280,
  /num/gpt-4o_zs@test300_seed0/parse/.initial=1.000,
  /num/gpt-4o_zs@test300_seed0/n/.initial=300,
  /num/gpt-4o_zs@test300_seed0/mae/.initial=14.54,
  /num/gpt-4o_zs@test300_seed0/f1/.initial=0.568,
  /num/gpt-4o_zs@test300_seed0/f1_5/.initial=0.753,
  /num/gpt-4o_zs@test300_seed0/f1_15/.initial=0.482,
  /num/gpt-4o_zs@test300_seed0/f1_20/.initial=0.310,
  /num/gpt-4o_zs@test300_seed0/p10/.initial=0.526,
  /num/gpt-4o_zs@test300_seed0/r10/.initial=0.616,
  /num/gpt-5.5_icl9@test300_seed0/miou/.initial=0.713,
  /num/gpt-5.5_icl9@test300_seed0/lo/.initial=0.701,
  /num/gpt-5.5_icl9@test300_seed0/hi/.initial=0.725,
  /num/gpt-5.5_icl9@test300_seed0/dir/.initial=0.443,
  /num/gpt-5.5_icl9@test300_seed0/parse/.initial=1.000,
  /num/gpt-5.5_icl9@test300_seed0/n/.initial=300,
  /num/gpt-5.5_icl9@test300_seed0/mae/.initial=11.91,
  /num/gpt-5.5_icl9@test300_seed0/f1/.initial=0.644,
  /num/gpt-5.5_icl9@test300_seed0/f1_5/.initial=0.737,
  /num/gpt-5.5_icl9@test300_seed0/f1_15/.initial=0.534,
  /num/gpt-5.5_icl9@test300_seed0/f1_20/.initial=0.463,
  /num/gpt-5.5_icl9@test300_seed0/p10/.initial=0.660,
  /num/gpt-5.5_icl9@test300_seed0/r10/.initial=0.628,
  /num/gpt-5.5_zs@test300_seed0/miou/.initial=0.700,
  /num/gpt-5.5_zs@test300_seed0/lo/.initial=0.688,
  /num/gpt-5.5_zs@test300_seed0/hi/.initial=0.711,
  /num/gpt-5.5_zs@test300_seed0/dir/.initial=0.397,
  /num/gpt-5.5_zs@test300_seed0/parse/.initial=1.000,
  /num/gpt-5.5_zs@test300_seed0/n/.initial=300,
  /num/gpt-5.5_zs@test300_seed0/mae/.initial=12.51,
  /num/gpt-5.5_zs@test300_seed0/f1/.initial=0.653,
  /num/gpt-5.5_zs@test300_seed0/f1_5/.initial=0.726,
  /num/gpt-5.5_zs@test300_seed0/f1_15/.initial=0.531,
  /num/gpt-5.5_zs@test300_seed0/f1_20/.initial=0.435,
  /num/gpt-5.5_zs@test300_seed0/p10/.initial=0.646,
  /num/gpt-5.5_zs@test300_seed0/r10/.initial=0.661,
  /num/gpt-6-astra_icl9@test300_seed0/miou/.initial=0.734,
  /num/gpt-6-astra_icl9@test300_seed0/lo/.initial=0.723,
  /num/gpt-6-astra_icl9@test300_seed0/hi/.initial=0.746,
  /num/gpt-6-astra_icl9@test300_seed0/dir/.initial=0.533,
  /num/gpt-6-astra_icl9@test300_seed0/parse/.initial=1.000,
  /num/gpt-6-astra_icl9@test300_seed0/n/.initial=300,
  /num/gpt-6-astra_icl9@test300_seed0/mae/.initial=10.75,
  /num/gpt-6-astra_icl9@test300_seed0/f1/.initial=0.701,
  /num/gpt-6-astra_icl9@test300_seed0/f1_5/.initial=0.756,
  /num/gpt-6-astra_icl9@test300_seed0/f1_15/.initial=0.630,
  /num/gpt-6-astra_icl9@test300_seed0/f1_20/.initial=0.559,
  /num/gpt-6-astra_icl9@test300_seed0/p10/.initial=0.702,
  /num/gpt-6-astra_icl9@test300_seed0/r10/.initial=0.700,
  /num/gpt-6-astra_zs@test300_seed0/miou/.initial=0.707,
  /num/gpt-6-astra_zs@test300_seed0/lo/.initial=0.694,
  /num/gpt-6-astra_zs@test300_seed0/hi/.initial=0.719,
  /num/gpt-6-astra_zs@test300_seed0/dir/.initial=0.490,
  /num/gpt-6-astra_zs@test300_seed0/parse/.initial=1.000,
  /num/gpt-6-astra_zs@test300_seed0/n/.initial=300,
  /num/gpt-6-astra_zs@test300_seed0/mae/.initial=12.10,
  /num/gpt-6-astra_zs@test300_seed0/f1/.initial=0.663,
  /num/gpt-6-astra_zs@test300_seed0/f1_5/.initial=0.717,
  /num/gpt-6-astra_zs@test300_seed0/f1_15/.initial=0.599,
  /num/gpt-6-astra_zs@test300_seed0/f1_20/.initial=0.515,
  /num/gpt-6-astra_zs@test300_seed0/p10/.initial=0.704,
  /num/gpt-6-astra_zs@test300_seed0/r10/.initial=0.626,
  /num/noop@test300_seed0/miou/.initial=0.593,
  /num/noop@test300_seed0/lo/.initial=0.582,
  /num/noop@test300_seed0/hi/.initial=0.603,
  /num/noop@test300_seed0/dir/.initial=0.237,
  /num/noop@test300_seed0/parse/.initial=1.000,
  /num/noop@test300_seed0/n/.initial=300,
  /num/noop@test300_seed0/mae/.initial=15.71,
  /num/noop@test300_seed0/f1/.initial=0.000,
  /num/noop@test300_seed0/f1_5/.initial=0.000,
  /num/noop@test300_seed0/f1_15/.initial=0.000,
  /num/noop@test300_seed0/f1_20/.initial=0.000,
  /num/noop@test300_seed0/p10/.initial=0.000,
  /num/noop@test300_seed0/r10/.initial=0.000,
  /num/noop@test_full/miou/.initial=0.587,
  /num/noop@test_full/lo/.initial=0.584,
  /num/noop@test_full/hi/.initial=0.589,
  /num/noop@test_full/dir/.initial=0.256,
  /num/noop@test_full/parse/.initial=1.000,
  /num/noop@test_full/n/.initial=3807,
  /num/ours_e36@test300_seed0/miou/.initial=0.782,
  /num/ours_e36@test300_seed0/lo/.initial=0.765,
  /num/ours_e36@test300_seed0/hi/.initial=0.799,
  /num/ours_e36@test300_seed0/dir/.initial=0.553,
  /num/ours_e36@test300_seed0/parse/.initial=1.000,
  /num/ours_e36@test300_seed0/n/.initial=300,
  /num/ours_e36@test300_seed0/mae/.initial=8.95,
  /num/ours_e36@test300_seed0/f1/.initial=0.732,
  /num/ours_e36@test300_seed0/f1_5/.initial=0.749,
  /num/ours_e36@test300_seed0/f1_15/.initial=0.692,
  /num/ours_e36@test300_seed0/f1_20/.initial=0.695,
  /num/ours_e36@test300_seed0/p10/.initial=0.735,
  /num/ours_e36@test300_seed0/r10/.initial=0.730,
  /num/ours_e36@test_full/miou/.initial=0.778,
  /num/ours_e36@test_full/lo/.initial=0.773,
  /num/ours_e36@test_full/hi/.initial=0.782,
  /num/ours_e36@test_full/dir/.initial=0.551,
  /num/ours_e36@test_full/parse/.initial=1.000,
  /num/ours_e36@test_full/n/.initial=3807,
  /num/ours_e36@ep1@val300_seed0/miou/.initial=0.738,
  /num/ours_e36@ep1@val300_seed0/lo/.initial=0.720,
  /num/ours_e36@ep1@val300_seed0/hi/.initial=0.755,
  /num/ours_e36@ep1@val300_seed0/dir/.initial=0.523,
  /num/ours_e36@ep1@val300_seed0/parse/.initial=1.000,
  /num/ours_e36@ep1@val300_seed0/n/.initial=300,
  /num/ours_e36@ep2@val300_seed0/miou/.initial=0.767,
  /num/ours_e36@ep2@val300_seed0/lo/.initial=0.751,
  /num/ours_e36@ep2@val300_seed0/hi/.initial=0.783,
  /num/ours_e36@ep2@val300_seed0/dir/.initial=0.537,
  /num/ours_e36@ep2@val300_seed0/parse/.initial=1.000,
  /num/ours_e36@ep2@val300_seed0/n/.initial=300,
  /num/ours_e36@ep3@val300_seed0/miou/.initial=0.782,
  /num/ours_e36@ep3@val300_seed0/lo/.initial=0.766,
  /num/ours_e36@ep3@val300_seed0/hi/.initial=0.798,
  /num/ours_e36@ep3@val300_seed0/dir/.initial=0.563,
  /num/ours_e36@ep3@val300_seed0/parse/.initial=1.000,
  /num/ours_e36@ep3@val300_seed0/n/.initial=300,
  /num/ours_e36@ep4@val300_seed0/miou/.initial=0.784,
  /num/ours_e36@ep4@val300_seed0/lo/.initial=0.768,
  /num/ours_e36@ep4@val300_seed0/hi/.initial=0.800,
  /num/ours_e36@ep4@val300_seed0/dir/.initial=0.570,
  /num/ours_e36@ep4@val300_seed0/parse/.initial=1.000,
  /num/ours_e36@ep4@val300_seed0/n/.initial=300,
  /num/ours_e36@ep5@val300_seed0/miou/.initial=0.782,
  /num/ours_e36@ep5@val300_seed0/lo/.initial=0.766,
  /num/ours_e36@ep5@val300_seed0/hi/.initial=0.797,
  /num/ours_e36@ep5@val300_seed0/dir/.initial=0.573,
  /num/ours_e36@ep5@val300_seed0/parse/.initial=1.000,
  /num/ours_e36@ep5@val300_seed0/n/.initial=300,
  /num/ours_e36_symarea@test_full/miou/.initial=0.710,
  /num/ours_e36_symarea@test_full/lo/.initial=0.707,
  /num/ours_e36_symarea@test_full/hi/.initial=0.713,
  /num/ours_e36_symarea@test_full/dir/.initial=0.256,
  /num/ours_e36_symarea@test_full/parse/.initial=1.000,
  /num/ours_e36_symarea@test_full/n/.initial=3807,
  /num/ours_e36_symmax@test_full/miou/.initial=0.636,
  /num/ours_e36_symmax@test_full/lo/.initial=0.633,
  /num/ours_e36_symmax@test_full/hi/.initial=0.639,
  /num/ours_e36_symmax@test_full/dir/.initial=0.256,
  /num/ours_e36_symmax@test_full/parse/.initial=1.000,
  /num/ours_e36_symmax@test_full/n/.initial=3807,
  /num/ours_e36_symmean@test300_seed0/miou/.initial=0.713,
  /num/ours_e36_symmean@test300_seed0/lo/.initial=0.700,
  /num/ours_e36_symmean@test300_seed0/hi/.initial=0.726,
  /num/ours_e36_symmean@test300_seed0/dir/.initial=0.237,
  /num/ours_e36_symmean@test300_seed0/parse/.initial=1.000,
  /num/ours_e36_symmean@test300_seed0/n/.initial=300,
  /num/ours_e36_symmean@test300_seed0/mae/.initial=12.31,
  /num/ours_e36_symmean@test300_seed0/f1/.initial=0.668,
  /num/ours_e36_symmean@test300_seed0/f1_5/.initial=0.771,
  /num/ours_e36_symmean@test300_seed0/f1_15/.initial=0.550,
  /num/ours_e36_symmean@test300_seed0/f1_20/.initial=0.108,
  /num/ours_e36_symmean@test300_seed0/p10/.initial=0.549,
  /num/ours_e36_symmean@test300_seed0/r10/.initial=0.853,
  /num/ours_e36_symmean@test_full/miou/.initial=0.710,
  /num/ours_e36_symmean@test_full/lo/.initial=0.707,
  /num/ours_e36_symmean@test_full/hi/.initial=0.713,
  /num/ours_e36_symmean@test_full/dir/.initial=0.256,
  /num/ours_e36_symmean@test_full/parse/.initial=1.000,
  /num/ours_e36_symmean@test_full/n/.initial=3807,
  /num/ours_en@test300_seed0/miou/.initial=0.790,
  /num/ours_en@test300_seed0/lo/.initial=0.773,
  /num/ours_en@test300_seed0/hi/.initial=0.807,
  /num/ours_en@test300_seed0/dir/.initial=0.580,
  /num/ours_en@test300_seed0/parse/.initial=1.000,
  /num/ours_en@test300_seed0/n/.initial=300,
  /num/ours_en@test300_seed0/mae/.initial=8.71,
  /num/ours_en@test300_seed0/f1/.initial=0.743,
  /num/ours_en@test300_seed0/f1_5/.initial=0.754,
  /num/ours_en@test300_seed0/f1_15/.initial=0.704,
  /num/ours_en@test300_seed0/f1_20/.initial=0.740,
  /num/ours_en@test300_seed0/p10/.initial=0.748,
  /num/ours_en@test300_seed0/r10/.initial=0.738,
  /num/ours_en@test_full/miou/.initial=0.784,
  /num/ours_en@test_full/lo/.initial=0.780,
  /num/ours_en@test_full/hi/.initial=0.789,
  /num/ours_en@test_full/dir/.initial=0.574,
  /num/ours_en@test_full/parse/.initial=1.000,
  /num/ours_en@test_full/n/.initial=3807,
  /num/ours_en@ep1@val300_seed0/miou/.initial=0.761,
  /num/ours_en@ep1@val300_seed0/lo/.initial=0.743,
  /num/ours_en@ep1@val300_seed0/hi/.initial=0.778,
  /num/ours_en@ep1@val300_seed0/dir/.initial=0.543,
  /num/ours_en@ep1@val300_seed0/parse/.initial=1.000,
  /num/ours_en@ep1@val300_seed0/n/.initial=300,
  /num/ours_en@ep2@val300_seed0/miou/.initial=0.780,
  /num/ours_en@ep2@val300_seed0/lo/.initial=0.764,
  /num/ours_en@ep2@val300_seed0/hi/.initial=0.797,
  /num/ours_en@ep2@val300_seed0/dir/.initial=0.547,
  /num/ours_en@ep2@val300_seed0/parse/.initial=1.000,
  /num/ours_en@ep2@val300_seed0/n/.initial=300,
  /num/ours_en@ep3@val300_seed0/miou/.initial=0.802,
  /num/ours_en@ep3@val300_seed0/lo/.initial=0.786,
  /num/ours_en@ep3@val300_seed0/hi/.initial=0.818,
  /num/ours_en@ep3@val300_seed0/dir/.initial=0.613,
  /num/ours_en@ep3@val300_seed0/parse/.initial=1.000,
  /num/ours_en@ep3@val300_seed0/n/.initial=300,
  /num/ours_en@ep4@val300_seed0/miou/.initial=0.794,
  /num/ours_en@ep4@val300_seed0/lo/.initial=0.778,
  /num/ours_en@ep4@val300_seed0/hi/.initial=0.809,
  /num/ours_en@ep4@val300_seed0/dir/.initial=0.633,
  /num/ours_en@ep4@val300_seed0/parse/.initial=1.000,
  /num/ours_en@ep4@val300_seed0/n/.initial=300,
  /num/ours_en@ep5@val300_seed0/miou/.initial=0.796,
  /num/ours_en@ep5@val300_seed0/lo/.initial=0.780,
  /num/ours_en@ep5@val300_seed0/hi/.initial=0.810,
  /num/ours_en@ep5@val300_seed0/dir/.initial=0.597,
  /num/ours_en@ep5@val300_seed0/parse/.initial=1.000,
  /num/ours_en@ep5@val300_seed0/n/.initial=300,
  /num/ours_en_bin30@test300_seed0/miou/.initial=0.733,
  /num/ours_en_bin30@test300_seed0/lo/.initial=0.717,
  /num/ours_en_bin30@test300_seed0/hi/.initial=0.748,
  /num/ours_en_bin30@test300_seed0/dir/.initial=0.570,
  /num/ours_en_bin30@test300_seed0/parse/.initial=1.000,
  /num/ours_en_bin30@test300_seed0/n/.initial=300,
  /num/ours_en_bin30@test300_seed0/mae/.initial=11.06,
  /num/ours_en_bin30@test300_seed0/f1/.initial=0.739,
  /num/ours_en_bin30@test300_seed0/f1_5/.initial=0.736,
  /num/ours_en_bin30@test300_seed0/f1_15/.initial=0.696,
  /num/ours_en_bin30@test300_seed0/f1_20/.initial=0.659,
  /num/ours_en_bin30@test300_seed0/p10/.initial=0.748,
  /num/ours_en_bin30@test300_seed0/r10/.initial=0.730,
  /num/ours_en_bin30@test_full/miou/.initial=0.739,
  /num/ours_en_bin30@test_full/lo/.initial=0.734,
  /num/ours_en_bin30@test_full/hi/.initial=0.743,
  /num/ours_en_bin30@test_full/dir/.initial=0.576,
  /num/ours_en_bin30@test_full/parse/.initial=1.000,
  /num/ours_en_bin30@test_full/n/.initial=3807,
  /num/ours_en_prejudge@test300_seed0/miou/.initial=0.785,
  /num/ours_en_prejudge@test300_seed0/lo/.initial=0.769,
  /num/ours_en_prejudge@test300_seed0/hi/.initial=0.801,
  /num/ours_en_prejudge@test300_seed0/dir/.initial=0.547,
  /num/ours_en_prejudge@test300_seed0/parse/.initial=1.000,
  /num/ours_en_prejudge@test300_seed0/n/.initial=300,
  /num/ours_en_prejudge@test300_seed0/mae/.initial=8.89,
  /num/ours_en_prejudge@test300_seed0/f1/.initial=0.735,
  /num/ours_en_prejudge@test300_seed0/f1_5/.initial=0.754,
  /num/ours_en_prejudge@test300_seed0/f1_15/.initial=0.712,
  /num/ours_en_prejudge@test300_seed0/f1_20/.initial=0.704,
  /num/ours_en_prejudge@test300_seed0/p10/.initial=0.740,
  /num/ours_en_prejudge@test300_seed0/r10/.initial=0.730,
  /num/ours_en_prejudge@ep1@val300_seed0/miou/.initial=0.779,
  /num/ours_en_prejudge@ep1@val300_seed0/lo/.initial=0.762,
  /num/ours_en_prejudge@ep1@val300_seed0/hi/.initial=0.795,
  /num/ours_en_prejudge@ep1@val300_seed0/dir/.initial=0.580,
  /num/ours_en_prejudge@ep1@val300_seed0/parse/.initial=1.000,
  /num/ours_en_prejudge@ep1@val300_seed0/n/.initial=300,
  /num/ours_en_prejudge@ep2@val300_seed0/miou/.initial=0.794,
  /num/ours_en_prejudge@ep2@val300_seed0/lo/.initial=0.777,
  /num/ours_en_prejudge@ep2@val300_seed0/hi/.initial=0.811,
  /num/ours_en_prejudge@ep2@val300_seed0/dir/.initial=0.610,
  /num/ours_en_prejudge@ep2@val300_seed0/parse/.initial=1.000,
  /num/ours_en_prejudge@ep2@val300_seed0/n/.initial=300,
  /num/ours_en_prejudge@ep3@val300_seed0/miou/.initial=0.791,
  /num/ours_en_prejudge@ep3@val300_seed0/lo/.initial=0.773,
  /num/ours_en_prejudge@ep3@val300_seed0/hi/.initial=0.808,
  /num/ours_en_prejudge@ep3@val300_seed0/dir/.initial=0.600,
  /num/ours_en_prejudge@ep3@val300_seed0/parse/.initial=1.000,
  /num/ours_en_prejudge@ep3@val300_seed0/n/.initial=300,
  /num/ours_en_prejudge@ep4@val300_seed0/miou/.initial=0.800,
  /num/ours_en_prejudge@ep4@val300_seed0/lo/.initial=0.782,
  /num/ours_en_prejudge@ep4@val300_seed0/hi/.initial=0.816,
  /num/ours_en_prejudge@ep4@val300_seed0/dir/.initial=0.627,
  /num/ours_en_prejudge@ep4@val300_seed0/parse/.initial=1.000,
  /num/ours_en_prejudge@ep4@val300_seed0/n/.initial=300,
  /num/ours_en_prejudge@ep5@val300_seed0/miou/.initial=0.804,
  /num/ours_en_prejudge@ep5@val300_seed0/lo/.initial=0.787,
  /num/ours_en_prejudge@ep5@val300_seed0/hi/.initial=0.820,
  /num/ours_en_prejudge@ep5@val300_seed0/dir/.initial=0.630,
  /num/ours_en_prejudge@ep5@val300_seed0/parse/.initial=1.000,
  /num/ours_en_prejudge@ep5@val300_seed0/n/.initial=300,
  /num/ours_en_q25@test300_seed0/miou/.initial=0.729,
  /num/ours_en_q25@test300_seed0/lo/.initial=0.714,
  /num/ours_en_q25@test300_seed0/hi/.initial=0.744,
  /num/ours_en_q25@test300_seed0/dir/.initial=0.577,
  /num/ours_en_q25@test300_seed0/parse/.initial=1.000,
  /num/ours_en_q25@test300_seed0/n/.initial=300,
  /num/ours_en_q25@test300_seed0/mae/.initial=11.42,
  /num/ours_en_q25@test300_seed0/f1/.initial=0.741,
  /num/ours_en_q25@test300_seed0/f1_5/.initial=0.738,
  /num/ours_en_q25@test300_seed0/f1_15/.initial=0.697,
  /num/ours_en_q25@test300_seed0/f1_20/.initial=0.658,
  /num/ours_en_q25@test300_seed0/p10/.initial=0.749,
  /num/ours_en_q25@test300_seed0/r10/.initial=0.733,
  /num/ours_en_q25@test_full/miou/.initial=0.724,
  /num/ours_en_q25@test_full/lo/.initial=0.720,
  /num/ours_en_q25@test_full/hi/.initial=0.728,
  /num/ours_en_q25@test_full/dir/.initial=0.576,
  /num/ours_en_q25@test_full/parse/.initial=1.000,
  /num/ours_en_q25@test_full/n/.initial=3807,
  /num/ours_en_symarea@test300_seed0/miou/.initial=0.711,
  /num/ours_en_symarea@test300_seed0/lo/.initial=0.700,
  /num/ours_en_symarea@test300_seed0/hi/.initial=0.724,
  /num/ours_en_symarea@test300_seed0/dir/.initial=0.237,
  /num/ours_en_symarea@test300_seed0/parse/.initial=1.000,
  /num/ours_en_symarea@test300_seed0/n/.initial=300,
  /num/ours_en_symarea@test300_seed0/mae/.initial=12.35,
  /num/ours_en_symarea@test300_seed0/f1/.initial=0.663,
  /num/ours_en_symarea@test300_seed0/f1_5/.initial=0.771,
  /num/ours_en_symarea@test300_seed0/f1_15/.initial=0.572,
  /num/ours_en_symarea@test300_seed0/f1_20/.initial=0.000,
  /num/ours_en_symarea@test300_seed0/p10/.initial=0.528,
  /num/ours_en_symarea@test300_seed0/r10/.initial=0.888,
  /num/ours_en_symarea@test_full/miou/.initial=0.709,
  /num/ours_en_symarea@test_full/lo/.initial=0.706,
  /num/ours_en_symarea@test_full/hi/.initial=0.712,
  /num/ours_en_symarea@test_full/dir/.initial=0.256,
  /num/ours_en_symarea@test_full/parse/.initial=1.000,
  /num/ours_en_symarea@test_full/n/.initial=3807,
  /num/ours_final@test300_seed0/miou/.initial=0.797,
  /num/ours_final@test300_seed0/lo/.initial=0.781,
  /num/ours_final@test300_seed0/hi/.initial=0.814,
  /num/ours_final@test300_seed0/dir/.initial=0.607,
  /num/ours_final@test300_seed0/parse/.initial=1.000,
  /num/ours_final@test300_seed0/n/.initial=300,
  /num/ours_final@test300_seed0/mae/.initial=8.33,
  /num/ours_final@test300_seed0/f1/.initial=0.766,
  /num/ours_final@test300_seed0/f1_5/.initial=0.777,
  /num/ours_final@test300_seed0/f1_15/.initial=0.737,
  /num/ours_final@test300_seed0/f1_20/.initial=0.721,
  /num/ours_final@test300_seed0/p10/.initial=0.778,
  /num/ours_final@test300_seed0/r10/.initial=0.755,
  /num/ours_final@ep1@val300_seed0/miou/.initial=0.765,
  /num/ours_final@ep1@val300_seed0/lo/.initial=0.748,
  /num/ours_final@ep1@val300_seed0/hi/.initial=0.781,
  /num/ours_final@ep1@val300_seed0/dir/.initial=0.593,
  /num/ours_final@ep1@val300_seed0/parse/.initial=1.000,
  /num/ours_final@ep1@val300_seed0/n/.initial=300,
  /num/ours_final@ep2@val300_seed0/miou/.initial=0.773,
  /num/ours_final@ep2@val300_seed0/lo/.initial=0.758,
  /num/ours_final@ep2@val300_seed0/hi/.initial=0.788,
  /num/ours_final@ep2@val300_seed0/dir/.initial=0.560,
  /num/ours_final@ep2@val300_seed0/parse/.initial=1.000,
  /num/ours_final@ep2@val300_seed0/n/.initial=300,
  /num/ours_final@ep3@val300_seed0/miou/.initial=0.776,
  /num/ours_final@ep3@val300_seed0/lo/.initial=0.760,
  /num/ours_final@ep3@val300_seed0/hi/.initial=0.792,
  /num/ours_final@ep3@val300_seed0/dir/.initial=0.570,
  /num/ours_final@ep3@val300_seed0/parse/.initial=1.000,
  /num/ours_final@ep3@val300_seed0/n/.initial=300,
  /num/ours_final@ep4@val300_seed0/miou/.initial=0.787,
  /num/ours_final@ep4@val300_seed0/lo/.initial=0.772,
  /num/ours_final@ep4@val300_seed0/hi/.initial=0.803,
  /num/ours_final@ep4@val300_seed0/dir/.initial=0.587,
  /num/ours_final@ep4@val300_seed0/parse/.initial=1.000,
  /num/ours_final@ep4@val300_seed0/n/.initial=300,
  /num/ours_final@ep5@val300_seed0/miou/.initial=0.785,
  /num/ours_final@ep5@val300_seed0/lo/.initial=0.770,
  /num/ours_final@ep5@val300_seed0/hi/.initial=0.801,
  /num/ours_final@ep5@val300_seed0/dir/.initial=0.570,
  /num/ours_final@ep5@val300_seed0/parse/.initial=1.000,
  /num/ours_final@ep5@val300_seed0/n/.initial=300,
  /num/ours_final_balanced@test300_seed0/miou/.initial=0.718,
  /num/ours_final_balanced@test300_seed0/lo/.initial=0.707,
  /num/ours_final_balanced@test300_seed0/hi/.initial=0.730,
  /num/ours_final_balanced@test300_seed0/dir/.initial=0.237,
  /num/ours_final_balanced@test300_seed0/parse/.initial=1.000,
  /num/ours_final_balanced@test300_seed0/n/.initial=300,
  /num/ours_final_balanced@test300_seed0/mae/.initial=11.97,
  /num/ours_final_balanced@test300_seed0/f1/.initial=0.678,
  /num/ours_final_balanced@test300_seed0/f1_5/.initial=0.779,
  /num/ours_final_balanced@test300_seed0/f1_15/.initial=0.567,
  /num/ours_final_balanced@test300_seed0/f1_20/.initial=0.121,
  /num/ours_final_balanced@test300_seed0/p10/.initial=0.549,
  /num/ours_final_balanced@test300_seed0/r10/.initial=0.886,
  /num/ours_final_bin30@test300_seed0/miou/.initial=0.743,
  /num/ours_final_bin30@test300_seed0/lo/.initial=0.728,
  /num/ours_final_bin30@test300_seed0/hi/.initial=0.757,
  /num/ours_final_bin30@test300_seed0/dir/.initial=0.603,
  /num/ours_final_bin30@test300_seed0/parse/.initial=1.000,
  /num/ours_final_bin30@test300_seed0/n/.initial=300,
  /num/ours_final_bin30@test300_seed0/mae/.initial=10.52,
  /num/ours_final_bin30@test300_seed0/f1/.initial=0.756,
  /num/ours_final_bin30@test300_seed0/f1_5/.initial=0.744,
  /num/ours_final_bin30@test300_seed0/f1_15/.initial=0.720,
  /num/ours_final_bin30@test300_seed0/f1_20/.initial=0.672,
  /num/ours_final_bin30@test300_seed0/p10/.initial=0.784,
  /num/ours_final_bin30@test300_seed0/r10/.initial=0.730,
  /num/ours_final_fixmag@test300_seed0/miou/.initial=0.747,
  /num/ours_final_fixmag@test300_seed0/lo/.initial=0.731,
  /num/ours_final_fixmag@test300_seed0/hi/.initial=0.763,
  /num/ours_final_fixmag@test300_seed0/dir/.initial=0.603,
  /num/ours_final_fixmag@test300_seed0/parse/.initial=1.000,
  /num/ours_final_fixmag@test300_seed0/n/.initial=300,
  /num/ours_final_fixmag@test300_seed0/mae/.initial=10.60,
  /num/ours_final_fixmag@test300_seed0/f1/.initial=0.751,
  /num/ours_final_fixmag@test300_seed0/f1_5/.initial=0.777,
  /num/ours_final_fixmag@test300_seed0/f1_15/.initial=0.691,
  /num/ours_final_fixmag@test300_seed0/f1_20/.initial=0.654,
  /num/ours_final_fixmag@test300_seed0/p10/.initial=0.725,
  /num/ours_final_fixmag@test300_seed0/r10/.initial=0.779,
  /num/ours_final_q10@test300_seed0/miou/.initial=0.778,
  /num/ours_final_q10@test300_seed0/lo/.initial=0.762,
  /num/ours_final_q10@test300_seed0/hi/.initial=0.794,
  /num/ours_final_q10@test300_seed0/dir/.initial=0.610,
  /num/ours_final_q10@test300_seed0/parse/.initial=1.000,
  /num/ours_final_q10@test300_seed0/n/.initial=300,
  /num/ours_final_q10@test300_seed0/mae/.initial=9.12,
  /num/ours_final_q10@test300_seed0/f1/.initial=0.756,
  /num/ours_final_q10@test300_seed0/f1_5/.initial=0.777,
  /num/ours_final_q10@test300_seed0/f1_15/.initial=0.720,
  /num/ours_final_q10@test300_seed0/f1_20/.initial=0.703,
  /num/ours_final_q10@test300_seed0/p10/.initial=0.784,
  /num/ours_final_q10@test300_seed0/r10/.initial=0.730,
  /num/ours_final_q25@test300_seed0/miou/.initial=0.731,
  /num/ours_final_q25@test300_seed0/lo/.initial=0.717,
  /num/ours_final_q25@test300_seed0/hi/.initial=0.746,
  /num/ours_final_q25@test300_seed0/dir/.initial=0.613,
  /num/ours_final_q25@test300_seed0/parse/.initial=1.000,
  /num/ours_final_q25@test300_seed0/n/.initial=300,
  /num/ours_final_q25@test300_seed0/mae/.initial=11.36,
  /num/ours_final_q25@test300_seed0/f1/.initial=0.763,
  /num/ours_final_q25@test300_seed0/f1_5/.initial=0.753,
  /num/ours_final_q25@test300_seed0/f1_15/.initial=0.717,
  /num/ours_final_q25@test300_seed0/f1_20/.initial=0.667,
  /num/ours_final_q25@test300_seed0/p10/.initial=0.780,
  /num/ours_final_q25@test300_seed0/r10/.initial=0.746,
  /num/ours_final_seed1@test300_seed0/miou/.initial=0.789,
  /num/ours_final_seed1@test300_seed0/lo/.initial=0.773,
  /num/ours_final_seed1@test300_seed0/hi/.initial=0.805,
  /num/ours_final_seed1@test300_seed0/dir/.initial=0.583,
  /num/ours_final_seed1@test300_seed0/parse/.initial=1.000,
  /num/ours_final_seed1@test300_seed0/n/.initial=300,
  /num/ours_final_seed1@ep1@val300_seed0/miou/.initial=0.757,
  /num/ours_final_seed1@ep1@val300_seed0/lo/.initial=0.740,
  /num/ours_final_seed1@ep1@val300_seed0/hi/.initial=0.773,
  /num/ours_final_seed1@ep1@val300_seed0/dir/.initial=0.567,
  /num/ours_final_seed1@ep1@val300_seed0/parse/.initial=1.000,
  /num/ours_final_seed1@ep1@val300_seed0/n/.initial=300,
  /num/ours_final_seed1@ep2@val300_seed0/miou/.initial=0.775,
  /num/ours_final_seed1@ep2@val300_seed0/lo/.initial=0.758,
  /num/ours_final_seed1@ep2@val300_seed0/hi/.initial=0.792,
  /num/ours_final_seed1@ep2@val300_seed0/dir/.initial=0.580,
  /num/ours_final_seed1@ep2@val300_seed0/parse/.initial=1.000,
  /num/ours_final_seed1@ep2@val300_seed0/n/.initial=300,
  /num/ours_final_seed1@ep3@val300_seed0/miou/.initial=0.784,
  /num/ours_final_seed1@ep3@val300_seed0/lo/.initial=0.768,
  /num/ours_final_seed1@ep3@val300_seed0/hi/.initial=0.800,
  /num/ours_final_seed1@ep3@val300_seed0/dir/.initial=0.597,
  /num/ours_final_seed1@ep3@val300_seed0/parse/.initial=1.000,
  /num/ours_final_seed1@ep3@val300_seed0/n/.initial=300,
  /num/ours_final_seed1@ep4@val300_seed0/miou/.initial=0.796,
  /num/ours_final_seed1@ep4@val300_seed0/lo/.initial=0.780,
  /num/ours_final_seed1@ep4@val300_seed0/hi/.initial=0.811,
  /num/ours_final_seed1@ep4@val300_seed0/dir/.initial=0.597,
  /num/ours_final_seed1@ep4@val300_seed0/parse/.initial=1.000,
  /num/ours_final_seed1@ep4@val300_seed0/n/.initial=300,
  /num/ours_final_seed1@ep5@val300_seed0/miou/.initial=0.796,
  /num/ours_final_seed1@ep5@val300_seed0/lo/.initial=0.781,
  /num/ours_final_seed1@ep5@val300_seed0/hi/.initial=0.812,
  /num/ours_final_seed1@ep5@val300_seed0/dir/.initial=0.603,
  /num/ours_final_seed1@ep5@val300_seed0/parse/.initial=1.000,
  /num/ours_final_seed1@ep5@val300_seed0/n/.initial=300,
  /num/ours_final_seed2@test300_seed0/miou/.initial=0.783,
  /num/ours_final_seed2@test300_seed0/lo/.initial=0.766,
  /num/ours_final_seed2@test300_seed0/hi/.initial=0.800,
  /num/ours_final_seed2@test300_seed0/dir/.initial=0.557,
  /num/ours_final_seed2@test300_seed0/parse/.initial=1.000,
  /num/ours_final_seed2@test300_seed0/n/.initial=300,
  /num/ours_final_seed2@ep1@val300_seed0/miou/.initial=0.769,
  /num/ours_final_seed2@ep1@val300_seed0/lo/.initial=0.753,
  /num/ours_final_seed2@ep1@val300_seed0/hi/.initial=0.786,
  /num/ours_final_seed2@ep1@val300_seed0/dir/.initial=0.590,
  /num/ours_final_seed2@ep1@val300_seed0/parse/.initial=1.000,
  /num/ours_final_seed2@ep1@val300_seed0/n/.initial=300,
  /num/ours_final_seed2@ep2@val300_seed0/miou/.initial=0.775,
  /num/ours_final_seed2@ep2@val300_seed0/lo/.initial=0.759,
  /num/ours_final_seed2@ep2@val300_seed0/hi/.initial=0.791,
  /num/ours_final_seed2@ep2@val300_seed0/dir/.initial=0.580,
  /num/ours_final_seed2@ep2@val300_seed0/parse/.initial=1.000,
  /num/ours_final_seed2@ep2@val300_seed0/n/.initial=300,
  /num/ours_final_seed2@ep3@val300_seed0/miou/.initial=0.786,
  /num/ours_final_seed2@ep3@val300_seed0/lo/.initial=0.770,
  /num/ours_final_seed2@ep3@val300_seed0/hi/.initial=0.802,
  /num/ours_final_seed2@ep3@val300_seed0/dir/.initial=0.620,
  /num/ours_final_seed2@ep3@val300_seed0/parse/.initial=1.000,
  /num/ours_final_seed2@ep3@val300_seed0/n/.initial=300,
  /num/ours_final_seed2@ep4@val300_seed0/miou/.initial=0.791,
  /num/ours_final_seed2@ep4@val300_seed0/lo/.initial=0.775,
  /num/ours_final_seed2@ep4@val300_seed0/hi/.initial=0.807,
  /num/ours_final_seed2@ep4@val300_seed0/dir/.initial=0.613,
  /num/ours_final_seed2@ep4@val300_seed0/parse/.initial=1.000,
  /num/ours_final_seed2@ep4@val300_seed0/n/.initial=300,
  /num/ours_final_seed2@ep5@val300_seed0/miou/.initial=0.795,
  /num/ours_final_seed2@ep5@val300_seed0/lo/.initial=0.779,
  /num/ours_final_seed2@ep5@val300_seed0/hi/.initial=0.811,
  /num/ours_final_seed2@ep5@val300_seed0/dir/.initial=0.630,
  /num/ours_final_seed2@ep5@val300_seed0/parse/.initial=1.000,
  /num/ours_final_seed2@ep5@val300_seed0/n/.initial=300,
  /num/ours_final_symarea@test300_seed0/miou/.initial=0.713,
  /num/ours_final_symarea@test300_seed0/lo/.initial=0.700,
  /num/ours_final_symarea@test300_seed0/hi/.initial=0.725,
  /num/ours_final_symarea@test300_seed0/dir/.initial=0.237,
  /num/ours_final_symarea@test300_seed0/parse/.initial=1.000,
  /num/ours_final_symarea@test300_seed0/n/.initial=300,
  /num/ours_final_symarea@test300_seed0/mae/.initial=12.27,
  /num/ours_final_symarea@test300_seed0/f1/.initial=0.666,
  /num/ours_final_symarea@test300_seed0/f1_5/.initial=0.771,
  /num/ours_final_symarea@test300_seed0/f1_15/.initial=0.556,
  /num/ours_final_symarea@test300_seed0/f1_20/.initial=0.061,
  /num/ours_final_symarea@test300_seed0/p10/.initial=0.533,
  /num/ours_final_symarea@test300_seed0/r10/.initial=0.888,
  /num/qwen4b_zs@test300_seed0/miou/.initial=0.549,
  /num/qwen4b_zs@test300_seed0/lo/.initial=0.527,
  /num/qwen4b_zs@test300_seed0/hi/.initial=0.571,
  /num/qwen4b_zs@test300_seed0/dir/.initial=0.227,
  /num/qwen4b_zs@test300_seed0/parse/.initial=1.000,
  /num/qwen4b_zs@test300_seed0/n/.initial=300,
  /num/qwen4b_zs@test300_seed0/mae/.initial=29.37,
  /num/qwen4b_zs@test300_seed0/f1/.initial=0.504,
  /num/qwen4b_zs@test300_seed0/f1_5/.initial=0.743,
  /num/qwen4b_zs@test300_seed0/f1_15/.initial=0.437,
  /num/qwen4b_zs@test300_seed0/f1_20/.initial=0.323,
  /num/qwen4b_zs@test300_seed0/p10/.initial=0.505,
  /num/qwen4b_zs@test300_seed0/r10/.initial=0.502,
  /num/qwen4b_zs@test_full/miou/.initial=0.548,
  /num/qwen4b_zs@test_full/lo/.initial=0.542,
  /num/qwen4b_zs@test_full/hi/.initial=0.554,
  /num/qwen4b_zs@test_full/dir/.initial=0.236,
  /num/qwen4b_zs@test_full/parse/.initial=1.000,
  /num/qwen4b_zs@test_full/n/.initial=3807,
  /num/rule_cal_blend@test_full/miou/.initial=0.686,
  /num/rule_cal_blend@test_full/lo/.initial=0.682,
  /num/rule_cal_blend@test_full/hi/.initial=0.689,
  /num/rule_cal_blend@test_full/dir/.initial=0.489,
  /num/rule_cal_blend@test_full/parse/.initial=1.000,
  /num/rule_cal_blend@test_full/n/.initial=3807,
  /num/rule_cal_blend@val300_seed0/miou/.initial=0.685,
  /num/rule_cal_blend@val300_seed0/lo/.initial=0.672,
  /num/rule_cal_blend@val300_seed0/hi/.initial=0.698,
  /num/rule_cal_blend@val300_seed0/dir/.initial=0.477,
  /num/rule_cal_blend@val300_seed0/parse/.initial=1.000,
  /num/rule_cal_blend@val300_seed0/n/.initial=300,
  /num/rule_cal_subj@test_full/miou/.initial=0.669,
  /num/rule_cal_subj@test_full/lo/.initial=0.665,
  /num/rule_cal_subj@test_full/hi/.initial=0.672,
  /num/rule_cal_subj@test_full/dir/.initial=0.460,
  /num/rule_cal_subj@test_full/parse/.initial=1.000,
  /num/rule_cal_subj@test_full/n/.initial=3807,
  /num/rule_cal_subj@val300_seed0/miou/.initial=0.668,
  /num/rule_cal_subj@val300_seed0/lo/.initial=0.655,
  /num/rule_cal_subj@val300_seed0/hi/.initial=0.681,
  /num/rule_cal_subj@val300_seed0/dir/.initial=0.443,
  /num/rule_cal_subj@val300_seed0/parse/.initial=1.000,
  /num/rule_cal_subj@val300_seed0/n/.initial=300,
  /num/rule_center_blend@test_full/miou/.initial=0.688,
  /num/rule_center_blend@test_full/lo/.initial=0.685,
  /num/rule_center_blend@test_full/hi/.initial=0.691,
  /num/rule_center_blend@test_full/dir/.initial=0.504,
  /num/rule_center_blend@test_full/parse/.initial=1.000,
  /num/rule_center_blend@test_full/n/.initial=3807,
  /num/rule_center_blend@val300_seed0/miou/.initial=0.692,
  /num/rule_center_blend@val300_seed0/lo/.initial=0.679,
  /num/rule_center_blend@val300_seed0/hi/.initial=0.704,
  /num/rule_center_blend@val300_seed0/dir/.initial=0.503,
  /num/rule_center_blend@val300_seed0/parse/.initial=1.000,
  /num/rule_center_blend@val300_seed0/n/.initial=300,
  /num/rule_center_fix@test_full/miou/.initial=0.686,
  /num/rule_center_fix@test_full/lo/.initial=0.682,
  /num/rule_center_fix@test_full/hi/.initial=0.689,
  /num/rule_center_fix@test_full/dir/.initial=0.481,
  /num/rule_center_fix@test_full/parse/.initial=1.000,
  /num/rule_center_fix@test_full/n/.initial=3807,
  /num/rule_center_fix@val300_seed0/miou/.initial=0.691,
  /num/rule_center_fix@val300_seed0/lo/.initial=0.678,
  /num/rule_center_fix@val300_seed0/hi/.initial=0.704,
  /num/rule_center_fix@val300_seed0/dir/.initial=0.517,
  /num/rule_center_fix@val300_seed0/parse/.initial=1.000,
  /num/rule_center_fix@val300_seed0/n/.initial=300,
  /num/rule_center_subj@test_full/miou/.initial=0.661,
  /num/rule_center_subj@test_full/lo/.initial=0.657,
  /num/rule_center_subj@test_full/hi/.initial=0.665,
  /num/rule_center_subj@test_full/dir/.initial=0.454,
  /num/rule_center_subj@test_full/parse/.initial=1.000,
  /num/rule_center_subj@test_full/n/.initial=3807,
  /num/rule_center_subj@val300_seed0/miou/.initial=0.664,
  /num/rule_center_subj@val300_seed0/lo/.initial=0.650,
  /num/rule_center_subj@val300_seed0/hi/.initial=0.677,
  /num/rule_center_subj@val300_seed0/dir/.initial=0.443,
  /num/rule_center_subj@val300_seed0/parse/.initial=1.000,
  /num/rule_center_subj@val300_seed0/n/.initial=300,
  /num/rule_hierarchy@test_full/miou/.initial=0.690,
  /num/rule_hierarchy@test_full/lo/.initial=0.686,
  /num/rule_hierarchy@test_full/hi/.initial=0.694,
  /num/rule_hierarchy@test_full/dir/.initial=0.419,
  /num/rule_hierarchy@test_full/parse/.initial=1.000,
  /num/rule_hierarchy@test_full/n/.initial=3807,
  /num/rule_hierarchy@val300_seed0/miou/.initial=0.699,
  /num/rule_hierarchy@val300_seed0/lo/.initial=0.684,
  /num/rule_hierarchy@val300_seed0/hi/.initial=0.713,
  /num/rule_hierarchy@val300_seed0/dir/.initial=0.440,
  /num/rule_hierarchy@val300_seed0/parse/.initial=1.000,
  /num/rule_hierarchy@val300_seed0/n/.initial=300,
  /num/rule_saliency@test300_seed0/miou/.initial=0.692,
  /num/rule_saliency@test300_seed0/lo/.initial=0.678,
  /num/rule_saliency@test300_seed0/hi/.initial=0.706,
  /num/rule_saliency@test300_seed0/dir/.initial=0.430,
  /num/rule_saliency@test300_seed0/parse/.initial=1.000,
  /num/rule_saliency@test300_seed0/n/.initial=300,
  /num/rule_saliency@test300_seed0/mae/.initial=13.14,
  /num/rule_saliency@test300_seed0/f1/.initial=0.636,
  /num/rule_saliency@test300_seed0/f1_5/.initial=0.686,
  /num/rule_saliency@test300_seed0/f1_15/.initial=0.565,
  /num/rule_saliency@test300_seed0/f1_20/.initial=0.546,
  /num/rule_saliency@test300_seed0/p10/.initial=0.582,
  /num/rule_saliency@test300_seed0/r10/.initial=0.700,
  /num/rule_saliency@test_full/miou/.initial=0.690,
  /num/rule_saliency@test_full/lo/.initial=0.686,
  /num/rule_saliency@test_full/hi/.initial=0.694,
  /num/rule_saliency@test_full/dir/.initial=0.419,
  /num/rule_saliency@test_full/parse/.initial=1.000,
  /num/rule_saliency@test_full/n/.initial=3807,
  /num/rule_thirds_blend@test_full/miou/.initial=0.571,
  /num/rule_thirds_blend@test_full/lo/.initial=0.568,
  /num/rule_thirds_blend@test_full/hi/.initial=0.575,
  /num/rule_thirds_blend@test_full/dir/.initial=0.198,
  /num/rule_thirds_blend@test_full/parse/.initial=1.000,
  /num/rule_thirds_blend@test_full/n/.initial=3807,
  /num/rule_thirds_blend@val300_seed0/miou/.initial=0.575,
  /num/rule_thirds_blend@val300_seed0/lo/.initial=0.562,
  /num/rule_thirds_blend@val300_seed0/hi/.initial=0.587,
  /num/rule_thirds_blend@val300_seed0/dir/.initial=0.200,
  /num/rule_thirds_blend@val300_seed0/parse/.initial=1.000,
  /num/rule_thirds_blend@val300_seed0/n/.initial=300,
  /num/rule_thirds_subj@test_full/miou/.initial=0.586,
  /num/rule_thirds_subj@test_full/lo/.initial=0.582,
  /num/rule_thirds_subj@test_full/hi/.initial=0.590,
  /num/rule_thirds_subj@test_full/dir/.initial=0.265,
  /num/rule_thirds_subj@test_full/parse/.initial=1.000,
  /num/rule_thirds_subj@test_full/n/.initial=3807,
  /num/rule_thirds_subj@val300_seed0/miou/.initial=0.584,
  /num/rule_thirds_subj@val300_seed0/lo/.initial=0.570,
  /num/rule_thirds_subj@val300_seed0/hi/.initial=0.598,
  /num/rule_thirds_subj@val300_seed0/dir/.initial=0.257,
  /num/rule_thirds_subj@val300_seed0/parse/.initial=1.000,
  /num/rule_thirds_subj@val300_seed0/n/.initial=300,
  /num/sym1d15k@test300_seed0/miou/.initial=0.709,
  /num/sym1d15k@test300_seed0/lo/.initial=0.697,
  /num/sym1d15k@test300_seed0/hi/.initial=0.722,
  /num/sym1d15k@test300_seed0/dir/.initial=0.237,
  /num/sym1d15k@test300_seed0/parse/.initial=1.000,
  /num/sym1d15k@test300_seed0/n/.initial=300,
  /num/sym1d15k@test300_seed0/mae/.initial=12.51,
  /num/sym1d15k@test300_seed0/f1/.initial=0.662,
  /num/sym1d15k@test300_seed0/f1_5/.initial=0.771,
  /num/sym1d15k@test300_seed0/f1_15/.initial=0.562,
  /num/sym1d15k@test300_seed0/f1_20/.initial=0.072,
  /num/sym1d15k@test300_seed0/p10/.initial=0.532,
  /num/sym1d15k@test300_seed0/r10/.initial=0.876,
  /num/sym1d15k@ep1@val300_seed0/miou/.initial=0.703,
  /num/sym1d15k@ep1@val300_seed0/lo/.initial=0.693,
  /num/sym1d15k@ep1@val300_seed0/hi/.initial=0.713,
  /num/sym1d15k@ep1@val300_seed0/dir/.initial=0.307,
  /num/sym1d15k@ep1@val300_seed0/parse/.initial=1.000,
  /num/sym1d15k@ep1@val300_seed0/n/.initial=300,
  /num/sym1d15k@ep10@val300_seed0/miou/.initial=0.703,
  /num/sym1d15k@ep10@val300_seed0/lo/.initial=0.692,
  /num/sym1d15k@ep10@val300_seed0/hi/.initial=0.713,
  /num/sym1d15k@ep10@val300_seed0/dir/.initial=0.307,
  /num/sym1d15k@ep10@val300_seed0/parse/.initial=1.000,
  /num/sym1d15k@ep10@val300_seed0/n/.initial=300,
  /num/sym1d15k@ep2@val300_seed0/miou/.initial=0.704,
  /num/sym1d15k@ep2@val300_seed0/lo/.initial=0.694,
  /num/sym1d15k@ep2@val300_seed0/hi/.initial=0.714,
  /num/sym1d15k@ep2@val300_seed0/dir/.initial=0.307,
  /num/sym1d15k@ep2@val300_seed0/parse/.initial=1.000,
  /num/sym1d15k@ep2@val300_seed0/n/.initial=300,
  /num/sym1d15k@ep3@val300_seed0/miou/.initial=0.704,
  /num/sym1d15k@ep3@val300_seed0/lo/.initial=0.694,
  /num/sym1d15k@ep3@val300_seed0/hi/.initial=0.715,
  /num/sym1d15k@ep3@val300_seed0/dir/.initial=0.307,
  /num/sym1d15k@ep3@val300_seed0/parse/.initial=1.000,
  /num/sym1d15k@ep3@val300_seed0/n/.initial=300,
  /num/sym1d15k@ep4@val300_seed0/miou/.initial=0.705,
  /num/sym1d15k@ep4@val300_seed0/lo/.initial=0.695,
  /num/sym1d15k@ep4@val300_seed0/hi/.initial=0.716,
  /num/sym1d15k@ep4@val300_seed0/dir/.initial=0.307,
  /num/sym1d15k@ep4@val300_seed0/parse/.initial=1.000,
  /num/sym1d15k@ep4@val300_seed0/n/.initial=300,
  /num/sym1d15k@ep5@val300_seed0/miou/.initial=0.705,
  /num/sym1d15k@ep5@val300_seed0/lo/.initial=0.694,
  /num/sym1d15k@ep5@val300_seed0/hi/.initial=0.715,
  /num/sym1d15k@ep5@val300_seed0/dir/.initial=0.307,
  /num/sym1d15k@ep5@val300_seed0/parse/.initial=1.000,
  /num/sym1d15k@ep5@val300_seed0/n/.initial=300,
  /num/sym1d15k@ep6@val300_seed0/miou/.initial=0.703,
  /num/sym1d15k@ep6@val300_seed0/lo/.initial=0.693,
  /num/sym1d15k@ep6@val300_seed0/hi/.initial=0.713,
  /num/sym1d15k@ep6@val300_seed0/dir/.initial=0.307,
  /num/sym1d15k@ep6@val300_seed0/parse/.initial=1.000,
  /num/sym1d15k@ep6@val300_seed0/n/.initial=300,
  /num/sym1d15k@ep7@val300_seed0/miou/.initial=0.703,
  /num/sym1d15k@ep7@val300_seed0/lo/.initial=0.693,
  /num/sym1d15k@ep7@val300_seed0/hi/.initial=0.713,
  /num/sym1d15k@ep7@val300_seed0/dir/.initial=0.307,
  /num/sym1d15k@ep7@val300_seed0/parse/.initial=1.000,
  /num/sym1d15k@ep7@val300_seed0/n/.initial=300,
  /num/sym1d15k@ep8@val300_seed0/miou/.initial=0.703,
  /num/sym1d15k@ep8@val300_seed0/lo/.initial=0.693,
  /num/sym1d15k@ep8@val300_seed0/hi/.initial=0.713,
  /num/sym1d15k@ep8@val300_seed0/dir/.initial=0.307,
  /num/sym1d15k@ep8@val300_seed0/parse/.initial=1.000,
  /num/sym1d15k@ep8@val300_seed0/n/.initial=300,
  /num/sym1d15k@ep9@val300_seed0/miou/.initial=0.702,
  /num/sym1d15k@ep9@val300_seed0/lo/.initial=0.692,
  /num/sym1d15k@ep9@val300_seed0/hi/.initial=0.713,
  /num/sym1d15k@ep9@val300_seed0/dir/.initial=0.307,
  /num/sym1d15k@ep9@val300_seed0/parse/.initial=1.000,
  /num/sym1d15k@ep9@val300_seed0/n/.initial=300,
  /num/unic@test300_seed0/miou/.initial=0.596,
  /num/unic@test300_seed0/lo/.initial=0.584,
  /num/unic@test300_seed0/hi/.initial=0.607,
  /num/unic@test300_seed0/dir/.initial=0.453,
  /num/unic@test300_seed0/parse/.initial=1.000,
  /num/unic@test300_seed0/n/.initial=300,
  /num/unic@test300_seed0/mae/.initial=13.72,
  /num/unic@test300_seed0/f1/.initial=0.213,
  /num/unic@test300_seed0/f1_5/.initial=0.410,
  /num/unic@test300_seed0/f1_15/.initial=0.126,
  /num/unic@test300_seed0/f1_20/.initial=0.054,
  /num/unic@test300_seed0/p10/.initial=0.765,
  /num/unic@test300_seed0/r10/.initial=0.124,
  /num/unic@test_full/miou/.initial=0.586,
  /num/unic@test_full/lo/.initial=0.583,
  /num/unic@test_full/hi/.initial=0.589,
  /num/unic@test_full/dir/.initial=0.455,
  /num/unic@test_full/parse/.initial=1.000,
  /num/unic@test_full/n/.initial=3807,
  /num/unic_clip0@test300_seed0/miou/.initial=0.643,
  /num/unic_clip0@test300_seed0/lo/.initial=0.631,
  /num/unic_clip0@test300_seed0/hi/.initial=0.654,
  /num/unic_clip0@test300_seed0/dir/.initial=0.413,
  /num/unic_clip0@test300_seed0/parse/.initial=1.000,
  /num/unic_clip0@test300_seed0/n/.initial=300,
  /num/unic_clip0@test300_seed0/mae/.initial=13.72,
  /num/unic_clip0@test300_seed0/f1/.initial=0.213,
  /num/unic_clip0@test300_seed0/f1_5/.initial=0.410,
  /num/unic_clip0@test300_seed0/f1_15/.initial=0.126,
  /num/unic_clip0@test300_seed0/f1_20/.initial=0.054,
  /num/unic_clip0@test300_seed0/p10/.initial=0.765,
  /num/unic_clip0@test300_seed0/r10/.initial=0.124,
  /num/unic_clip0@test_full/miou/.initial=0.635,
  /num/unic_clip0@test_full/lo/.initial=0.632,
  /num/unic_clip0@test_full/hi/.initial=0.638,
  /num/unic_clip0@test_full/dir/.initial=0.434,
  /num/unic_clip0@test_full/parse/.initial=1.000,
  /num/unic_clip0@test_full/n/.initial=3807,
  /num/InternVL3-2B-hf_icl9/miou/.initial=0.606,
  /num/InternVL3-2B-hf_icl9/lo/.initial=0.603,
  /num/InternVL3-2B-hf_icl9/hi/.initial=0.609,
  /num/InternVL3-2B-hf_icl9/dir/.initial=0.255,
  /num/InternVL3-2B-hf_icl9/parse/.initial=1.000,
  /num/InternVL3-2B-hf_icl9/n/.initial=3807,
  /num/InternVL3-2B-hf_zs/miou/.initial=0.611,
  /num/InternVL3-2B-hf_zs/lo/.initial=0.608,
  /num/InternVL3-2B-hf_zs/hi/.initial=0.614,
  /num/InternVL3-2B-hf_zs/dir/.initial=0.256,
  /num/InternVL3-2B-hf_zs/parse/.initial=1.000,
  /num/InternVL3-2B-hf_zs/n/.initial=3807,
  /num/abl_en_15k_s8090/miou/.initial=0.788,
  /num/abl_en_15k_s8090/lo/.initial=0.772,
  /num/abl_en_15k_s8090/hi/.initial=0.805,
  /num/abl_en_15k_s8090/dir/.initial=0.573,
  /num/abl_en_15k_s8090/parse/.initial=1.000,
  /num/abl_en_15k_s8090/n/.initial=300,
  /num/abl_en_15k_s8090@ep1/miou/.initial=0.754,
  /num/abl_en_15k_s8090@ep1/lo/.initial=0.738,
  /num/abl_en_15k_s8090@ep1/hi/.initial=0.771,
  /num/abl_en_15k_s8090@ep1/dir/.initial=0.520,
  /num/abl_en_15k_s8090@ep1/parse/.initial=1.000,
  /num/abl_en_15k_s8090@ep1/n/.initial=300,
  /num/abl_en_15k_s8090@ep2/miou/.initial=0.781,
  /num/abl_en_15k_s8090@ep2/lo/.initial=0.765,
  /num/abl_en_15k_s8090@ep2/hi/.initial=0.797,
  /num/abl_en_15k_s8090@ep2/dir/.initial=0.593,
  /num/abl_en_15k_s8090@ep2/parse/.initial=1.000,
  /num/abl_en_15k_s8090@ep2/n/.initial=300,
  /num/abl_en_15k_s8090@ep3/miou/.initial=0.778,
  /num/abl_en_15k_s8090@ep3/lo/.initial=0.761,
  /num/abl_en_15k_s8090@ep3/hi/.initial=0.793,
  /num/abl_en_15k_s8090@ep3/dir/.initial=0.573,
  /num/abl_en_15k_s8090@ep3/parse/.initial=1.000,
  /num/abl_en_15k_s8090@ep3/n/.initial=300,
  /num/abl_en_15k_s8090@ep4/miou/.initial=0.793,
  /num/abl_en_15k_s8090@ep4/lo/.initial=0.777,
  /num/abl_en_15k_s8090@ep4/hi/.initial=0.808,
  /num/abl_en_15k_s8090@ep4/dir/.initial=0.613,
  /num/abl_en_15k_s8090@ep4/parse/.initial=1.000,
  /num/abl_en_15k_s8090@ep4/n/.initial=300,
  /num/abl_en_15k_s8090@ep5/miou/.initial=0.794,
  /num/abl_en_15k_s8090@ep5/lo/.initial=0.779,
  /num/abl_en_15k_s8090@ep5/hi/.initial=0.810,
  /num/abl_en_15k_s8090@ep5/dir/.initial=0.610,
  /num/abl_en_15k_s8090@ep5/parse/.initial=1.000,
  /num/abl_en_15k_s8090@ep5/n/.initial=300,
  /num/abl_en_5k_s8090/miou/.initial=0.786,
  /num/abl_en_5k_s8090/lo/.initial=0.770,
  /num/abl_en_5k_s8090/hi/.initial=0.802,
  /num/abl_en_5k_s8090/dir/.initial=0.553,
  /num/abl_en_5k_s8090/parse/.initial=1.000,
  /num/abl_en_5k_s8090/n/.initial=300,
  /num/abl_en_5k_s8090@ep1/miou/.initial=0.757,
  /num/abl_en_5k_s8090@ep1/lo/.initial=0.741,
  /num/abl_en_5k_s8090@ep1/hi/.initial=0.775,
  /num/abl_en_5k_s8090@ep1/dir/.initial=0.567,
  /num/abl_en_5k_s8090@ep1/parse/.initial=1.000,
  /num/abl_en_5k_s8090@ep1/n/.initial=300,
  /num/abl_en_5k_s8090@ep2/miou/.initial=0.776,
  /num/abl_en_5k_s8090@ep2/lo/.initial=0.759,
  /num/abl_en_5k_s8090@ep2/hi/.initial=0.791,
  /num/abl_en_5k_s8090@ep2/dir/.initial=0.583,
  /num/abl_en_5k_s8090@ep2/parse/.initial=1.000,
  /num/abl_en_5k_s8090@ep2/n/.initial=300,
  /num/abl_en_5k_s8090@ep3/miou/.initial=0.785,
  /num/abl_en_5k_s8090@ep3/lo/.initial=0.771,
  /num/abl_en_5k_s8090@ep3/hi/.initial=0.800,
  /num/abl_en_5k_s8090@ep3/dir/.initial=0.573,
  /num/abl_en_5k_s8090@ep3/parse/.initial=1.000,
  /num/abl_en_5k_s8090@ep3/n/.initial=300,
  /num/abl_en_5k_s8090@ep4/miou/.initial=0.786,
  /num/abl_en_5k_s8090@ep4/lo/.initial=0.770,
  /num/abl_en_5k_s8090@ep4/hi/.initial=0.801,
  /num/abl_en_5k_s8090@ep4/dir/.initial=0.527,
  /num/abl_en_5k_s8090@ep4/parse/.initial=1.000,
  /num/abl_en_5k_s8090@ep4/n/.initial=300,
  /num/abl_en_5k_s8090@ep5/miou/.initial=0.782,
  /num/abl_en_5k_s8090@ep5/lo/.initial=0.765,
  /num/abl_en_5k_s8090@ep5/hi/.initial=0.797,
  /num/abl_en_5k_s8090@ep5/dir/.initial=0.533,
  /num/abl_en_5k_s8090@ep5/parse/.initial=1.000,
  /num/abl_en_5k_s8090@ep5/n/.initial=300,
  /num/abl_en_r1to0/miou/.initial=0.745,
  /num/abl_en_r1to0/lo/.initial=0.740,
  /num/abl_en_r1to0/hi/.initial=0.749,
  /num/abl_en_r1to0/dir/.initial=0.536,
  /num/abl_en_r1to0/parse/.initial=1.000,
  /num/abl_en_r1to0/n/.initial=3807,
  /num/abl_en_r1to0@ep1/miou/.initial=0.697,
  /num/abl_en_r1to0@ep1/lo/.initial=0.679,
  /num/abl_en_r1to0@ep1/hi/.initial=0.716,
  /num/abl_en_r1to0@ep1/dir/.initial=0.507,
  /num/abl_en_r1to0@ep1/parse/.initial=1.000,
  /num/abl_en_r1to0@ep1/n/.initial=300,
  /num/abl_en_r1to0@ep2/miou/.initial=0.732,
  /num/abl_en_r1to0@ep2/lo/.initial=0.714,
  /num/abl_en_r1to0@ep2/hi/.initial=0.749,
  /num/abl_en_r1to0@ep2/dir/.initial=0.547,
  /num/abl_en_r1to0@ep2/parse/.initial=1.000,
  /num/abl_en_r1to0@ep2/n/.initial=300,
  /num/abl_en_r1to0@ep3/miou/.initial=0.746,
  /num/abl_en_r1to0@ep3/lo/.initial=0.728,
  /num/abl_en_r1to0@ep3/hi/.initial=0.763,
  /num/abl_en_r1to0@ep3/dir/.initial=0.517,
  /num/abl_en_r1to0@ep3/parse/.initial=1.000,
  /num/abl_en_r1to0@ep3/n/.initial=300,
  /num/abl_en_r1to0@ep4/miou/.initial=0.754,
  /num/abl_en_r1to0@ep4/lo/.initial=0.737,
  /num/abl_en_r1to0@ep4/hi/.initial=0.771,
  /num/abl_en_r1to0@ep4/dir/.initial=0.543,
  /num/abl_en_r1to0@ep4/parse/.initial=1.000,
  /num/abl_en_r1to0@ep4/n/.initial=300,
  /num/abl_en_r1to0@ep5/miou/.initial=0.754,
  /num/abl_en_r1to0@ep5/lo/.initial=0.737,
  /num/abl_en_r1to0@ep5/hi/.initial=0.769,
  /num/abl_en_r1to0@ep5/dir/.initial=0.540,
  /num/abl_en_r1to0@ep5/parse/.initial=1.000,
  /num/abl_en_r1to0@ep5/n/.initial=300,
  /num/abl_en_r1to0_s8090/miou/.initial=0.741,
  /num/abl_en_r1to0_s8090/lo/.initial=0.737,
  /num/abl_en_r1to0_s8090/hi/.initial=0.746,
  /num/abl_en_r1to0_s8090/dir/.initial=0.498,
  /num/abl_en_r1to0_s8090/parse/.initial=0.999,
  /num/abl_en_r1to0_s8090/n/.initial=3807,
  /num/abl_en_r1to0_s8090@ep1/miou/.initial=0.752,
  /num/abl_en_r1to0_s8090@ep1/lo/.initial=0.734,
  /num/abl_en_r1to0_s8090@ep1/hi/.initial=0.769,
  /num/abl_en_r1to0_s8090@ep1/dir/.initial=0.533,
  /num/abl_en_r1to0_s8090@ep1/parse/.initial=1.000,
  /num/abl_en_r1to0_s8090@ep1/n/.initial=300,
  /num/abl_en_r1to0_s8090@ep2/miou/.initial=0.757,
  /num/abl_en_r1to0_s8090@ep2/lo/.initial=0.741,
  /num/abl_en_r1to0_s8090@ep2/hi/.initial=0.772,
  /num/abl_en_r1to0_s8090@ep2/dir/.initial=0.540,
  /num/abl_en_r1to0_s8090@ep2/parse/.initial=1.000,
  /num/abl_en_r1to0_s8090@ep2/n/.initial=300,
  /num/abl_en_r1to0_s8090@ep3/miou/.initial=0.758,
  /num/abl_en_r1to0_s8090@ep3/lo/.initial=0.742,
  /num/abl_en_r1to0_s8090@ep3/hi/.initial=0.775,
  /num/abl_en_r1to0_s8090@ep3/dir/.initial=0.530,
  /num/abl_en_r1to0_s8090@ep3/parse/.initial=1.000,
  /num/abl_en_r1to0_s8090@ep3/n/.initial=300,
  /num/abl_en_r1to0_s8090@ep4/miou/.initial=0.756,
  /num/abl_en_r1to0_s8090@ep4/lo/.initial=0.740,
  /num/abl_en_r1to0_s8090@ep4/hi/.initial=0.772,
  /num/abl_en_r1to0_s8090@ep4/dir/.initial=0.577,
  /num/abl_en_r1to0_s8090@ep4/parse/.initial=1.000,
  /num/abl_en_r1to0_s8090@ep4/n/.initial=300,
  /num/abl_en_r1to0_s8090@ep5/miou/.initial=0.755,
  /num/abl_en_r1to0_s8090@ep5/lo/.initial=0.738,
  /num/abl_en_r1to0_s8090@ep5/hi/.initial=0.772,
  /num/abl_en_r1to0_s8090@ep5/dir/.initial=0.540,
  /num/abl_en_r1to0_s8090@ep5/parse/.initial=1.000,
  /num/abl_en_r1to0_s8090@ep5/n/.initial=300,
  /num/abl_en_r1to1_s8090/miou/.initial=0.760,
  /num/abl_en_r1to1_s8090/lo/.initial=0.755,
  /num/abl_en_r1to1_s8090/hi/.initial=0.765,
  /num/abl_en_r1to1_s8090/dir/.initial=0.525,
  /num/abl_en_r1to1_s8090/parse/.initial=1.000,
  /num/abl_en_r1to1_s8090/n/.initial=3807,
  /num/abl_en_r1to1_s8090@ep1/miou/.initial=0.762,
  /num/abl_en_r1to1_s8090@ep1/lo/.initial=0.746,
  /num/abl_en_r1to1_s8090@ep1/hi/.initial=0.778,
  /num/abl_en_r1to1_s8090@ep1/dir/.initial=0.550,
  /num/abl_en_r1to1_s8090@ep1/parse/.initial=1.000,
  /num/abl_en_r1to1_s8090@ep1/n/.initial=300,
  /num/abl_en_r1to1_s8090@ep2/miou/.initial=0.777,
  /num/abl_en_r1to1_s8090@ep2/lo/.initial=0.762,
  /num/abl_en_r1to1_s8090@ep2/hi/.initial=0.793,
  /num/abl_en_r1to1_s8090@ep2/dir/.initial=0.580,
  /num/abl_en_r1to1_s8090@ep2/parse/.initial=1.000,
  /num/abl_en_r1to1_s8090@ep2/n/.initial=300,
  /num/abl_en_r1to1_s8090@ep3/miou/.initial=0.774,
  /num/abl_en_r1to1_s8090@ep3/lo/.initial=0.758,
  /num/abl_en_r1to1_s8090@ep3/hi/.initial=0.789,
  /num/abl_en_r1to1_s8090@ep3/dir/.initial=0.560,
  /num/abl_en_r1to1_s8090@ep3/parse/.initial=1.000,
  /num/abl_en_r1to1_s8090@ep3/n/.initial=300,
  /num/abl_en_r1to1_s8090@ep4/miou/.initial=0.775,
  /num/abl_en_r1to1_s8090@ep4/lo/.initial=0.760,
  /num/abl_en_r1to1_s8090@ep4/hi/.initial=0.790,
  /num/abl_en_r1to1_s8090@ep4/dir/.initial=0.563,
  /num/abl_en_r1to1_s8090@ep4/parse/.initial=1.000,
  /num/abl_en_r1to1_s8090@ep4/n/.initial=300,
  /num/abl_en_r1to1_s8090@ep5/miou/.initial=0.773,
  /num/abl_en_r1to1_s8090@ep5/lo/.initial=0.759,
  /num/abl_en_r1to1_s8090@ep5/hi/.initial=0.788,
  /num/abl_en_r1to1_s8090@ep5/dir/.initial=0.557,
  /num/abl_en_r1to1_s8090@ep5/parse/.initial=1.000,
  /num/abl_en_r1to1_s8090@ep5/n/.initial=300,
  /num/abl_en_r1to2_s8090/miou/.initial=0.773,
  /num/abl_en_r1to2_s8090/lo/.initial=0.758,
  /num/abl_en_r1to2_s8090/hi/.initial=0.788,
  /num/abl_en_r1to2_s8090/dir/.initial=0.563,
  /num/abl_en_r1to2_s8090/parse/.initial=1.000,
  /num/abl_en_r1to2_s8090/n/.initial=300,
  /num/abl_en_r1to2_s8090@ep1/miou/.initial=0.750,
  /num/abl_en_r1to2_s8090@ep1/lo/.initial=0.732,
  /num/abl_en_r1to2_s8090@ep1/hi/.initial=0.768,
  /num/abl_en_r1to2_s8090@ep1/dir/.initial=0.513,
  /num/abl_en_r1to2_s8090@ep1/parse/.initial=1.000,
  /num/abl_en_r1to2_s8090@ep1/n/.initial=300,
  /num/abl_en_r1to2_s8090@ep2/miou/.initial=0.762,
  /num/abl_en_r1to2_s8090@ep2/lo/.initial=0.746,
  /num/abl_en_r1to2_s8090@ep2/hi/.initial=0.778,
  /num/abl_en_r1to2_s8090@ep2/dir/.initial=0.547,
  /num/abl_en_r1to2_s8090@ep2/parse/.initial=1.000,
  /num/abl_en_r1to2_s8090@ep2/n/.initial=300,
  /num/abl_en_r1to2_s8090@ep3/miou/.initial=0.773,
  /num/abl_en_r1to2_s8090@ep3/lo/.initial=0.759,
  /num/abl_en_r1to2_s8090@ep3/hi/.initial=0.788,
  /num/abl_en_r1to2_s8090@ep3/dir/.initial=0.520,
  /num/abl_en_r1to2_s8090@ep3/parse/.initial=1.000,
  /num/abl_en_r1to2_s8090@ep3/n/.initial=300,
  /num/abl_en_r1to2_s8090@ep4/miou/.initial=0.782,
  /num/abl_en_r1to2_s8090@ep4/lo/.initial=0.767,
  /num/abl_en_r1to2_s8090@ep4/hi/.initial=0.796,
  /num/abl_en_r1to2_s8090@ep4/dir/.initial=0.553,
  /num/abl_en_r1to2_s8090@ep4/parse/.initial=1.000,
  /num/abl_en_r1to2_s8090@ep4/n/.initial=300,
  /num/abl_en_r1to2_s8090@ep5/miou/.initial=0.781,
  /num/abl_en_r1to2_s8090@ep5/lo/.initial=0.766,
  /num/abl_en_r1to2_s8090@ep5/hi/.initial=0.796,
  /num/abl_en_r1to2_s8090@ep5/dir/.initial=0.560,
  /num/abl_en_r1to2_s8090@ep5/parse/.initial=1.000,
  /num/abl_en_r1to2_s8090@ep5/n/.initial=300,
  /num/abl_en_scale10k_s8090/miou/.initial=0.791,
  /num/abl_en_scale10k_s8090/lo/.initial=0.774,
  /num/abl_en_scale10k_s8090/hi/.initial=0.809,
  /num/abl_en_scale10k_s8090/dir/.initial=0.580,
  /num/abl_en_scale10k_s8090/parse/.initial=1.000,
  /num/abl_en_scale10k_s8090/n/.initial=300,
  /num/abl_en_scale10k_s8090@ep1/miou/.initial=0.743,
  /num/abl_en_scale10k_s8090@ep1/lo/.initial=0.725,
  /num/abl_en_scale10k_s8090@ep1/hi/.initial=0.761,
  /num/abl_en_scale10k_s8090@ep1/dir/.initial=0.533,
  /num/abl_en_scale10k_s8090@ep1/parse/.initial=1.000,
  /num/abl_en_scale10k_s8090@ep1/n/.initial=300,
  /num/abl_en_scale10k_s8090@ep2/miou/.initial=0.786,
  /num/abl_en_scale10k_s8090@ep2/lo/.initial=0.770,
  /num/abl_en_scale10k_s8090@ep2/hi/.initial=0.803,
  /num/abl_en_scale10k_s8090@ep2/dir/.initial=0.597,
  /num/abl_en_scale10k_s8090@ep2/parse/.initial=1.000,
  /num/abl_en_scale10k_s8090@ep2/n/.initial=300,
  /num/abl_en_scale10k_s8090@ep3/miou/.initial=0.787,
  /num/abl_en_scale10k_s8090@ep3/lo/.initial=0.770,
  /num/abl_en_scale10k_s8090@ep3/hi/.initial=0.802,
  /num/abl_en_scale10k_s8090@ep3/dir/.initial=0.587,
  /num/abl_en_scale10k_s8090@ep3/parse/.initial=1.000,
  /num/abl_en_scale10k_s8090@ep3/n/.initial=300,
  /num/abl_en_scale10k_s8090@ep4/miou/.initial=0.795,
  /num/abl_en_scale10k_s8090@ep4/lo/.initial=0.778,
  /num/abl_en_scale10k_s8090@ep4/hi/.initial=0.811,
  /num/abl_en_scale10k_s8090@ep4/dir/.initial=0.587,
  /num/abl_en_scale10k_s8090@ep4/parse/.initial=1.000,
  /num/abl_en_scale10k_s8090@ep4/n/.initial=300,
  /num/abl_en_scale10k_s8090@ep5/miou/.initial=0.795,
  /num/abl_en_scale10k_s8090@ep5/lo/.initial=0.778,
  /num/abl_en_scale10k_s8090@ep5/hi/.initial=0.811,
  /num/abl_en_scale10k_s8090@ep5/dir/.initial=0.593,
  /num/abl_en_scale10k_s8090@ep5/parse/.initial=1.000,
  /num/abl_en_scale10k_s8090@ep5/n/.initial=300,
  /num/abl_en_scalefull_s8090/miou/.initial=0.790,
  /num/abl_en_scalefull_s8090/lo/.initial=0.771,
  /num/abl_en_scalefull_s8090/hi/.initial=0.807,
  /num/abl_en_scalefull_s8090/dir/.initial=0.590,
  /num/abl_en_scalefull_s8090/parse/.initial=1.000,
  /num/abl_en_scalefull_s8090/n/.initial=300,
  /num/abl_en_scalefull_s8090@ep1/miou/.initial=0.766,
  /num/abl_en_scalefull_s8090@ep1/lo/.initial=0.749,
  /num/abl_en_scalefull_s8090@ep1/hi/.initial=0.783,
  /num/abl_en_scalefull_s8090@ep1/dir/.initial=0.573,
  /num/abl_en_scalefull_s8090@ep1/parse/.initial=1.000,
  /num/abl_en_scalefull_s8090@ep1/n/.initial=300,
  /num/abl_en_scalefull_s8090@ep2/miou/.initial=0.787,
  /num/abl_en_scalefull_s8090@ep2/lo/.initial=0.772,
  /num/abl_en_scalefull_s8090@ep2/hi/.initial=0.802,
  /num/abl_en_scalefull_s8090@ep2/dir/.initial=0.603,
  /num/abl_en_scalefull_s8090@ep2/parse/.initial=1.000,
  /num/abl_en_scalefull_s8090@ep2/n/.initial=300,
  /num/abl_en_scalefull_s8090@ep3/miou/.initial=0.784,
  /num/abl_en_scalefull_s8090@ep3/lo/.initial=0.767,
  /num/abl_en_scalefull_s8090@ep3/hi/.initial=0.799,
  /num/abl_en_scalefull_s8090@ep3/dir/.initial=0.557,
  /num/abl_en_scalefull_s8090@ep3/parse/.initial=1.000,
  /num/abl_en_scalefull_s8090@ep3/n/.initial=300,
  /num/abl_en_scalefull_s8090@ep4/miou/.initial=0.790,
  /num/abl_en_scalefull_s8090@ep4/lo/.initial=0.774,
  /num/abl_en_scalefull_s8090@ep4/hi/.initial=0.806,
  /num/abl_en_scalefull_s8090@ep4/dir/.initial=0.580,
  /num/abl_en_scalefull_s8090@ep4/parse/.initial=1.000,
  /num/abl_en_scalefull_s8090@ep4/n/.initial=300,
  /num/abl_en_scalefull_s8090@ep5/miou/.initial=0.792,
  /num/abl_en_scalefull_s8090@ep5/lo/.initial=0.776,
  /num/abl_en_scalefull_s8090@ep5/hi/.initial=0.807,
  /num/abl_en_scalefull_s8090@ep5/dir/.initial=0.603,
  /num/abl_en_scalefull_s8090@ep5/parse/.initial=1.000,
  /num/abl_en_scalefull_s8090@ep5/n/.initial=300,
  /num/abl_final_r1to0/miou/.initial=0.782,
  /num/abl_final_r1to0/lo/.initial=0.765,
  /num/abl_final_r1to0/hi/.initial=0.799,
  /num/abl_final_r1to0/dir/.initial=0.540,
  /num/abl_final_r1to0/parse/.initial=1.000,
  /num/abl_final_r1to0/n/.initial=300,
  /num/abl_final_r1to0@s1875/miou/.initial=0.767,
  /num/abl_final_r1to0@s1875/lo/.initial=0.752,
  /num/abl_final_r1to0@s1875/hi/.initial=0.784,
  /num/abl_final_r1to0@s1875/dir/.initial=0.580,
  /num/abl_final_r1to0@s1875/parse/.initial=1.000,
  /num/abl_final_r1to0@s1875/n/.initial=300,
  /num/abl_final_r1to0@s3750/miou/.initial=0.766,
  /num/abl_final_r1to0@s3750/lo/.initial=0.750,
  /num/abl_final_r1to0@s3750/hi/.initial=0.782,
  /num/abl_final_r1to0@s3750/dir/.initial=0.547,
  /num/abl_final_r1to0@s3750/parse/.initial=1.000,
  /num/abl_final_r1to0@s3750/n/.initial=300,
  /num/abl_final_r1to0@s5625/miou/.initial=0.780,
  /num/abl_final_r1to0@s5625/lo/.initial=0.764,
  /num/abl_final_r1to0@s5625/hi/.initial=0.796,
  /num/abl_final_r1to0@s5625/dir/.initial=0.593,
  /num/abl_final_r1to0@s5625/parse/.initial=1.000,
  /num/abl_final_r1to0@s5625/n/.initial=300,
  /num/abl_final_r1to0@s7500/miou/.initial=0.781,
  /num/abl_final_r1to0@s7500/lo/.initial=0.765,
  /num/abl_final_r1to0@s7500/hi/.initial=0.797,
  /num/abl_final_r1to0@s7500/dir/.initial=0.567,
  /num/abl_final_r1to0@s7500/parse/.initial=1.000,
  /num/abl_final_r1to0@s7500/n/.initial=300,
  /num/abl_final_r1to0@s9375/miou/.initial=0.779,
  /num/abl_final_r1to0@s9375/lo/.initial=0.763,
  /num/abl_final_r1to0@s9375/hi/.initial=0.795,
  /num/abl_final_r1to0@s9375/dir/.initial=0.530,
  /num/abl_final_r1to0@s9375/parse/.initial=1.000,
  /num/abl_final_r1to0@s9375/n/.initial=300,
  /num/abl_final_r1to2/miou/.initial=0.792,
  /num/abl_final_r1to2/lo/.initial=0.775,
  /num/abl_final_r1to2/hi/.initial=0.808,
  /num/abl_final_r1to2/dir/.initial=0.590,
  /num/abl_final_r1to2/parse/.initial=1.000,
  /num/abl_final_r1to2/n/.initial=300,
  /num/abl_final_r1to2@s1875/miou/.initial=0.759,
  /num/abl_final_r1to2@s1875/lo/.initial=0.743,
  /num/abl_final_r1to2@s1875/hi/.initial=0.776,
  /num/abl_final_r1to2@s1875/dir/.initial=0.547,
  /num/abl_final_r1to2@s1875/parse/.initial=1.000,
  /num/abl_final_r1to2@s1875/n/.initial=300,
  /num/abl_final_r1to2@s3750/miou/.initial=0.785,
  /num/abl_final_r1to2@s3750/lo/.initial=0.769,
  /num/abl_final_r1to2@s3750/hi/.initial=0.800,
  /num/abl_final_r1to2@s3750/dir/.initial=0.577,
  /num/abl_final_r1to2@s3750/parse/.initial=1.000,
  /num/abl_final_r1to2@s3750/n/.initial=300,
  /num/abl_final_r1to2@s5625/miou/.initial=0.795,
  /num/abl_final_r1to2@s5625/lo/.initial=0.780,
  /num/abl_final_r1to2@s5625/hi/.initial=0.810,
  /num/abl_final_r1to2@s5625/dir/.initial=0.617,
  /num/abl_final_r1to2@s5625/parse/.initial=1.000,
  /num/abl_final_r1to2@s5625/n/.initial=300,
  /num/abl_final_r1to2@s7500/miou/.initial=0.798,
  /num/abl_final_r1to2@s7500/lo/.initial=0.783,
  /num/abl_final_r1to2@s7500/hi/.initial=0.813,
  /num/abl_final_r1to2@s7500/dir/.initial=0.620,
  /num/abl_final_r1to2@s7500/parse/.initial=1.000,
  /num/abl_final_r1to2@s7500/n/.initial=300,
  /num/abl_final_r1to2@s9375/miou/.initial=0.796,
  /num/abl_final_r1to2@s9375/lo/.initial=0.781,
  /num/abl_final_r1to2@s9375/hi/.initial=0.811,
  /num/abl_final_r1to2@s9375/dir/.initial=0.597,
  /num/abl_final_r1to2@s9375/parse/.initial=1.000,
  /num/abl_final_r1to2@s9375/n/.initial=300,
  /num/abl_final_rs4780/miou/.initial=0.784,
  /num/abl_final_rs4780/lo/.initial=0.769,
  /num/abl_final_rs4780/hi/.initial=0.800,
  /num/abl_final_rs4780/dir/.initial=0.570,
  /num/abl_final_rs4780/parse/.initial=1.000,
  /num/abl_final_rs4780/n/.initial=300,
  /num/abl_final_rs4780@s1875/miou/.initial=0.771,
  /num/abl_final_rs4780@s1875/lo/.initial=0.755,
  /num/abl_final_rs4780@s1875/hi/.initial=0.787,
  /num/abl_final_rs4780@s1875/dir/.initial=0.570,
  /num/abl_final_rs4780@s1875/parse/.initial=1.000,
  /num/abl_final_rs4780@s1875/n/.initial=300,
  /num/abl_final_rs4780@s3750/miou/.initial=0.772,
  /num/abl_final_rs4780@s3750/lo/.initial=0.755,
  /num/abl_final_rs4780@s3750/hi/.initial=0.788,
  /num/abl_final_rs4780@s3750/dir/.initial=0.533,
  /num/abl_final_rs4780@s3750/parse/.initial=1.000,
  /num/abl_final_rs4780@s3750/n/.initial=300,
  /num/abl_final_rs4780@s5625/miou/.initial=0.784,
  /num/abl_final_rs4780@s5625/lo/.initial=0.769,
  /num/abl_final_rs4780@s5625/hi/.initial=0.800,
  /num/abl_final_rs4780@s5625/dir/.initial=0.577,
  /num/abl_final_rs4780@s5625/parse/.initial=1.000,
  /num/abl_final_rs4780@s5625/n/.initial=300,
  /num/abl_final_rs4780@s7500/miou/.initial=0.788,
  /num/abl_final_rs4780@s7500/lo/.initial=0.773,
  /num/abl_final_rs4780@s7500/hi/.initial=0.804,
  /num/abl_final_rs4780@s7500/dir/.initial=0.577,
  /num/abl_final_rs4780@s7500/parse/.initial=1.000,
  /num/abl_final_rs4780@s7500/n/.initial=300,
  /num/abl_final_rs4780@s9375/miou/.initial=0.791,
  /num/abl_final_rs4780@s9375/lo/.initial=0.776,
  /num/abl_final_rs4780@s9375/hi/.initial=0.805,
  /num/abl_final_rs4780@s9375/dir/.initial=0.607,
  /num/abl_final_rs4780@s9375/parse/.initial=1.000,
  /num/abl_final_rs4780@s9375/n/.initial=300,
  /num/abl_reason5b3k@ep1/miou/.initial=0.718,
  /num/abl_reason5b3k@ep1/lo/.initial=0.701,
  /num/abl_reason5b3k@ep1/hi/.initial=0.735,
  /num/abl_reason5b3k@ep1/dir/.initial=0.460,
  /num/abl_reason5b3k@ep1/parse/.initial=1.000,
  /num/abl_reason5b3k@ep1/n/.initial=300,
  /num/abl_reason5b3k_mix@ep1/miou/.initial=0.709,
  /num/abl_reason5b3k_mix@ep1/lo/.initial=0.692,
  /num/abl_reason5b3k_mix@ep1/hi/.initial=0.726,
  /num/abl_reason5b3k_mix@ep1/dir/.initial=0.477,
  /num/abl_reason5b3k_mix@ep1/parse/.initial=1.000,
  /num/abl_reason5b3k_mix@ep1/n/.initial=300,
  /num/abl_reason5b3k_r1to0p5@ep1/miou/.initial=0.731,
  /num/abl_reason5b3k_r1to0p5@ep1/lo/.initial=0.714,
  /num/abl_reason5b3k_r1to0p5@ep1/hi/.initial=0.749,
  /num/abl_reason5b3k_r1to0p5@ep1/dir/.initial=0.510,
  /num/abl_reason5b3k_r1to0p5@ep1/parse/.initial=1.000,
  /num/abl_reason5b3k_r1to0p5@ep1/n/.initial=300,
  /num/abl_reason5b3k_r1to2@ep1/miou/.initial=0.751,
  /num/abl_reason5b3k_r1to2@ep1/lo/.initial=0.734,
  /num/abl_reason5b3k_r1to2@ep1/hi/.initial=0.768,
  /num/abl_reason5b3k_r1to2@ep1/dir/.initial=0.530,
  /num/abl_reason5b3k_r1to2@ep1/parse/.initial=1.000,
  /num/abl_reason5b3k_r1to2@ep1/n/.initial=300,
  /num/align_4b_reason2/miou/.initial=0.641,
  /num/align_4b_reason2/lo/.initial=0.624,
  /num/align_4b_reason2/hi/.initial=0.658,
  /num/align_4b_reason2/dir/.initial=0.270,
  /num/align_4b_reason2/parse/.initial=1.000,
  /num/align_4b_reason2/n/.initial=300,
  /num/align_direct_ep8/miou/.initial=0.821,
  /num/align_direct_ep8/lo/.initial=0.806,
  /num/align_direct_ep8/hi/.initial=0.836,
  /num/align_direct_ep8/dir/.initial=0.610,
  /num/align_direct_ep8/parse/.initial=1.000,
  /num/align_direct_ep8/n/.initial=300,
  /num/align_e36_ep4/miou/.initial=0.782,
  /num/align_e36_ep4/lo/.initial=0.765,
  /num/align_e36_ep4/hi/.initial=0.799,
  /num/align_e36_ep4/dir/.initial=0.553,
  /num/align_e36_ep4/parse/.initial=1.000,
  /num/align_e36_ep4/n/.initial=300,
  /num/axis2d15k/miou/.initial=0.717,
  /num/axis2d15k/lo/.initial=0.706,
  /num/axis2d15k/hi/.initial=0.729,
  /num/axis2d15k/dir/.initial=0.237,
  /num/axis2d15k/parse/.initial=1.000,
  /num/axis2d15k/n/.initial=300,
  /num/axis2d15k@ep1/miou/.initial=0.706,
  /num/axis2d15k@ep1/lo/.initial=0.696,
  /num/axis2d15k@ep1/hi/.initial=0.715,
  /num/axis2d15k@ep1/dir/.initial=0.307,
  /num/axis2d15k@ep1/parse/.initial=1.000,
  /num/axis2d15k@ep1/n/.initial=300,
  /num/axis2d15k@ep10/miou/.initial=0.706,
  /num/axis2d15k@ep10/lo/.initial=0.695,
  /num/axis2d15k@ep10/hi/.initial=0.717,
  /num/axis2d15k@ep10/dir/.initial=0.303,
  /num/axis2d15k@ep10/parse/.initial=0.997,
  /num/axis2d15k@ep10/n/.initial=300,
  /num/axis2d15k@ep2/miou/.initial=0.702,
  /num/axis2d15k@ep2/lo/.initial=0.693,
  /num/axis2d15k@ep2/hi/.initial=0.711,
  /num/axis2d15k@ep2/dir/.initial=0.307,
  /num/axis2d15k@ep2/parse/.initial=1.000,
  /num/axis2d15k@ep2/n/.initial=300,
  /num/axis2d15k@ep3/miou/.initial=0.710,
  /num/axis2d15k@ep3/lo/.initial=0.700,
  /num/axis2d15k@ep3/hi/.initial=0.721,
  /num/axis2d15k@ep3/dir/.initial=0.307,
  /num/axis2d15k@ep3/parse/.initial=1.000,
  /num/axis2d15k@ep3/n/.initial=300,
  /num/axis2d15k@ep4/miou/.initial=0.709,
  /num/axis2d15k@ep4/lo/.initial=0.699,
  /num/axis2d15k@ep4/hi/.initial=0.719,
  /num/axis2d15k@ep4/dir/.initial=0.307,
  /num/axis2d15k@ep4/parse/.initial=1.000,
  /num/axis2d15k@ep4/n/.initial=300,
  /num/axis2d15k@ep5/miou/.initial=0.709,
  /num/axis2d15k@ep5/lo/.initial=0.699,
  /num/axis2d15k@ep5/hi/.initial=0.719,
  /num/axis2d15k@ep5/dir/.initial=0.307,
  /num/axis2d15k@ep5/parse/.initial=1.000,
  /num/axis2d15k@ep5/n/.initial=300,
  /num/axis2d15k@ep6/miou/.initial=0.707,
  /num/axis2d15k@ep6/lo/.initial=0.697,
  /num/axis2d15k@ep6/hi/.initial=0.717,
  /num/axis2d15k@ep6/dir/.initial=0.307,
  /num/axis2d15k@ep6/parse/.initial=1.000,
  /num/axis2d15k@ep6/n/.initial=300,
  /num/axis2d15k@ep7/miou/.initial=0.706,
  /num/axis2d15k@ep7/lo/.initial=0.694,
  /num/axis2d15k@ep7/hi/.initial=0.717,
  /num/axis2d15k@ep7/dir/.initial=0.303,
  /num/axis2d15k@ep7/parse/.initial=0.997,
  /num/axis2d15k@ep7/n/.initial=300,
  /num/axis2d15k@ep8/miou/.initial=0.708,
  /num/axis2d15k@ep8/lo/.initial=0.698,
  /num/axis2d15k@ep8/hi/.initial=0.718,
  /num/axis2d15k@ep8/dir/.initial=0.307,
  /num/axis2d15k@ep8/parse/.initial=1.000,
  /num/axis2d15k@ep8/n/.initial=300,
  /num/axis2d15k@ep9/miou/.initial=0.706,
  /num/axis2d15k@ep9/lo/.initial=0.695,
  /num/axis2d15k@ep9/hi/.initial=0.717,
  /num/axis2d15k@ep9/dir/.initial=0.303,
  /num/axis2d15k@ep9/parse/.initial=0.997,
  /num/axis2d15k@ep9/n/.initial=300,
  /num/claude-haiku_icl9/miou/.initial=0.688,
  /num/claude-haiku_icl9/lo/.initial=0.676,
  /num/claude-haiku_icl9/hi/.initial=0.701,
  /num/claude-haiku_icl9/dir/.initial=0.343,
  /num/claude-haiku_icl9/parse/.initial=1.000,
  /num/claude-haiku_icl9/n/.initial=300,
  /num/claude-haiku_icl9txt/miou/.initial=0.691,
  /num/claude-haiku_icl9txt/lo/.initial=0.678,
  /num/claude-haiku_icl9txt/hi/.initial=0.704,
  /num/claude-haiku_icl9txt/dir/.initial=0.397,
  /num/claude-haiku_icl9txt/parse/.initial=1.000,
  /num/claude-haiku_icl9txt/n/.initial=300,
  /num/claude-haiku_zs/miou/.initial=0.690,
  /num/claude-haiku_zs/lo/.initial=0.678,
  /num/claude-haiku_zs/hi/.initial=0.702,
  /num/claude-haiku_zs/dir/.initial=0.407,
  /num/claude-haiku_zs/parse/.initial=1.000,
  /num/claude-haiku_zs/n/.initial=300,
  /num/claude-opus_icl9/miou/.initial=0.748,
  /num/claude-opus_icl9/lo/.initial=0.737,
  /num/claude-opus_icl9/hi/.initial=0.760,
  /num/claude-opus_icl9/dir/.initial=0.523,
  /num/claude-opus_icl9/parse/.initial=1.000,
  /num/claude-opus_icl9/n/.initial=300,
  /num/claude-opus_icl9_reason/miou/.initial=0.751,
  /num/claude-opus_icl9_reason/lo/.initial=0.739,
  /num/claude-opus_icl9_reason/hi/.initial=0.762,
  /num/claude-opus_icl9_reason/dir/.initial=0.523,
  /num/claude-opus_icl9_reason/parse/.initial=1.000,
  /num/claude-opus_icl9_reason/n/.initial=300,
  /num/claude-opus_icl9_set1/miou/.initial=0.747,
  /num/claude-opus_icl9_set1/lo/.initial=0.735,
  /num/claude-opus_icl9_set1/hi/.initial=0.758,
  /num/claude-opus_icl9_set1/dir/.initial=0.550,
  /num/claude-opus_icl9_set1/parse/.initial=1.000,
  /num/claude-opus_icl9_set1/n/.initial=300,
  /num/claude-opus_icl9_set2/miou/.initial=0.747,
  /num/claude-opus_icl9_set2/lo/.initial=0.735,
  /num/claude-opus_icl9_set2/hi/.initial=0.759,
  /num/claude-opus_icl9_set2/dir/.initial=0.510,
  /num/claude-opus_icl9_set2/parse/.initial=1.000,
  /num/claude-opus_icl9_set2/n/.initial=300,
  /num/claude-opus_icl9txt/miou/.initial=0.742,
  /num/claude-opus_icl9txt/lo/.initial=0.730,
  /num/claude-opus_icl9txt/hi/.initial=0.753,
  /num/claude-opus_icl9txt/dir/.initial=0.543,
  /num/claude-opus_icl9txt/parse/.initial=1.000,
  /num/claude-opus_icl9txt/n/.initial=300,
  /num/claude-opus_zs/miou/.initial=0.724,
  /num/claude-opus_zs/lo/.initial=0.712,
  /num/claude-opus_zs/hi/.initial=0.736,
  /num/claude-opus_zs/dir/.initial=0.520,
  /num/claude-opus_zs/parse/.initial=1.000,
  /num/claude-opus_zs/n/.initial=300,
  /num/claude-sonnet_icl9/miou/.initial=0.730,
  /num/claude-sonnet_icl9/lo/.initial=0.718,
  /num/claude-sonnet_icl9/hi/.initial=0.742,
  /num/claude-sonnet_icl9/dir/.initial=0.433,
  /num/claude-sonnet_icl9/parse/.initial=1.000,
  /num/claude-sonnet_icl9/n/.initial=300,
  /num/claude-sonnet_icl9txt/miou/.initial=0.728,
  /num/claude-sonnet_icl9txt/lo/.initial=0.717,
  /num/claude-sonnet_icl9txt/hi/.initial=0.740,
  /num/claude-sonnet_icl9txt/dir/.initial=0.430,
  /num/claude-sonnet_icl9txt/parse/.initial=1.000,
  /num/claude-sonnet_icl9txt/n/.initial=300,
  /num/claude-sonnet_zs/miou/.initial=0.727,
  /num/claude-sonnet_zs/lo/.initial=0.715,
  /num/claude-sonnet_zs/hi/.initial=0.738,
  /num/claude-sonnet_zs/dir/.initial=0.447,
  /num/claude-sonnet_zs/parse/.initial=1.000,
  /num/claude-sonnet_zs/n/.initial=300,
  /num/clip_mlp_s0/miou/.initial=0.712,
  /num/clip_mlp_s0/lo/.initial=0.701,
  /num/clip_mlp_s0/hi/.initial=0.722,
  /num/clip_mlp_s0/dir/.initial=0.380,
  /num/clip_mlp_s0/parse/.initial=1.000,
  /num/clip_mlp_s0/n/.initial=300,
  /num/clip_mlp_s1/miou/.initial=0.716,
  /num/clip_mlp_s1/lo/.initial=0.706,
  /num/clip_mlp_s1/hi/.initial=0.727,
  /num/clip_mlp_s1/dir/.initial=0.373,
  /num/clip_mlp_s1/parse/.initial=1.000,
  /num/clip_mlp_s1/n/.initial=300,
  /num/clip_mlp_s2/miou/.initial=0.710,
  /num/clip_mlp_s2/lo/.initial=0.700,
  /num/clip_mlp_s2/hi/.initial=0.720,
  /num/clip_mlp_s2/dir/.initial=0.317,
  /num/clip_mlp_s2/parse/.initial=1.000,
  /num/clip_mlp_s2/n/.initial=300,
  /num/cnn_s0/miou/.initial=0.712,
  /num/cnn_s0/lo/.initial=0.700,
  /num/cnn_s0/hi/.initial=0.723,
  /num/cnn_s0/dir/.initial=0.400,
  /num/cnn_s0/parse/.initial=1.000,
  /num/cnn_s0/n/.initial=300,
  /num/cnn_s1/miou/.initial=0.709,
  /num/cnn_s1/lo/.initial=0.697,
  /num/cnn_s1/hi/.initial=0.720,
  /num/cnn_s1/dir/.initial=0.350,
  /num/cnn_s1/parse/.initial=1.000,
  /num/cnn_s1/n/.initial=300,
  /num/cnn_s2/miou/.initial=0.708,
  /num/cnn_s2/lo/.initial=0.698,
  /num/cnn_s2/hi/.initial=0.719,
  /num/cnn_s2/dir/.initial=0.400,
  /num/cnn_s2/parse/.initial=1.000,
  /num/cnn_s2/n/.initial=300,
  /num/constant/miou/.initial=0.692,
  /num/constant/lo/.initial=0.690,
  /num/constant/hi/.initial=0.695,
  /num/constant/dir/.initial=0.295,
  /num/constant/parse/.initial=1.000,
  /num/constant/n/.initial=3807,
  /num/dino_cls_s0/miou/.initial=0.722,
  /num/dino_cls_s0/lo/.initial=0.708,
  /num/dino_cls_s0/hi/.initial=0.735,
  /num/dino_cls_s0/dir/.initial=0.437,
  /num/dino_cls_s0/parse/.initial=1.000,
  /num/dino_cls_s0/n/.initial=300,
  /num/dino_cls_s1/miou/.initial=0.727,
  /num/dino_cls_s1/lo/.initial=0.713,
  /num/dino_cls_s1/hi/.initial=0.740,
  /num/dino_cls_s1/dir/.initial=0.417,
  /num/dino_cls_s1/parse/.initial=1.000,
  /num/dino_cls_s1/n/.initial=300,
  /num/dino_cls_s2/miou/.initial=0.726,
  /num/dino_cls_s2/lo/.initial=0.713,
  /num/dino_cls_s2/hi/.initial=0.739,
  /num/dino_cls_s2/dir/.initial=0.420,
  /num/dino_cls_s2/parse/.initial=1.000,
  /num/dino_cls_s2/n/.initial=300,
  /num/dino_grid3_s0/miou/.initial=0.781,
  /num/dino_grid3_s0/lo/.initial=0.769,
  /num/dino_grid3_s0/hi/.initial=0.793,
  /num/dino_grid3_s0/dir/.initial=0.630,
  /num/dino_grid3_s0/parse/.initial=1.000,
  /num/dino_grid3_s0/n/.initial=300,
  /num/dino_grid3_s1/miou/.initial=0.787,
  /num/dino_grid3_s1/lo/.initial=0.775,
  /num/dino_grid3_s1/hi/.initial=0.799,
  /num/dino_grid3_s1/dir/.initial=0.640,
  /num/dino_grid3_s1/parse/.initial=1.000,
  /num/dino_grid3_s1/n/.initial=300,
  /num/dino_grid3_s2/miou/.initial=0.781,
  /num/dino_grid3_s2/lo/.initial=0.768,
  /num/dino_grid3_s2/hi/.initial=0.793,
  /num/dino_grid3_s2/dir/.initial=0.597,
  /num/dino_grid3_s2/parse/.initial=1.000,
  /num/dino_grid3_s2/n/.initial=300,
  /num/dino_mean_s0/miou/.initial=0.774,
  /num/dino_mean_s0/lo/.initial=0.761,
  /num/dino_mean_s0/hi/.initial=0.786,
  /num/dino_mean_s0/dir/.initial=0.543,
  /num/dino_mean_s0/parse/.initial=1.000,
  /num/dino_mean_s0/n/.initial=300,
  /num/dino_mean_s1/miou/.initial=0.774,
  /num/dino_mean_s1/lo/.initial=0.761,
  /num/dino_mean_s1/hi/.initial=0.787,
  /num/dino_mean_s1/dir/.initial=0.587,
  /num/dino_mean_s1/parse/.initial=1.000,
  /num/dino_mean_s1/n/.initial=300,
  /num/dino_mean_s2/miou/.initial=0.772,
  /num/dino_mean_s2/lo/.initial=0.760,
  /num/dino_mean_s2/hi/.initial=0.784,
  /num/dino_mean_s2/dir/.initial=0.547,
  /num/dino_mean_s2/parse/.initial=1.000,
  /num/dino_mean_s2/n/.initial=300,
  /num/dino_mlp_s0/miou/.initial=0.726,
  /num/dino_mlp_s0/lo/.initial=0.715,
  /num/dino_mlp_s0/hi/.initial=0.738,
  /num/dino_mlp_s0/dir/.initial=0.397,
  /num/dino_mlp_s0/parse/.initial=1.000,
  /num/dino_mlp_s0/n/.initial=300,
  /num/dino_mlp_s1/miou/.initial=0.721,
  /num/dino_mlp_s1/lo/.initial=0.709,
  /num/dino_mlp_s1/hi/.initial=0.732,
  /num/dino_mlp_s1/dir/.initial=0.437,
  /num/dino_mlp_s1/parse/.initial=1.000,
  /num/dino_mlp_s1/n/.initial=300,
  /num/dino_mlp_s2/miou/.initial=0.723,
  /num/dino_mlp_s2/lo/.initial=0.712,
  /num/dino_mlp_s2/hi/.initial=0.733,
  /num/dino_mlp_s2/dir/.initial=0.440,
  /num/dino_mlp_s2/parse/.initial=1.000,
  /num/dino_mlp_s2/n/.initial=300,
  /num/direct15k/miou/.initial=0.789,
  /num/direct15k/lo/.initial=0.775,
  /num/direct15k/hi/.initial=0.804,
  /num/direct15k/dir/.initial=0.587,
  /num/direct15k/parse/.initial=1.000,
  /num/direct15k/n/.initial=300,
  /num/direct15k@ep1/miou/.initial=0.771,
  /num/direct15k@ep1/lo/.initial=0.754,
  /num/direct15k@ep1/hi/.initial=0.787,
  /num/direct15k@ep1/dir/.initial=0.577,
  /num/direct15k@ep1/parse/.initial=1.000,
  /num/direct15k@ep1/n/.initial=300,
  /num/direct15k@ep10/miou/.initial=0.807,
  /num/direct15k@ep10/lo/.initial=0.794,
  /num/direct15k@ep10/hi/.initial=0.822,
  /num/direct15k@ep10/dir/.initial=0.607,
  /num/direct15k@ep10/parse/.initial=1.000,
  /num/direct15k@ep10/n/.initial=300,
  /num/direct15k@ep2/miou/.initial=0.789,
  /num/direct15k@ep2/lo/.initial=0.773,
  /num/direct15k@ep2/hi/.initial=0.805,
  /num/direct15k@ep2/dir/.initial=0.617,
  /num/direct15k@ep2/parse/.initial=1.000,
  /num/direct15k@ep2/n/.initial=300,
  /num/direct15k@ep3/miou/.initial=0.798,
  /num/direct15k@ep3/lo/.initial=0.783,
  /num/direct15k@ep3/hi/.initial=0.813,
  /num/direct15k@ep3/dir/.initial=0.597,
  /num/direct15k@ep3/parse/.initial=1.000,
  /num/direct15k@ep3/n/.initial=300,
  /num/direct15k@ep4/miou/.initial=0.798,
  /num/direct15k@ep4/lo/.initial=0.783,
  /num/direct15k@ep4/hi/.initial=0.814,
  /num/direct15k@ep4/dir/.initial=0.583,
  /num/direct15k@ep4/parse/.initial=1.000,
  /num/direct15k@ep4/n/.initial=300,
  /num/direct15k@ep5/miou/.initial=0.809,
  /num/direct15k@ep5/lo/.initial=0.795,
  /num/direct15k@ep5/hi/.initial=0.824,
  /num/direct15k@ep5/dir/.initial=0.603,
  /num/direct15k@ep5/parse/.initial=1.000,
  /num/direct15k@ep5/n/.initial=300,
  /num/direct15k@ep6/miou/.initial=0.805,
  /num/direct15k@ep6/lo/.initial=0.791,
  /num/direct15k@ep6/hi/.initial=0.819,
  /num/direct15k@ep6/dir/.initial=0.597,
  /num/direct15k@ep6/parse/.initial=1.000,
  /num/direct15k@ep6/n/.initial=300,
  /num/direct15k@ep7/miou/.initial=0.805,
  /num/direct15k@ep7/lo/.initial=0.791,
  /num/direct15k@ep7/hi/.initial=0.819,
  /num/direct15k@ep7/dir/.initial=0.607,
  /num/direct15k@ep7/parse/.initial=1.000,
  /num/direct15k@ep7/n/.initial=300,
  /num/direct15k@ep8/miou/.initial=0.805,
  /num/direct15k@ep8/lo/.initial=0.792,
  /num/direct15k@ep8/hi/.initial=0.820,
  /num/direct15k@ep8/dir/.initial=0.600,
  /num/direct15k@ep8/parse/.initial=1.000,
  /num/direct15k@ep8/n/.initial=300,
  /num/direct15k@ep9/miou/.initial=0.807,
  /num/direct15k@ep9/lo/.initial=0.794,
  /num/direct15k@ep9/hi/.initial=0.822,
  /num/direct15k@ep9/dir/.initial=0.613,
  /num/direct15k@ep9/parse/.initial=1.000,
  /num/direct15k@ep9/n/.initial=300,
  /num/direct15k_balanced/miou/.initial=0.716,
  /num/direct15k_balanced/lo/.initial=0.706,
  /num/direct15k_balanced/hi/.initial=0.728,
  /num/direct15k_balanced/dir/.initial=0.237,
  /num/direct15k_balanced/parse/.initial=1.000,
  /num/direct15k_balanced/n/.initial=300,
  /num/direct15k_bin30/miou/.initial=0.731,
  /num/direct15k_bin30/lo/.initial=0.717,
  /num/direct15k_bin30/hi/.initial=0.745,
  /num/direct15k_bin30/dir/.initial=0.553,
  /num/direct15k_bin30/parse/.initial=1.000,
  /num/direct15k_bin30/n/.initial=300,
  /num/direct15k_fixmag/miou/.initial=0.744,
  /num/direct15k_fixmag/lo/.initial=0.730,
  /num/direct15k_fixmag/hi/.initial=0.757,
  /num/direct15k_fixmag/dir/.initial=0.583,
  /num/direct15k_fixmag/parse/.initial=1.000,
  /num/direct15k_fixmag/n/.initial=300,
  /num/direct15k_q10/miou/.initial=0.775,
  /num/direct15k_q10/lo/.initial=0.761,
  /num/direct15k_q10/hi/.initial=0.789,
  /num/direct15k_q10/dir/.initial=0.560,
  /num/direct15k_q10/parse/.initial=1.000,
  /num/direct15k_q10/n/.initial=300,
  /num/direct15k_q25/miou/.initial=0.736,
  /num/direct15k_q25/lo/.initial=0.723,
  /num/direct15k_q25/hi/.initial=0.749,
  /num/direct15k_q25/dir/.initial=0.563,
  /num/direct15k_q25/parse/.initial=1.000,
  /num/direct15k_q25/n/.initial=300,
  /num/direct15k_seed1/miou/.initial=0.812,
  /num/direct15k_seed1/lo/.initial=0.798,
  /num/direct15k_seed1/hi/.initial=0.827,
  /num/direct15k_seed1/dir/.initial=0.610,
  /num/direct15k_seed1/parse/.initial=1.000,
  /num/direct15k_seed1/n/.initial=300,
  /num/direct15k_seed1@ep1/miou/.initial=0.775,
  /num/direct15k_seed1@ep1/lo/.initial=0.758,
  /num/direct15k_seed1@ep1/hi/.initial=0.793,
  /num/direct15k_seed1@ep1/dir/.initial=0.533,
  /num/direct15k_seed1@ep1/parse/.initial=1.000,
  /num/direct15k_seed1@ep1/n/.initial=300,
  /num/direct15k_seed1@ep10/miou/.initial=0.816,
  /num/direct15k_seed1@ep10/lo/.initial=0.803,
  /num/direct15k_seed1@ep10/hi/.initial=0.831,
  /num/direct15k_seed1@ep10/dir/.initial=0.623,
  /num/direct15k_seed1@ep10/parse/.initial=1.000,
  /num/direct15k_seed1@ep10/n/.initial=300,
  /num/direct15k_seed1@ep2/miou/.initial=0.796,
  /num/direct15k_seed1@ep2/lo/.initial=0.780,
  /num/direct15k_seed1@ep2/hi/.initial=0.812,
  /num/direct15k_seed1@ep2/dir/.initial=0.640,
  /num/direct15k_seed1@ep2/parse/.initial=1.000,
  /num/direct15k_seed1@ep2/n/.initial=300,
  /num/direct15k_seed1@ep3/miou/.initial=0.803,
  /num/direct15k_seed1@ep3/lo/.initial=0.787,
  /num/direct15k_seed1@ep3/hi/.initial=0.820,
  /num/direct15k_seed1@ep3/dir/.initial=0.617,
  /num/direct15k_seed1@ep3/parse/.initial=1.000,
  /num/direct15k_seed1@ep3/n/.initial=300,
  /num/direct15k_seed1@ep4/miou/.initial=0.805,
  /num/direct15k_seed1@ep4/lo/.initial=0.789,
  /num/direct15k_seed1@ep4/hi/.initial=0.821,
  /num/direct15k_seed1@ep4/dir/.initial=0.620,
  /num/direct15k_seed1@ep4/parse/.initial=1.000,
  /num/direct15k_seed1@ep4/n/.initial=300,
  /num/direct15k_seed1@ep5/miou/.initial=0.806,
  /num/direct15k_seed1@ep5/lo/.initial=0.792,
  /num/direct15k_seed1@ep5/hi/.initial=0.820,
  /num/direct15k_seed1@ep5/dir/.initial=0.610,
  /num/direct15k_seed1@ep5/parse/.initial=1.000,
  /num/direct15k_seed1@ep5/n/.initial=300,
  /num/direct15k_seed1@ep6/miou/.initial=0.814,
  /num/direct15k_seed1@ep6/lo/.initial=0.800,
  /num/direct15k_seed1@ep6/hi/.initial=0.827,
  /num/direct15k_seed1@ep6/dir/.initial=0.623,
  /num/direct15k_seed1@ep6/parse/.initial=1.000,
  /num/direct15k_seed1@ep6/n/.initial=300,
  /num/direct15k_seed1@ep7/miou/.initial=0.813,
  /num/direct15k_seed1@ep7/lo/.initial=0.799,
  /num/direct15k_seed1@ep7/hi/.initial=0.827,
  /num/direct15k_seed1@ep7/dir/.initial=0.623,
  /num/direct15k_seed1@ep7/parse/.initial=1.000,
  /num/direct15k_seed1@ep7/n/.initial=300,
  /num/direct15k_seed1@ep8/miou/.initial=0.817,
  /num/direct15k_seed1@ep8/lo/.initial=0.804,
  /num/direct15k_seed1@ep8/hi/.initial=0.831,
  /num/direct15k_seed1@ep8/dir/.initial=0.613,
  /num/direct15k_seed1@ep8/parse/.initial=1.000,
  /num/direct15k_seed1@ep8/n/.initial=300,
  /num/direct15k_seed1@ep9/miou/.initial=0.815,
  /num/direct15k_seed1@ep9/lo/.initial=0.802,
  /num/direct15k_seed1@ep9/hi/.initial=0.830,
  /num/direct15k_seed1@ep9/dir/.initial=0.613,
  /num/direct15k_seed1@ep9/parse/.initial=1.000,
  /num/direct15k_seed1@ep9/n/.initial=300,
  /num/direct15k_seed2/miou/.initial=0.800,
  /num/direct15k_seed2/lo/.initial=0.786,
  /num/direct15k_seed2/hi/.initial=0.816,
  /num/direct15k_seed2/dir/.initial=0.573,
  /num/direct15k_seed2/parse/.initial=1.000,
  /num/direct15k_seed2/n/.initial=300,
  /num/direct15k_seed2@ep1/miou/.initial=0.769,
  /num/direct15k_seed2@ep1/lo/.initial=0.753,
  /num/direct15k_seed2@ep1/hi/.initial=0.785,
  /num/direct15k_seed2@ep1/dir/.initial=0.527,
  /num/direct15k_seed2@ep1/parse/.initial=1.000,
  /num/direct15k_seed2@ep1/n/.initial=300,
  /num/direct15k_seed2@ep10/miou/.initial=0.812,
  /num/direct15k_seed2@ep10/lo/.initial=0.797,
  /num/direct15k_seed2@ep10/hi/.initial=0.826,
  /num/direct15k_seed2@ep10/dir/.initial=0.633,
  /num/direct15k_seed2@ep10/parse/.initial=1.000,
  /num/direct15k_seed2@ep10/n/.initial=300,
  /num/direct15k_seed2@ep2/miou/.initial=0.787,
  /num/direct15k_seed2@ep2/lo/.initial=0.772,
  /num/direct15k_seed2@ep2/hi/.initial=0.803,
  /num/direct15k_seed2@ep2/dir/.initial=0.577,
  /num/direct15k_seed2@ep2/parse/.initial=1.000,
  /num/direct15k_seed2@ep2/n/.initial=300,
  /num/direct15k_seed2@ep3/miou/.initial=0.791,
  /num/direct15k_seed2@ep3/lo/.initial=0.776,
  /num/direct15k_seed2@ep3/hi/.initial=0.807,
  /num/direct15k_seed2@ep3/dir/.initial=0.607,
  /num/direct15k_seed2@ep3/parse/.initial=1.000,
  /num/direct15k_seed2@ep3/n/.initial=300,
  /num/direct15k_seed2@ep4/miou/.initial=0.810,
  /num/direct15k_seed2@ep4/lo/.initial=0.795,
  /num/direct15k_seed2@ep4/hi/.initial=0.824,
  /num/direct15k_seed2@ep4/dir/.initial=0.627,
  /num/direct15k_seed2@ep4/parse/.initial=1.000,
  /num/direct15k_seed2@ep4/n/.initial=300,
  /num/direct15k_seed2@ep5/miou/.initial=0.811,
  /num/direct15k_seed2@ep5/lo/.initial=0.797,
  /num/direct15k_seed2@ep5/hi/.initial=0.826,
  /num/direct15k_seed2@ep5/dir/.initial=0.650,
  /num/direct15k_seed2@ep5/parse/.initial=1.000,
  /num/direct15k_seed2@ep5/n/.initial=300,
  /num/direct15k_seed2@ep6/miou/.initial=0.814,
  /num/direct15k_seed2@ep6/lo/.initial=0.800,
  /num/direct15k_seed2@ep6/hi/.initial=0.828,
  /num/direct15k_seed2@ep6/dir/.initial=0.640,
  /num/direct15k_seed2@ep6/parse/.initial=1.000,
  /num/direct15k_seed2@ep6/n/.initial=300,
  /num/direct15k_seed2@ep7/miou/.initial=0.815,
  /num/direct15k_seed2@ep7/lo/.initial=0.802,
  /num/direct15k_seed2@ep7/hi/.initial=0.829,
  /num/direct15k_seed2@ep7/dir/.initial=0.637,
  /num/direct15k_seed2@ep7/parse/.initial=1.000,
  /num/direct15k_seed2@ep7/n/.initial=300,
  /num/direct15k_seed2@ep8/miou/.initial=0.813,
  /num/direct15k_seed2@ep8/lo/.initial=0.800,
  /num/direct15k_seed2@ep8/hi/.initial=0.827,
  /num/direct15k_seed2@ep8/dir/.initial=0.610,
  /num/direct15k_seed2@ep8/parse/.initial=1.000,
  /num/direct15k_seed2@ep8/n/.initial=300,
  /num/direct15k_seed2@ep9/miou/.initial=0.812,
  /num/direct15k_seed2@ep9/lo/.initial=0.798,
  /num/direct15k_seed2@ep9/hi/.initial=0.826,
  /num/direct15k_seed2@ep9/dir/.initial=0.623,
  /num/direct15k_seed2@ep9/parse/.initial=1.000,
  /num/direct15k_seed2@ep9/n/.initial=300,
  /num/direct15k_symarea/miou/.initial=0.709,
  /num/direct15k_symarea/lo/.initial=0.697,
  /num/direct15k_symarea/hi/.initial=0.721,
  /num/direct15k_symarea/dir/.initial=0.237,
  /num/direct15k_symarea/parse/.initial=1.000,
  /num/direct15k_symarea/n/.initial=300,
  /num/direct21k/miou/.initial=0.812,
  /num/direct21k/lo/.initial=0.808,
  /num/direct21k/hi/.initial=0.816,
  /num/direct21k/dir/.initial=0.598,
  /num/direct21k/parse/.initial=1.000,
  /num/direct21k/n/.initial=3807,
  /num/direct21k@ep1/miou/.initial=0.793,
  /num/direct21k@ep1/lo/.initial=0.777,
  /num/direct21k@ep1/hi/.initial=0.809,
  /num/direct21k@ep1/dir/.initial=0.600,
  /num/direct21k@ep1/parse/.initial=1.000,
  /num/direct21k@ep1/n/.initial=300,
  /num/direct21k@ep10/miou/.initial=0.821,
  /num/direct21k@ep10/lo/.initial=0.807,
  /num/direct21k@ep10/hi/.initial=0.835,
  /num/direct21k@ep10/dir/.initial=0.637,
  /num/direct21k@ep10/parse/.initial=1.000,
  /num/direct21k@ep10/n/.initial=300,
  /num/direct21k@ep2/miou/.initial=0.800,
  /num/direct21k@ep2/lo/.initial=0.783,
  /num/direct21k@ep2/hi/.initial=0.816,
  /num/direct21k@ep2/dir/.initial=0.600,
  /num/direct21k@ep2/parse/.initial=1.000,
  /num/direct21k@ep2/n/.initial=300,
  /num/direct21k@ep3/miou/.initial=0.793,
  /num/direct21k@ep3/lo/.initial=0.776,
  /num/direct21k@ep3/hi/.initial=0.810,
  /num/direct21k@ep3/dir/.initial=0.600,
  /num/direct21k@ep3/parse/.initial=1.000,
  /num/direct21k@ep3/n/.initial=300,
  /num/direct21k@ep4/miou/.initial=0.804,
  /num/direct21k@ep4/lo/.initial=0.789,
  /num/direct21k@ep4/hi/.initial=0.820,
  /num/direct21k@ep4/dir/.initial=0.613,
  /num/direct21k@ep4/parse/.initial=1.000,
  /num/direct21k@ep4/n/.initial=300,
  /num/direct21k@ep5/miou/.initial=0.825,
  /num/direct21k@ep5/lo/.initial=0.810,
  /num/direct21k@ep5/hi/.initial=0.839,
  /num/direct21k@ep5/dir/.initial=0.647,
  /num/direct21k@ep5/parse/.initial=1.000,
  /num/direct21k@ep5/n/.initial=300,
  /num/direct21k@ep6/miou/.initial=0.818,
  /num/direct21k@ep6/lo/.initial=0.805,
  /num/direct21k@ep6/hi/.initial=0.832,
  /num/direct21k@ep6/dir/.initial=0.643,
  /num/direct21k@ep6/parse/.initial=1.000,
  /num/direct21k@ep6/n/.initial=300,
  /num/direct21k@ep7/miou/.initial=0.825,
  /num/direct21k@ep7/lo/.initial=0.811,
  /num/direct21k@ep7/hi/.initial=0.838,
  /num/direct21k@ep7/dir/.initial=0.640,
  /num/direct21k@ep7/parse/.initial=1.000,
  /num/direct21k@ep7/n/.initial=300,
  /num/direct21k@ep8/miou/.initial=0.825,
  /num/direct21k@ep8/lo/.initial=0.811,
  /num/direct21k@ep8/hi/.initial=0.839,
  /num/direct21k@ep8/dir/.initial=0.650,
  /num/direct21k@ep8/parse/.initial=1.000,
  /num/direct21k@ep8/n/.initial=300,
  /num/direct21k@ep9/miou/.initial=0.823,
  /num/direct21k@ep9/lo/.initial=0.809,
  /num/direct21k@ep9/hi/.initial=0.836,
  /num/direct21k@ep9/dir/.initial=0.637,
  /num/direct21k@ep9/parse/.initial=1.000,
  /num/direct21k@ep9/n/.initial=300,
  /num/direct21k_bin30/miou/.initial=0.753,
  /num/direct21k_bin30/lo/.initial=0.749,
  /num/direct21k_bin30/hi/.initial=0.757,
  /num/direct21k_bin30/dir/.initial=0.560,
  /num/direct21k_bin30/parse/.initial=1.000,
  /num/direct21k_bin30/n/.initial=3807,
  /num/direct21k_q10/miou/.initial=0.796,
  /num/direct21k_q10/lo/.initial=0.792,
  /num/direct21k_q10/hi/.initial=0.799,
  /num/direct21k_q10/dir/.initial=0.582,
  /num/direct21k_q10/parse/.initial=1.000,
  /num/direct21k_q10/n/.initial=3807,
  /num/direct21k_q25/miou/.initial=0.752,
  /num/direct21k_q25/lo/.initial=0.749,
  /num/direct21k_q25/hi/.initial=0.756,
  /num/direct21k_q25/dir/.initial=0.561,
  /num/direct21k_q25/parse/.initial=1.000,
  /num/direct21k_q25/n/.initial=3807,
  /num/direct21k_symarea/miou/.initial=0.712,
  /num/direct21k_symarea/lo/.initial=0.709,
  /num/direct21k_symarea/hi/.initial=0.715,
  /num/direct21k_symarea/dir/.initial=0.256,
  /num/direct21k_symarea/parse/.initial=1.000,
  /num/direct21k_symarea/n/.initial=3807,
  /num/direct21k_symmax/miou/.initial=0.651,
  /num/direct21k_symmax/lo/.initial=0.647,
  /num/direct21k_symmax/hi/.initial=0.655,
  /num/direct21k_symmax/dir/.initial=0.256,
  /num/direct21k_symmax/parse/.initial=1.000,
  /num/direct21k_symmax/n/.initial=3807,
  /num/direct21k_symmean/miou/.initial=0.711,
  /num/direct21k_symmean/lo/.initial=0.708,
  /num/direct21k_symmean/hi/.initial=0.715,
  /num/direct21k_symmean/dir/.initial=0.256,
  /num/direct21k_symmean/parse/.initial=1.000,
  /num/direct21k_symmean/n/.initial=3807,
  /num/gaic/miou/.initial=0.397,
  /num/gaic/lo/.initial=0.394,
  /num/gaic/hi/.initial=0.400,
  /num/gaic/dir/.initial=0.336,
  /num/gaic/parse/.initial=1.000,
  /num/gaic/n/.initial=3807,
  /num/gpt-4o_icl9/miou/.initial=0.676,
  /num/gpt-4o_icl9/lo/.initial=0.663,
  /num/gpt-4o_icl9/hi/.initial=0.689,
  /num/gpt-4o_icl9/dir/.initial=0.233,
  /num/gpt-4o_icl9/parse/.initial=1.000,
  /num/gpt-4o_icl9/n/.initial=300,
  /num/gpt-4o_zs/miou/.initial=0.672,
  /num/gpt-4o_zs/lo/.initial=0.660,
  /num/gpt-4o_zs/hi/.initial=0.684,
  /num/gpt-4o_zs/dir/.initial=0.280,
  /num/gpt-4o_zs/parse/.initial=1.000,
  /num/gpt-4o_zs/n/.initial=300,
  /num/gpt-5.5_icl9/miou/.initial=0.713,
  /num/gpt-5.5_icl9/lo/.initial=0.701,
  /num/gpt-5.5_icl9/hi/.initial=0.725,
  /num/gpt-5.5_icl9/dir/.initial=0.443,
  /num/gpt-5.5_icl9/parse/.initial=1.000,
  /num/gpt-5.5_icl9/n/.initial=300,
  /num/gpt-5.5_zs/miou/.initial=0.700,
  /num/gpt-5.5_zs/lo/.initial=0.688,
  /num/gpt-5.5_zs/hi/.initial=0.711,
  /num/gpt-5.5_zs/dir/.initial=0.397,
  /num/gpt-5.5_zs/parse/.initial=1.000,
  /num/gpt-5.5_zs/n/.initial=300,
  /num/gpt-6-astra_icl9/miou/.initial=0.734,
  /num/gpt-6-astra_icl9/lo/.initial=0.723,
  /num/gpt-6-astra_icl9/hi/.initial=0.746,
  /num/gpt-6-astra_icl9/dir/.initial=0.533,
  /num/gpt-6-astra_icl9/parse/.initial=1.000,
  /num/gpt-6-astra_icl9/n/.initial=300,
  /num/gpt-6-astra_zs/miou/.initial=0.707,
  /num/gpt-6-astra_zs/lo/.initial=0.694,
  /num/gpt-6-astra_zs/hi/.initial=0.719,
  /num/gpt-6-astra_zs/dir/.initial=0.490,
  /num/gpt-6-astra_zs/parse/.initial=1.000,
  /num/gpt-6-astra_zs/n/.initial=300,
  /num/noop/miou/.initial=0.587,
  /num/noop/lo/.initial=0.584,
  /num/noop/hi/.initial=0.589,
  /num/noop/dir/.initial=0.256,
  /num/noop/parse/.initial=1.000,
  /num/noop/n/.initial=3807,
  /num/ours_e36/miou/.initial=0.778,
  /num/ours_e36/lo/.initial=0.773,
  /num/ours_e36/hi/.initial=0.782,
  /num/ours_e36/dir/.initial=0.551,
  /num/ours_e36/parse/.initial=1.000,
  /num/ours_e36/n/.initial=3807,
  /num/ours_e36@ep1/miou/.initial=0.738,
  /num/ours_e36@ep1/lo/.initial=0.720,
  /num/ours_e36@ep1/hi/.initial=0.755,
  /num/ours_e36@ep1/dir/.initial=0.523,
  /num/ours_e36@ep1/parse/.initial=1.000,
  /num/ours_e36@ep1/n/.initial=300,
  /num/ours_e36@ep2/miou/.initial=0.767,
  /num/ours_e36@ep2/lo/.initial=0.751,
  /num/ours_e36@ep2/hi/.initial=0.783,
  /num/ours_e36@ep2/dir/.initial=0.537,
  /num/ours_e36@ep2/parse/.initial=1.000,
  /num/ours_e36@ep2/n/.initial=300,
  /num/ours_e36@ep3/miou/.initial=0.782,
  /num/ours_e36@ep3/lo/.initial=0.766,
  /num/ours_e36@ep3/hi/.initial=0.798,
  /num/ours_e36@ep3/dir/.initial=0.563,
  /num/ours_e36@ep3/parse/.initial=1.000,
  /num/ours_e36@ep3/n/.initial=300,
  /num/ours_e36@ep4/miou/.initial=0.784,
  /num/ours_e36@ep4/lo/.initial=0.768,
  /num/ours_e36@ep4/hi/.initial=0.800,
  /num/ours_e36@ep4/dir/.initial=0.570,
  /num/ours_e36@ep4/parse/.initial=1.000,
  /num/ours_e36@ep4/n/.initial=300,
  /num/ours_e36@ep5/miou/.initial=0.782,
  /num/ours_e36@ep5/lo/.initial=0.766,
  /num/ours_e36@ep5/hi/.initial=0.797,
  /num/ours_e36@ep5/dir/.initial=0.573,
  /num/ours_e36@ep5/parse/.initial=1.000,
  /num/ours_e36@ep5/n/.initial=300,
  /num/ours_e36_symarea/miou/.initial=0.710,
  /num/ours_e36_symarea/lo/.initial=0.707,
  /num/ours_e36_symarea/hi/.initial=0.713,
  /num/ours_e36_symarea/dir/.initial=0.256,
  /num/ours_e36_symarea/parse/.initial=1.000,
  /num/ours_e36_symarea/n/.initial=3807,
  /num/ours_e36_symmax/miou/.initial=0.636,
  /num/ours_e36_symmax/lo/.initial=0.633,
  /num/ours_e36_symmax/hi/.initial=0.639,
  /num/ours_e36_symmax/dir/.initial=0.256,
  /num/ours_e36_symmax/parse/.initial=1.000,
  /num/ours_e36_symmax/n/.initial=3807,
  /num/ours_e36_symmean/miou/.initial=0.710,
  /num/ours_e36_symmean/lo/.initial=0.707,
  /num/ours_e36_symmean/hi/.initial=0.713,
  /num/ours_e36_symmean/dir/.initial=0.256,
  /num/ours_e36_symmean/parse/.initial=1.000,
  /num/ours_e36_symmean/n/.initial=3807,
  /num/ours_en/miou/.initial=0.784,
  /num/ours_en/lo/.initial=0.780,
  /num/ours_en/hi/.initial=0.789,
  /num/ours_en/dir/.initial=0.574,
  /num/ours_en/parse/.initial=1.000,
  /num/ours_en/n/.initial=3807,
  /num/ours_en@ep1/miou/.initial=0.761,
  /num/ours_en@ep1/lo/.initial=0.743,
  /num/ours_en@ep1/hi/.initial=0.778,
  /num/ours_en@ep1/dir/.initial=0.543,
  /num/ours_en@ep1/parse/.initial=1.000,
  /num/ours_en@ep1/n/.initial=300,
  /num/ours_en@ep2/miou/.initial=0.780,
  /num/ours_en@ep2/lo/.initial=0.764,
  /num/ours_en@ep2/hi/.initial=0.797,
  /num/ours_en@ep2/dir/.initial=0.547,
  /num/ours_en@ep2/parse/.initial=1.000,
  /num/ours_en@ep2/n/.initial=300,
  /num/ours_en@ep3/miou/.initial=0.802,
  /num/ours_en@ep3/lo/.initial=0.786,
  /num/ours_en@ep3/hi/.initial=0.818,
  /num/ours_en@ep3/dir/.initial=0.613,
  /num/ours_en@ep3/parse/.initial=1.000,
  /num/ours_en@ep3/n/.initial=300,
  /num/ours_en@ep4/miou/.initial=0.794,
  /num/ours_en@ep4/lo/.initial=0.778,
  /num/ours_en@ep4/hi/.initial=0.809,
  /num/ours_en@ep4/dir/.initial=0.633,
  /num/ours_en@ep4/parse/.initial=1.000,
  /num/ours_en@ep4/n/.initial=300,
  /num/ours_en@ep5/miou/.initial=0.796,
  /num/ours_en@ep5/lo/.initial=0.780,
  /num/ours_en@ep5/hi/.initial=0.810,
  /num/ours_en@ep5/dir/.initial=0.597,
  /num/ours_en@ep5/parse/.initial=1.000,
  /num/ours_en@ep5/n/.initial=300,
  /num/ours_en_bin30/miou/.initial=0.739,
  /num/ours_en_bin30/lo/.initial=0.734,
  /num/ours_en_bin30/hi/.initial=0.743,
  /num/ours_en_bin30/dir/.initial=0.576,
  /num/ours_en_bin30/parse/.initial=1.000,
  /num/ours_en_bin30/n/.initial=3807,
  /num/ours_en_prejudge/miou/.initial=0.785,
  /num/ours_en_prejudge/lo/.initial=0.769,
  /num/ours_en_prejudge/hi/.initial=0.801,
  /num/ours_en_prejudge/dir/.initial=0.547,
  /num/ours_en_prejudge/parse/.initial=1.000,
  /num/ours_en_prejudge/n/.initial=300,
  /num/ours_en_prejudge@ep1/miou/.initial=0.779,
  /num/ours_en_prejudge@ep1/lo/.initial=0.762,
  /num/ours_en_prejudge@ep1/hi/.initial=0.795,
  /num/ours_en_prejudge@ep1/dir/.initial=0.580,
  /num/ours_en_prejudge@ep1/parse/.initial=1.000,
  /num/ours_en_prejudge@ep1/n/.initial=300,
  /num/ours_en_prejudge@ep2/miou/.initial=0.794,
  /num/ours_en_prejudge@ep2/lo/.initial=0.777,
  /num/ours_en_prejudge@ep2/hi/.initial=0.811,
  /num/ours_en_prejudge@ep2/dir/.initial=0.610,
  /num/ours_en_prejudge@ep2/parse/.initial=1.000,
  /num/ours_en_prejudge@ep2/n/.initial=300,
  /num/ours_en_prejudge@ep3/miou/.initial=0.791,
  /num/ours_en_prejudge@ep3/lo/.initial=0.773,
  /num/ours_en_prejudge@ep3/hi/.initial=0.808,
  /num/ours_en_prejudge@ep3/dir/.initial=0.600,
  /num/ours_en_prejudge@ep3/parse/.initial=1.000,
  /num/ours_en_prejudge@ep3/n/.initial=300,
  /num/ours_en_prejudge@ep4/miou/.initial=0.800,
  /num/ours_en_prejudge@ep4/lo/.initial=0.782,
  /num/ours_en_prejudge@ep4/hi/.initial=0.816,
  /num/ours_en_prejudge@ep4/dir/.initial=0.627,
  /num/ours_en_prejudge@ep4/parse/.initial=1.000,
  /num/ours_en_prejudge@ep4/n/.initial=300,
  /num/ours_en_prejudge@ep5/miou/.initial=0.804,
  /num/ours_en_prejudge@ep5/lo/.initial=0.787,
  /num/ours_en_prejudge@ep5/hi/.initial=0.820,
  /num/ours_en_prejudge@ep5/dir/.initial=0.630,
  /num/ours_en_prejudge@ep5/parse/.initial=1.000,
  /num/ours_en_prejudge@ep5/n/.initial=300,
  /num/ours_en_q25/miou/.initial=0.724,
  /num/ours_en_q25/lo/.initial=0.720,
  /num/ours_en_q25/hi/.initial=0.728,
  /num/ours_en_q25/dir/.initial=0.576,
  /num/ours_en_q25/parse/.initial=1.000,
  /num/ours_en_q25/n/.initial=3807,
  /num/ours_en_symarea/miou/.initial=0.709,
  /num/ours_en_symarea/lo/.initial=0.706,
  /num/ours_en_symarea/hi/.initial=0.712,
  /num/ours_en_symarea/dir/.initial=0.256,
  /num/ours_en_symarea/parse/.initial=1.000,
  /num/ours_en_symarea/n/.initial=3807,
  /num/ours_final/miou/.initial=0.797,
  /num/ours_final/lo/.initial=0.781,
  /num/ours_final/hi/.initial=0.814,
  /num/ours_final/dir/.initial=0.607,
  /num/ours_final/parse/.initial=1.000,
  /num/ours_final/n/.initial=300,
  /num/ours_final@ep1/miou/.initial=0.765,
  /num/ours_final@ep1/lo/.initial=0.748,
  /num/ours_final@ep1/hi/.initial=0.781,
  /num/ours_final@ep1/dir/.initial=0.593,
  /num/ours_final@ep1/parse/.initial=1.000,
  /num/ours_final@ep1/n/.initial=300,
  /num/ours_final@ep2/miou/.initial=0.773,
  /num/ours_final@ep2/lo/.initial=0.758,
  /num/ours_final@ep2/hi/.initial=0.788,
  /num/ours_final@ep2/dir/.initial=0.560,
  /num/ours_final@ep2/parse/.initial=1.000,
  /num/ours_final@ep2/n/.initial=300,
  /num/ours_final@ep3/miou/.initial=0.776,
  /num/ours_final@ep3/lo/.initial=0.760,
  /num/ours_final@ep3/hi/.initial=0.792,
  /num/ours_final@ep3/dir/.initial=0.570,
  /num/ours_final@ep3/parse/.initial=1.000,
  /num/ours_final@ep3/n/.initial=300,
  /num/ours_final@ep4/miou/.initial=0.787,
  /num/ours_final@ep4/lo/.initial=0.772,
  /num/ours_final@ep4/hi/.initial=0.803,
  /num/ours_final@ep4/dir/.initial=0.587,
  /num/ours_final@ep4/parse/.initial=1.000,
  /num/ours_final@ep4/n/.initial=300,
  /num/ours_final@ep5/miou/.initial=0.785,
  /num/ours_final@ep5/lo/.initial=0.770,
  /num/ours_final@ep5/hi/.initial=0.801,
  /num/ours_final@ep5/dir/.initial=0.570,
  /num/ours_final@ep5/parse/.initial=1.000,
  /num/ours_final@ep5/n/.initial=300,
  /num/ours_final_balanced/miou/.initial=0.718,
  /num/ours_final_balanced/lo/.initial=0.707,
  /num/ours_final_balanced/hi/.initial=0.730,
  /num/ours_final_balanced/dir/.initial=0.237,
  /num/ours_final_balanced/parse/.initial=1.000,
  /num/ours_final_balanced/n/.initial=300,
  /num/ours_final_bin30/miou/.initial=0.743,
  /num/ours_final_bin30/lo/.initial=0.728,
  /num/ours_final_bin30/hi/.initial=0.757,
  /num/ours_final_bin30/dir/.initial=0.603,
  /num/ours_final_bin30/parse/.initial=1.000,
  /num/ours_final_bin30/n/.initial=300,
  /num/ours_final_fixmag/miou/.initial=0.747,
  /num/ours_final_fixmag/lo/.initial=0.731,
  /num/ours_final_fixmag/hi/.initial=0.763,
  /num/ours_final_fixmag/dir/.initial=0.603,
  /num/ours_final_fixmag/parse/.initial=1.000,
  /num/ours_final_fixmag/n/.initial=300,
  /num/ours_final_q10/miou/.initial=0.778,
  /num/ours_final_q10/lo/.initial=0.762,
  /num/ours_final_q10/hi/.initial=0.794,
  /num/ours_final_q10/dir/.initial=0.610,
  /num/ours_final_q10/parse/.initial=1.000,
  /num/ours_final_q10/n/.initial=300,
  /num/ours_final_q25/miou/.initial=0.731,
  /num/ours_final_q25/lo/.initial=0.717,
  /num/ours_final_q25/hi/.initial=0.746,
  /num/ours_final_q25/dir/.initial=0.613,
  /num/ours_final_q25/parse/.initial=1.000,
  /num/ours_final_q25/n/.initial=300,
  /num/ours_final_seed1/miou/.initial=0.789,
  /num/ours_final_seed1/lo/.initial=0.773,
  /num/ours_final_seed1/hi/.initial=0.805,
  /num/ours_final_seed1/dir/.initial=0.583,
  /num/ours_final_seed1/parse/.initial=1.000,
  /num/ours_final_seed1/n/.initial=300,
  /num/ours_final_seed1@ep1/miou/.initial=0.757,
  /num/ours_final_seed1@ep1/lo/.initial=0.740,
  /num/ours_final_seed1@ep1/hi/.initial=0.773,
  /num/ours_final_seed1@ep1/dir/.initial=0.567,
  /num/ours_final_seed1@ep1/parse/.initial=1.000,
  /num/ours_final_seed1@ep1/n/.initial=300,
  /num/ours_final_seed1@ep2/miou/.initial=0.775,
  /num/ours_final_seed1@ep2/lo/.initial=0.758,
  /num/ours_final_seed1@ep2/hi/.initial=0.792,
  /num/ours_final_seed1@ep2/dir/.initial=0.580,
  /num/ours_final_seed1@ep2/parse/.initial=1.000,
  /num/ours_final_seed1@ep2/n/.initial=300,
  /num/ours_final_seed1@ep3/miou/.initial=0.784,
  /num/ours_final_seed1@ep3/lo/.initial=0.768,
  /num/ours_final_seed1@ep3/hi/.initial=0.800,
  /num/ours_final_seed1@ep3/dir/.initial=0.597,
  /num/ours_final_seed1@ep3/parse/.initial=1.000,
  /num/ours_final_seed1@ep3/n/.initial=300,
  /num/ours_final_seed1@ep4/miou/.initial=0.796,
  /num/ours_final_seed1@ep4/lo/.initial=0.780,
  /num/ours_final_seed1@ep4/hi/.initial=0.811,
  /num/ours_final_seed1@ep4/dir/.initial=0.597,
  /num/ours_final_seed1@ep4/parse/.initial=1.000,
  /num/ours_final_seed1@ep4/n/.initial=300,
  /num/ours_final_seed1@ep5/miou/.initial=0.796,
  /num/ours_final_seed1@ep5/lo/.initial=0.781,
  /num/ours_final_seed1@ep5/hi/.initial=0.812,
  /num/ours_final_seed1@ep5/dir/.initial=0.603,
  /num/ours_final_seed1@ep5/parse/.initial=1.000,
  /num/ours_final_seed1@ep5/n/.initial=300,
  /num/ours_final_seed2/miou/.initial=0.783,
  /num/ours_final_seed2/lo/.initial=0.766,
  /num/ours_final_seed2/hi/.initial=0.800,
  /num/ours_final_seed2/dir/.initial=0.557,
  /num/ours_final_seed2/parse/.initial=1.000,
  /num/ours_final_seed2/n/.initial=300,
  /num/ours_final_seed2@ep1/miou/.initial=0.769,
  /num/ours_final_seed2@ep1/lo/.initial=0.753,
  /num/ours_final_seed2@ep1/hi/.initial=0.786,
  /num/ours_final_seed2@ep1/dir/.initial=0.590,
  /num/ours_final_seed2@ep1/parse/.initial=1.000,
  /num/ours_final_seed2@ep1/n/.initial=300,
  /num/ours_final_seed2@ep2/miou/.initial=0.775,
  /num/ours_final_seed2@ep2/lo/.initial=0.759,
  /num/ours_final_seed2@ep2/hi/.initial=0.791,
  /num/ours_final_seed2@ep2/dir/.initial=0.580,
  /num/ours_final_seed2@ep2/parse/.initial=1.000,
  /num/ours_final_seed2@ep2/n/.initial=300,
  /num/ours_final_seed2@ep3/miou/.initial=0.786,
  /num/ours_final_seed2@ep3/lo/.initial=0.770,
  /num/ours_final_seed2@ep3/hi/.initial=0.802,
  /num/ours_final_seed2@ep3/dir/.initial=0.620,
  /num/ours_final_seed2@ep3/parse/.initial=1.000,
  /num/ours_final_seed2@ep3/n/.initial=300,
  /num/ours_final_seed2@ep4/miou/.initial=0.791,
  /num/ours_final_seed2@ep4/lo/.initial=0.775,
  /num/ours_final_seed2@ep4/hi/.initial=0.807,
  /num/ours_final_seed2@ep4/dir/.initial=0.613,
  /num/ours_final_seed2@ep4/parse/.initial=1.000,
  /num/ours_final_seed2@ep4/n/.initial=300,
  /num/ours_final_seed2@ep5/miou/.initial=0.795,
  /num/ours_final_seed2@ep5/lo/.initial=0.779,
  /num/ours_final_seed2@ep5/hi/.initial=0.811,
  /num/ours_final_seed2@ep5/dir/.initial=0.630,
  /num/ours_final_seed2@ep5/parse/.initial=1.000,
  /num/ours_final_seed2@ep5/n/.initial=300,
  /num/ours_final_symarea/miou/.initial=0.713,
  /num/ours_final_symarea/lo/.initial=0.700,
  /num/ours_final_symarea/hi/.initial=0.725,
  /num/ours_final_symarea/dir/.initial=0.237,
  /num/ours_final_symarea/parse/.initial=1.000,
  /num/ours_final_symarea/n/.initial=300,
  /num/qwen4b_zs/miou/.initial=0.548,
  /num/qwen4b_zs/lo/.initial=0.542,
  /num/qwen4b_zs/hi/.initial=0.554,
  /num/qwen4b_zs/dir/.initial=0.236,
  /num/qwen4b_zs/parse/.initial=1.000,
  /num/qwen4b_zs/n/.initial=3807,
  /num/rule_cal_blend/miou/.initial=0.686,
  /num/rule_cal_blend/lo/.initial=0.682,
  /num/rule_cal_blend/hi/.initial=0.689,
  /num/rule_cal_blend/dir/.initial=0.489,
  /num/rule_cal_blend/parse/.initial=1.000,
  /num/rule_cal_blend/n/.initial=3807,
  /num/rule_cal_subj/miou/.initial=0.669,
  /num/rule_cal_subj/lo/.initial=0.665,
  /num/rule_cal_subj/hi/.initial=0.672,
  /num/rule_cal_subj/dir/.initial=0.460,
  /num/rule_cal_subj/parse/.initial=1.000,
  /num/rule_cal_subj/n/.initial=3807,
  /num/rule_center_blend/miou/.initial=0.688,
  /num/rule_center_blend/lo/.initial=0.685,
  /num/rule_center_blend/hi/.initial=0.691,
  /num/rule_center_blend/dir/.initial=0.504,
  /num/rule_center_blend/parse/.initial=1.000,
  /num/rule_center_blend/n/.initial=3807,
  /num/rule_center_fix/miou/.initial=0.686,
  /num/rule_center_fix/lo/.initial=0.682,
  /num/rule_center_fix/hi/.initial=0.689,
  /num/rule_center_fix/dir/.initial=0.481,
  /num/rule_center_fix/parse/.initial=1.000,
  /num/rule_center_fix/n/.initial=3807,
  /num/rule_center_subj/miou/.initial=0.661,
  /num/rule_center_subj/lo/.initial=0.657,
  /num/rule_center_subj/hi/.initial=0.665,
  /num/rule_center_subj/dir/.initial=0.454,
  /num/rule_center_subj/parse/.initial=1.000,
  /num/rule_center_subj/n/.initial=3807,
  /num/rule_hierarchy/miou/.initial=0.690,
  /num/rule_hierarchy/lo/.initial=0.686,
  /num/rule_hierarchy/hi/.initial=0.694,
  /num/rule_hierarchy/dir/.initial=0.419,
  /num/rule_hierarchy/parse/.initial=1.000,
  /num/rule_hierarchy/n/.initial=3807,
  /num/rule_saliency/miou/.initial=0.690,
  /num/rule_saliency/lo/.initial=0.686,
  /num/rule_saliency/hi/.initial=0.694,
  /num/rule_saliency/dir/.initial=0.419,
  /num/rule_saliency/parse/.initial=1.000,
  /num/rule_saliency/n/.initial=3807,
  /num/rule_thirds_blend/miou/.initial=0.571,
  /num/rule_thirds_blend/lo/.initial=0.568,
  /num/rule_thirds_blend/hi/.initial=0.575,
  /num/rule_thirds_blend/dir/.initial=0.198,
  /num/rule_thirds_blend/parse/.initial=1.000,
  /num/rule_thirds_blend/n/.initial=3807,
  /num/rule_thirds_subj/miou/.initial=0.586,
  /num/rule_thirds_subj/lo/.initial=0.582,
  /num/rule_thirds_subj/hi/.initial=0.590,
  /num/rule_thirds_subj/dir/.initial=0.265,
  /num/rule_thirds_subj/parse/.initial=1.000,
  /num/rule_thirds_subj/n/.initial=3807,
  /num/sym1d15k/miou/.initial=0.709,
  /num/sym1d15k/lo/.initial=0.697,
  /num/sym1d15k/hi/.initial=0.722,
  /num/sym1d15k/dir/.initial=0.237,
  /num/sym1d15k/parse/.initial=1.000,
  /num/sym1d15k/n/.initial=300,
  /num/sym1d15k@ep1/miou/.initial=0.703,
  /num/sym1d15k@ep1/lo/.initial=0.693,
  /num/sym1d15k@ep1/hi/.initial=0.713,
  /num/sym1d15k@ep1/dir/.initial=0.307,
  /num/sym1d15k@ep1/parse/.initial=1.000,
  /num/sym1d15k@ep1/n/.initial=300,
  /num/sym1d15k@ep10/miou/.initial=0.703,
  /num/sym1d15k@ep10/lo/.initial=0.692,
  /num/sym1d15k@ep10/hi/.initial=0.713,
  /num/sym1d15k@ep10/dir/.initial=0.307,
  /num/sym1d15k@ep10/parse/.initial=1.000,
  /num/sym1d15k@ep10/n/.initial=300,
  /num/sym1d15k@ep2/miou/.initial=0.704,
  /num/sym1d15k@ep2/lo/.initial=0.694,
  /num/sym1d15k@ep2/hi/.initial=0.714,
  /num/sym1d15k@ep2/dir/.initial=0.307,
  /num/sym1d15k@ep2/parse/.initial=1.000,
  /num/sym1d15k@ep2/n/.initial=300,
  /num/sym1d15k@ep3/miou/.initial=0.704,
  /num/sym1d15k@ep3/lo/.initial=0.694,
  /num/sym1d15k@ep3/hi/.initial=0.715,
  /num/sym1d15k@ep3/dir/.initial=0.307,
  /num/sym1d15k@ep3/parse/.initial=1.000,
  /num/sym1d15k@ep3/n/.initial=300,
  /num/sym1d15k@ep4/miou/.initial=0.705,
  /num/sym1d15k@ep4/lo/.initial=0.695,
  /num/sym1d15k@ep4/hi/.initial=0.716,
  /num/sym1d15k@ep4/dir/.initial=0.307,
  /num/sym1d15k@ep4/parse/.initial=1.000,
  /num/sym1d15k@ep4/n/.initial=300,
  /num/sym1d15k@ep5/miou/.initial=0.705,
  /num/sym1d15k@ep5/lo/.initial=0.694,
  /num/sym1d15k@ep5/hi/.initial=0.715,
  /num/sym1d15k@ep5/dir/.initial=0.307,
  /num/sym1d15k@ep5/parse/.initial=1.000,
  /num/sym1d15k@ep5/n/.initial=300,
  /num/sym1d15k@ep6/miou/.initial=0.703,
  /num/sym1d15k@ep6/lo/.initial=0.693,
  /num/sym1d15k@ep6/hi/.initial=0.713,
  /num/sym1d15k@ep6/dir/.initial=0.307,
  /num/sym1d15k@ep6/parse/.initial=1.000,
  /num/sym1d15k@ep6/n/.initial=300,
  /num/sym1d15k@ep7/miou/.initial=0.703,
  /num/sym1d15k@ep7/lo/.initial=0.693,
  /num/sym1d15k@ep7/hi/.initial=0.713,
  /num/sym1d15k@ep7/dir/.initial=0.307,
  /num/sym1d15k@ep7/parse/.initial=1.000,
  /num/sym1d15k@ep7/n/.initial=300,
  /num/sym1d15k@ep8/miou/.initial=0.703,
  /num/sym1d15k@ep8/lo/.initial=0.693,
  /num/sym1d15k@ep8/hi/.initial=0.713,
  /num/sym1d15k@ep8/dir/.initial=0.307,
  /num/sym1d15k@ep8/parse/.initial=1.000,
  /num/sym1d15k@ep8/n/.initial=300,
  /num/sym1d15k@ep9/miou/.initial=0.702,
  /num/sym1d15k@ep9/lo/.initial=0.692,
  /num/sym1d15k@ep9/hi/.initial=0.713,
  /num/sym1d15k@ep9/dir/.initial=0.307,
  /num/sym1d15k@ep9/parse/.initial=1.000,
  /num/sym1d15k@ep9/n/.initial=300,
  /num/unic/miou/.initial=0.586,
  /num/unic/lo/.initial=0.583,
  /num/unic/hi/.initial=0.589,
  /num/unic/dir/.initial=0.455,
  /num/unic/parse/.initial=1.000,
  /num/unic/n/.initial=3807,
  /num/unic_clip0/miou/.initial=0.635,
  /num/unic_clip0/lo/.initial=0.632,
  /num/unic_clip0/hi/.initial=0.638,
  /num/unic_clip0/dir/.initial=0.434,
  /num/unic_clip0/parse/.initial=1.000,
  /num/unic_clip0/n/.initial=3807,
}

\newcommand{\res}[2]{\pgfkeysvalueof{/num/#1/#2}}

\newcommand{\best}[1]{\textbf{#1}}
\newcommand{\second}[1]{\underline{#1}}

\ifarxiv  \fi
\newcommand{\R}[1]{{\textbf{%
  \ifstrequal{#1}{1}{\textcolor{red}{R#1}}{%
  \ifstrequal{#1}{2}{\textcolor{blue}{R#1}}{%
  \ifstrequal{#1}{3}{\textcolor{magenta}{R#1}}{%
  \ifstrequal{#1}{4}{\textcolor{teal}{R#1}}{\textcolor{cyan}{R#1}}}}}}}}

\makeatletter
\renewcommand{\paragraph}{\@startsection{paragraph}{4}{\z@}{-0.5em}{-0.5em}{\normalfont\normalsize\bfseries}}
\makeatother

\usepackage{xr}
\makeatletter
\newcommand*{\addFileDependency}[1]{\typeout{(#1)}\@addtofilelist{#1}\IfFileExists{#1}{}{\typeout{No file #1.}}}
\makeatother

\definecolor{cvprblue}{rgb}{0.21,0.49,0.74}
\usepackage[pagebackref,breaklinks,colorlinks,allcolors=cvprblue]{hyperref}
\usepackage[capitalize]{cleveref}
\crefname{section}{Sec.}{Secs.}
\crefname{table}{Table}{Tables}
\crefname{figure}{Fig.}{Figs.}
\crefname{equation}{Eq.}{Eqs.}
\ifarxiv \crefname{appendix}{App.}{Apps.} \else \crefname{appendix}{Suppl.}{Suppls.} \fi
\input{suppl_labels}

\usepackage{placeins} %
\usepackage{colortbl}
\definecolor{lightred}{RGB}{253,191,191}
\definecolor{lightorange}{RGB}{255,223,191}
\definecolor{lightyellow}{RGB}{254,240,198}
\colorlet{rankfirst}{lightred}
\colorlet{ranksecond}{lightorange}
\colorlet{rankthird}{lightyellow}
\newcommand{\rankA}{\cellcolor{rankfirst}}
\newcommand{\rankB}{\cellcolor{ranksecond}}
\newcommand{\rankC}{\cellcolor{rankthird}}
\newcommand{\rankkey}[2]{{\setlength{\fboxsep}{1pt}\colorbox{#1}{\strut #2}}}
\begin{document}
\title{\paperTitle}
\author{\authorBlock}
\twocolumn[{%
\renewcommand\twocolumn[1][]{#1}%
\maketitle
\begin{center}
\centering
\captionsetup{type=figure}\vspace*{-14pt}
\includegraphics[width=\textwidth]{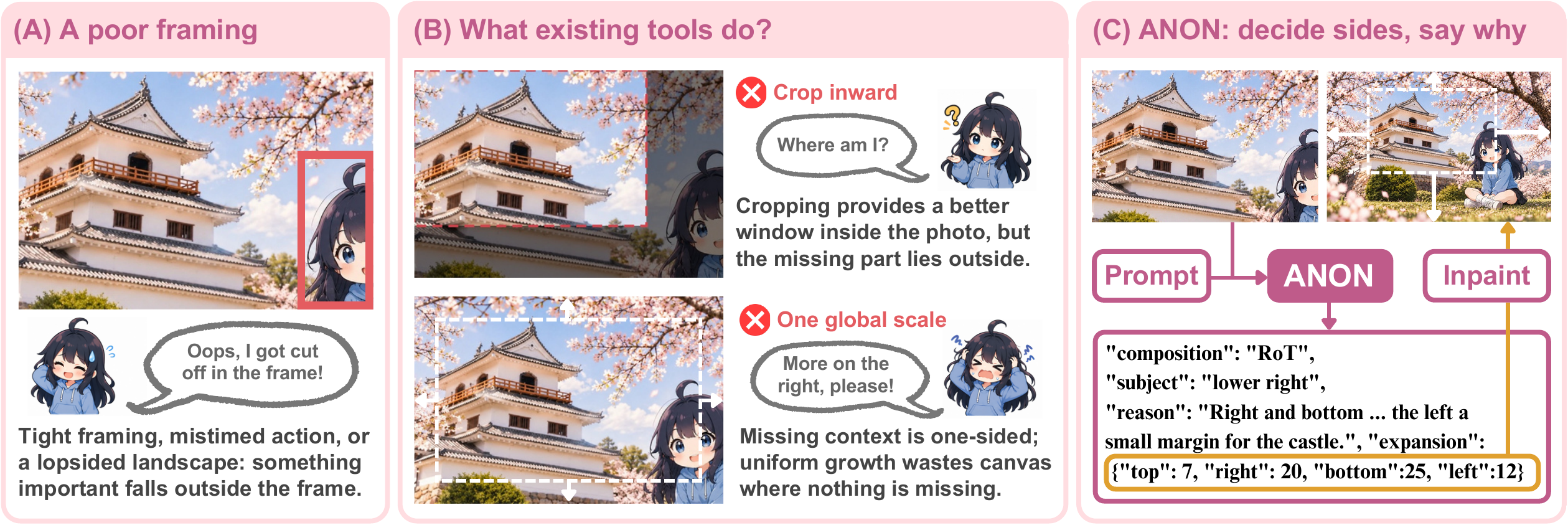} %
\vspace{-15pt}
\captionof{figure}{\textbf{The problem and our answer.} (A) A poorly framed photo cuts the subject at the border. (B) Existing tools: cropping inward keeps the missing part outside, and one global scale spends canvas where nothing is missing. (C) \anon{} reads the photo and returns a rationale and one expansion ratio per side; an inpainting model fills the enlarged canvas. Schematic example: the ratios and the filled result are illustrative; real outputs are shown in \cref{fig:qual,fig:apps}.} %
\label{fig:teaser}
\end{center}
\vspace{0pt} %
}]

\begin{abstract}
Outpainting models decide what to paint, but the sides to extend and the amount on each side are still set by the user. We present \anon{} (\anonlong), a fine-tuned vision-language model that predicts four outward expansion percentages and a short rationale from a badly framed photo. It decides the asymmetric expansion in one step, and an interchangeable outpainter fills the new area. On 300 test crops, it reaches a canvas mIoU of \res{ours_final@test300_seed0}{miou}, against \res{claude-opus_icl9@test300_seed0}{miou} for the strongest closed-source frontier VLM with in-context examples and \res{constant@test300_seed0}{miou} for a constant expansion; its numbers-only variant reaches \res{direct15k@test300_seed0}{miou}. In a user study, 74 participants prefer its frames to the frontier VLM's in 59\% of trials. The numbers follow directional edits to the rationale on 82--91\% of crops, while rewording leaves them unchanged; 90.6\% of its verifiable claims are judged correct.
\end{abstract}

\section{Introduction}
\label{sec:intro}

Photographic composition is decided by where the subject sits, whether the border cuts it, and how much room surrounds it. Tight or off-center framing can cut subjects, leave little space in the direction they face, or crowd one side. Cropping inward discards content; generative outpainting~\cite{rombach2022ldm} can expand the canvas, but the direction and extent of expansion must still be specified. Poor choices can give realistic fills without improving composition. Commercial tools such as Pixel Auto Frame (Pixel-AF) on Google Pixel phones already apply generative expansion to everyday photos, so the framing decision is made automatically in a shipped product.
We study how to decide how far to expand each side of an existing photo, and how to explain that decision. Automatic cropping selects a window inside the image: GAIC scores candidate boxes~\cite{zeng2020gaic}, and CROP uses a VLM to analyze composition before choosing a box~\cite{dong2026crop}. Outward cropping predicts one enlargement level shared by all borders, outpaints the ring, then searches for a crop~\cite{zhong2021outward}; UNIC recommends a camera move or zoom that requires re-shooting~\cite{liu2023unic}. None of them outputs a per-side outward expansion for an existing photo together with an explanation. Even with nine in-context examples, the strongest commercial VLM trails our planner by 0.049 in canvas mIoU (\cref{sec:results}). Reliable composition repair also requires explanations grounded in the image and consistent with the proposed expansion.

We propose \anon{} (\anonlong), a planner that predicts per-side expansion ratios together with a short rationale (\cref{fig:teaser}). It then hands the enlarged canvas to an outpainter. Reverse-cropped complete photos provide the geometric labels. Each rationale is written by two isolated teacher sessions: one observes the crop without the answer, and the other explains the answer without seeing the image (\cref{sec:rationale}). %
Our contributions are: %
\begin{enumerate}[leftmargin=*,nosep]
  \item \textbf{A new task: the expansion decision.} We formulate one-step prediction of four outward percentages for an existing photo, evaluated against the photographer's own frame by canvas mIoU and side MAE.
  \item \textbf{A planner and its data.} Trained on 15,000 reverse-cropped pairs whose rationales are screened by rules, an AI judge and human review of flagged rows, \anon{} reaches \res{ours_final@test300_seed0}{miou} canvas mIoU on 300 test crops, against \res{claude-opus_icl9@test300_seed0}{miou} for the best commercial VLM, and comparable to \anonnum{} (\res{direct15k@test300_seed0}{miou}); 74 participants prefer its frames to the VLM's in 59\% of trials under the same fill.
  \item \textbf{A rationale that can be checked and edited.} When its directional wording is edited, the grown sides follow the text on 82--91\% of crops, while rewording preserves them; 90.6\% of its verifiable claims about the photo are judged correct.
\end{enumerate}
\section{Related Work}
\label{sec:related}

\noindent \textbf{Cropping within the frame.}
Automatic cropping selects a better-composed window inside a photo, a subjective task with several valid answers. Saliency~\cite{suh2003thumbnail,sun2013soat} and composition rules or retargeting~\cite{datta2006aesthetics,liu2010optimizing,avidan2007seam,rubinstein2010retargetme} gave way to learning from professional edits~\cite{yan2013learning}, composition and boundary cues~\cite{fang2014automatic}, and professional photos and dense view pairs~\cite{chen2017vfn,wei2018cpc,chen2017fcdb}, with windows adjusted or scored~\cite{li2018a2rl,zeng2019gaic,zeng2020gaic} using attention and aesthetics~\cite{wang2019deepcrop,yang2026gafic}, candidate relations~\cite{li2020cgs}, boundary context~\cite{pan2021transview}, multiple valid crops~\cite{jia2022rethinking}, subjects~\cite{zhang2022humancrop,yang2023sacnet,wong2025aescrop} or composition rules~\cite{hong2021cacnet}. Recent work handles user-generated content~\cite{su2024s2cnet} and text queries~\cite{zhong2023clipcrop}, adopts vision-language pretraining~\cite{yuan2024vlpcrop}, and prompts VLMs with retrieved examples~\cite{lee2025cropper} or step-by-step analysis~\cite{dong2026crop}; concurrent work explains crops and 360$^\circ$ view selections~\cite{lee2026photoguider}. GenCrop~\cite{hong2024gencrop} and ProCrop~\cite{zhang2026procrop} outpaint professional photos to obtain cropping supervision; we reverse this direction and crop complete photos to obtain expansion supervision, which composition~\cite{zhang2021cadb,zhao2025picd,lee2018kupcp,mai2016composition} and aesthetic~\cite{murray2012ava,kong2016aadb,ren2017flickraes,yang2022para,he2022tad66k} datasets do not provide. Cropping shares our goal of better composition but chooses content \emph{inside} the view, whereas a subject cut by the border needs content \emph{outside} it. %

\noindent \textbf{Reframing beyond the border.}
When part of the subject lies outside the photo, no crop can restore it; the frame itself must change. Before capture, methods recommend views or camera motions~\cite{su2012viewrec,rawat2015context,su2021vapnet,sheng2025drlcomp,li2025spas}, steer a robot camera~\cite{alzayer2021autophoto}, search a 3D aesthetic field~\cite{tang2026aestheticfield} or visualize the adjustment as video~\cite{yao2025ppc}, all of which require re-shooting. After capture, outward cropping~\cite{zhong2021outward} predicts one enlargement level shared by all borders, outpaints the ring and searches for a crop; UNIC~\cite{liu2023unic} extrapolates features to recommend a camera view; and commercial tools expand photos automatically~\cite{google2024pixel9} or along user-set borders and aspect ratios~\cite{adobe2026genexpand}. Like outward cropping and these tools, \anon{} keeps the photo and leaves the fill to a generator; unlike them, it sets the amount for \emph{each side} in one step, explains why, and keeps the whole input inside the canvas. %

\noindent \textbf{Outpainting and generative fill.} %
Outpainting must extend a photo consistently, which grows harder with the added area. Image extrapolation progressed from CNN and GAN extenders~\cite{sabini2018outpainting,wang2019srn,yang2019verylong,teterwak2019boundless,kim2021poi,lin2022infinitygan,cheng2022inout,yao2022queryotr} to large-mask inpainting~\cite{suvorov2022lama} and diffusion models that fill through uncropping~\cite{saharia2022palette}, resampling~\cite{lugmayr2022repaint}, latent blending~\cite{avrahami2023bld}, conditional control~\cite{zhang2023controlnet,ju2024brushnet} and latent-diffusion backbones~\cite{rombach2022ldm,podell2024sdxl,bfl2025kontext}. Recent outpainters generate convincing extensions~\cite{zhuang2024powerpaint,yu2024shadow,yang2024vip,zhang2024pqdiff}, and studies of object expansion~\cite{eshratifar2024salient} and border seams~\cite{hou2025seamless} identify where fills still need care. Throughout this line the canvas is an input: composition-planned~\cite{song2025proout} and layout-controlled~\cite{kim2026blueprint} artwork outpainting, video outpainting~\cite{dehan2022video,fan2023m3ddm,wang2024motia,chen2024followyourcanvas,jeon2026seen2scene}, and aspect-ratio pipelines that warp~\cite{xu2025halo}, prune and repaint~\cite{shen2024prunerepaint} or outpaint fixed bands~\cite{paull2023stableremaster,wang2026pinterestcanvas} all start from a given target size. They decide \emph{what} to paint; \anon{} decides \emph{where} and \emph{how much}, and so complements them.

\noindent \textbf{VLMs for composition guidance.}
Composition depends on content as well as geometry, so instruction-tuned multimodal models~\cite{liu2023llava,bai2025qwen25vl,zhu2025internvl3,qwen2026qwen35} are a natural tool: they critique photographs~\cite{qi2025photographerseye,huang2024aesexpert}, learn aesthetics from language~\cite{ke2023vila,sheng2023aesclip,zhou2024uniaa}, describe and score quality~\cite{wu2024qinstruct,wu2024qalign,yang2026fgiaa,wu2026iqat1,you2025deqa,zhang2023liqe}, and give aesthetic guidance with rationales~\cite{du2026venus,li2026shuttermuse}. Benchmarks find their low-level and aesthetic perception uneven~\cite{wu2024qbench,huang2024aesbench}, which motivates task-specific fine-tuning, supervised or with verifiable rewards~\cite{liu2025visualrft}. MLLMs also emit executable edits, as retouching parameters with reasoning~\cite{dutt2025monetgpt,wu2026retouchiq} or iterative edit steps~\cite{zeng2025mira}. For framing, PhotoFramer~\cite{you2026photoframer} gives reframing instructions with generated exemplars, and the concurrent GeoComposer~\cite{li2026geocomposer} makes such exemplars geometrically consistent; an exemplar shows a target but leaves the expansion geometry \emph{implicit}. \anon{} makes it explicit: like detectors that regress four side distances~\cite{yu2016unitbox,tian2019fcos} and sequence models that emit box coordinates or layouts as tokens~\cite{chen2022pix2seq,chen2023shikra,peng2024kosmos2,feng2023layoutgpt}, it outputs executable per-side amounts as text.

\noindent \textbf{Checkable rationales and model judges.} %
A rationale helps only if its visual claims are correct and the decision follows it. Explanations generated before or with an answer~\cite{wei2022cot,camburu2018esnli,park2018multimodal,sammani2022nlxgpt} can read fluently without reflecting the prediction process~\cite{turpin2023unfaithful,jacovi2020faithful,ye2022unreliability,chen2025reasoning} or the image~\cite{uppaal2025journey}. Counterfactual interventions test whether explanations and predictions change together~\cite{lanham2023measuring,wiegreffe2021measuring,atanasova2023faithfulness,menzat2026counterfactual}, and program execution can enforce this~\cite{lyu2023faithfulcot}; as such tests measure self-consistency rather than faithfulness~\cite{parcalabescu2024measuring}, our interventions support only a consistency claim. Rationale distillation~\cite{hsieh2023distilling} often exposes the teacher to the answer~\cite{zelikman2022star,wang2023scott,dutt2025monetgpt}, inviting post-hoc justification. Like Prism~\cite{qiao2024prism}, which decouples perception from reasoning, we split the teacher in two, here so that the observation is written \emph{blind} to the answer and the explanation blind to the image, making every visual claim in a training rationale checkable. Model judges agree with humans in many settings~\cite{zheng2023judging,liu2023geval,gu2024judgesurvey,xiong2025llavacritic} yet show self-preference, order and egocentric biases~\cite{panickssery2024selfpref,wataoka2024selfpref,koo2024cobbler,chen2024mllmjudge}, so our judge only flags doubtful observations and a person decides. Hallucination benchmarks and language-prior analyses~\cite{li2023pope,rohrbach2018chair,guan2024hallusionbench,leng2024vcd,huang2024opera} frame the errors we report, which we score per atomic claim~\cite{min2023factscore,jing2024faithscore}.
\section{Method}
\label{sec:method}

\begin{figure*}[t]
\centering
\includegraphics[width=0.95\textwidth]{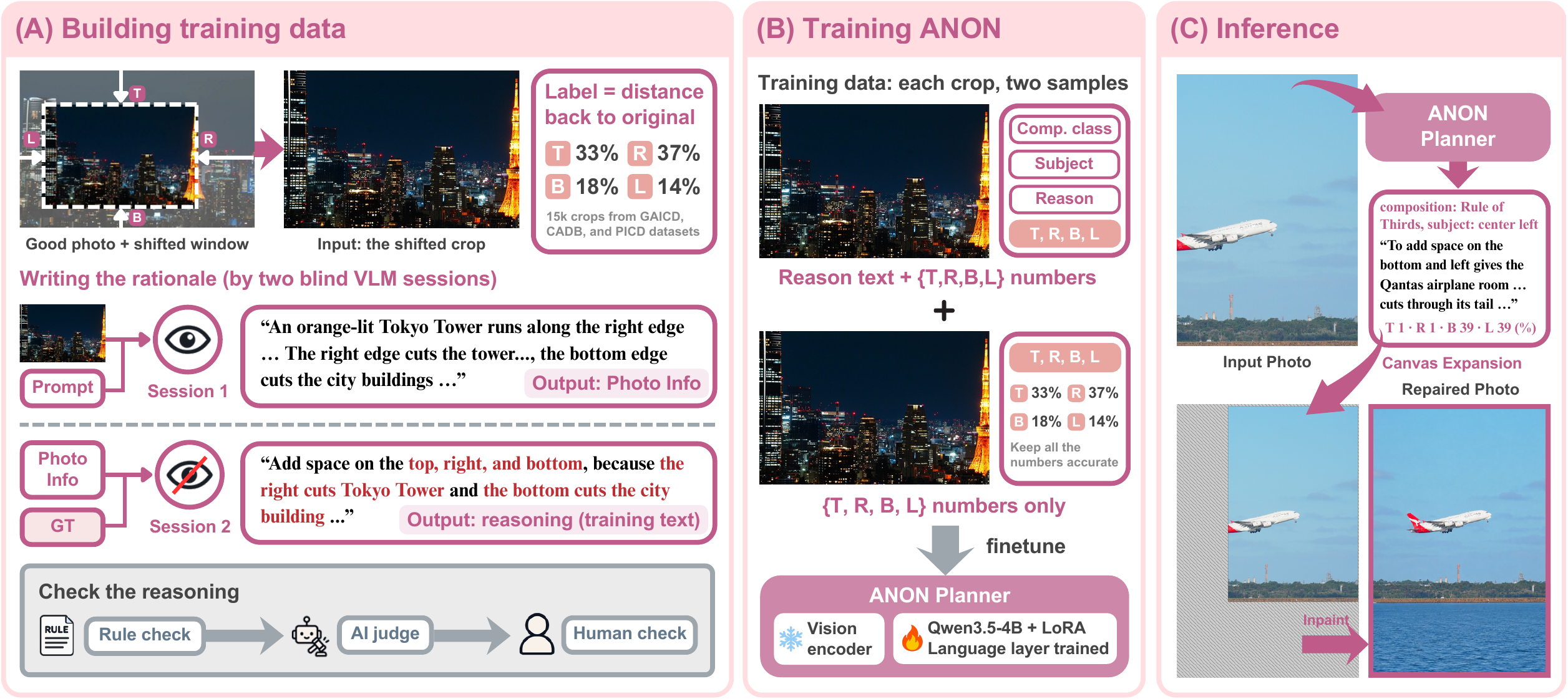} %
\caption{\textbf{Overview.} (a) Reverse cropping provides per-side distances to the reference frame (GT). Two isolated teacher sessions write the rationale: Session~1 sees the crop but not the answer and outputs an observation (Photo Info); Session~2 sees only facts (the observation, the subject cell and the GT sides to grow), never the image, and outputs the reasoning used as training text. Three checks follow: rules (names the sides to grow, no percentages, 20--150 words), an AI judge that re-reads the observation against the image (9.5\% of rows flagged), and a person who checks every flagged row, edits the observation, then reruns Session~2 and the checks. (b) Each crop provides a rationale sample and a numbers-only sample; the vision encoder is frozen and the language layers are trained with LoRA. (c) The planner fine-tuned in (b) reads the photo and the prompt and predicts composition class, subject cell, rationale and four percentages; these expand the canvas, which an interchangeable outpainter fills. Both examples are the authors' own photos with real outputs.}
\label{fig:pipeline}
\vspace{-8pt} %
\end{figure*}

\subsection{Task}
\label{sec:task}
The input is a badly framed photo $I$ (width $w$, height $h$), whose subject may be cut by the border. We predict four non-negative expansion ratios $e = (e_t, e_r, e_b, e_l)$: top and bottom relative to $h$, right and left relative to $w$, reported in percent. With the input normalized to $[0,1]^2$, the enlarged canvas is $B(e) = [-e_l,\, 1+e_r] \times [-e_t,\, 1+e_b]$. This expansion retains the full input frame. Accuracy is the IoU between predicted and reference canvases, $\mathrm{IoU}(B(\hat e), B(e^*))$ (\cref{sec:setup}).

\subsection{Reverse-synthesized training data}
\label{sec:data}
\begin{table}[t]
\caption{\textbf{Dataset statistics.} Target photos and reverse-cropped samples per source after the 15\%-side filter; the test set is fixed to 300 crops. Selection rule: how source photos are selected.} %
\label{tab:dataset}

\vspace{-8pt}

\papertablestyle
\fitpapertable{%
\begin{tabular}{lccc}
\toprule
Source & Targets (train/val) & Crops (train/val/test) & Selection rule \\
\midrule
GAICD & 2,289 / 168 & 5,000 / 473 / 96 & has a candidate box with MOS $\geq 4$ \\
CADB & 2,299 / 168 & 5,000 / 440 / 100 & 80\% mean score $\geq 3$, 20\% $< 3$ \\
PICD & 2,279 / 168 & 5,000 / 442 / 104 & 24 classes, balanced \\
Total & 6,867 / 504 & 15,000 / 1,355 / 300 & rationales: 100\% of train and val rows \\
\bottomrule
\end{tabular}}
\vspace{-6pt}
\end{table}

Pairs of a badly framed photo and its intended framing are hard to collect, so we build them backwards: from a complete photo we cut a displaced crop as the input, and the per-side distances from the crop back to the full photo, which we call the \emph{reference frame (GT)}, are the answer. The reference frame is the photographer's own frame, from well-composed photos to everyday snapshots. For a crop box $(x_1, y_1, x_2, y_2)$ in a $W \times H$ reference frame, with $w = x_2 - x_1$ and $h = y_2 - y_1$, %
\begin{equation}
e^* = \Bigl(\frac{y_1}{h},\ \frac{W - x_2}{w},\ \frac{H - y_2}{h},\ \frac{x_1}{w}\Bigr),
\label{eq:label}
\end{equation}
rounded to integer percentages in the target text. %
We sample from GAICD~\cite{zeng2020gaic}, CADB and PICD using the selection rules in \cref{tab:dataset}. BiRefNet~\cite{zheng2024birefnet} foreground masks, filtered by DeepGaze~IIE~\cite{linardos2021deepgaze} fixation density, locate the subject (supplement); %
displaced crops cover 50--80\% of the photo area and span three subject-retention tiers, from pushing the subject to a border to removing a piece (construction details in the supplement). We use 15,000 training crops (5,000 per source) and 1,355 validation crops, each with at least one side requiring 15\% or more for rationale supervision. The test set is unfiltered: 300 crops drawn at random from the 3,807-crop test pool, including 40 with all sides below 15\%. Source photos are disjoint across splits. %

\subsection{Two-stage rationale supervision} %
\label{sec:rationale}
A checkable rationale needs its statements about the image to be separable from the answer it justifies, so each rationale is written in two steps, each in a fresh, isolated session of the teacher (Claude Sonnet 5.5). \textbf{Observe}: the teacher sees the crop only, without the answer, and writes the visible facts: where the subject sits, where a face or gaze points, which borders cut the subject or an important object, which sides are empty, whether there are strong lines. ``Cut'' requires part of the object to lie outside the frame; a complete object close to the border is ``near''. \textbf{Explain}: a new session receives no image, only the observation, the subject's $3\times3$ cell from the detector and the sides to grow (those at 15\% or more). It is asked for a rationale of at most 100 words (the validator's hard limit is 150) that names the sides with the fixed phrase ``add space on the \emph{X}''. With $c$ the subject cell, the two steps are
\begin{equation}
o = T_{\mathrm{obs}}(I), \qquad z = T_{\mathrm{exp}}\bigl(o,\, c,\, S\bigr),
\label{eq:teacher}
\end{equation}
with $S = \{\, s : e^*_s \ge 0.15 \,\}$.
Because $T_{\mathrm{exp}}$ receives no image, visual claims beyond the detected cell $c$ can be checked against the observation $o$. %

Each rationale goes through three checks. (1) \emph{Rules}: name the largest side, claim no side at or below 5\%, deny no grown side, omit percentages and avoid crop suggestions. (2) \emph{AI judge}: an isolated session compares the observation with the image and flags unsupported or incorrect claims. (3) \emph{Human verdict}: every flagged observation is reviewed by a person and corrected where wrong, then used to regenerate and revalidate the rationale.
The judge flags 1,551 of 16,355 training and validation rows (9.5\%), so human review covers only these rows. Of the final 15,000 rationales, 99.8\% name exactly the grown sides. Review statistics and repeated judge errors that motivate human adjudication are in the supplement.

\subsection{Model, training and outpainting}
\label{sec:method_inference}
\label{sec:model}
\label{sec:outpaint}
The planner is \vlm fine-tuned with LoRA while keeping the vision encoder frozen (\cref{sec:setup}). Its output is a fixed-format text with four fields: the target composition class (the dataset's label, ``unknown'' when absent), the subject's $3\times3$ cell, the rationale, and the four percentages. To train both explanation and numerical prediction, we turn the same 15,000 crops into two kinds of samples (\cref{sec:ablation}): a \emph{rationale sample}, whose prompt $x^{\mathrm{rat}}$ asks for all four fields and whose target $y^{\mathrm{rat}} = (\text{class}, c, z, e^*)$ is one JSON object, and a \emph{numbers-only sample}, whose prompt $x^{\mathrm{num}}$ asks for $e^*$ alone. Both are trained together with next-token cross-entropy on the target tokens only,
\begin{equation}
\mathcal{L}(\theta) = -\!\!\sum_{(x,y) \in \mathcal{D}} \sum_{k=1}^{|y|} \log p_\theta(y_k \mid y_{<k}, x),
\label{eq:loss}
\end{equation}
where $\mathcal{D} = \mathcal{D}_{\mathrm{rat}} \cup \mathcal{D}_{\mathrm{num}}$ with $|\mathcal{D}_{\mathrm{rat}}| = |\mathcal{D}_{\mathrm{num}}| = 15{,}000$, shuffled into one set, and the loss is averaged over target tokens in each batch. At inference, the planner receives $x^{\mathrm{rat}}$ %
 (schedule and model selection in \cref{sec:setup}).

At inference, the four ratios define the enlarged canvas and its fill mask for an interchangeable outpainter (\cref{sec:generation}). We report raw outpainter outputs, including the small changes within the input region (kept-region LPIPS 0.053 for GPT Image 2.5; supplement). %

\section{Experiments}
\label{sec:experiments}

\subsection{Setup}
\label{sec:setup}
\paragraph{Data and metrics.} Training uses the 15,000 crops of \cref{sec:data}; the epoch is chosen on a fixed sample of 300 validation crops (splits in the supplementary material), and every method is evaluated once on the same fixed 300 test crops. The main metric is \emph{canvas mIoU} over the $N$ test crops, and \emph{side MAE} is the mean absolute difference in percentage points over the four sides, with inward predictions clipped to 0, as an inward crop adds no canvas (GAIC $=$ no-op): %
\begin{align}
\mathrm{mIoU} &= \frac{1}{N} \sum_{i=1}^{N} \mathrm{IoU}\bigl(B(\hat e_i), B(e^*_i)\bigr), \label{eq:miou}\\
\mathrm{MAE} &= \frac{100}{4N} \sum_{i=1}^{N} \sum_{s} \bigl|\max(\hat e_{is}, 0) - e^*_{is}\bigr|, \label{eq:mae}
\end{align}
with $B$ defined in \cref{sec:task}. %
Canvas mIoU measures geometric recovery; perceived composition is evaluated separately (\cref{sec:generation}). We also report added-region IoU in the supplement. \emph{Direction F1} pools all $4N$ sides, using $y_{is}=\mathbb{1}[e^{*}_{is}>\tau]$ and $\hat y_{is}=\mathbb{1}[\max(\hat e_{is},0)>\tau]$, with $\tau=10\%$: $F_1 = 2\,\mathrm{TP}/(2\,\mathrm{TP}+\mathrm{FP}+\mathrm{FN})$. This threshold approximately balances positive and negative validation sides; threshold sensitivity is reported in the supplement. Tables report one run per model with 95\% bootstrap CIs over test crops; $p$-values come from a two-sided exact sign test on per-crop IoU differences (supplement). Training variability is assessed separately using three seeds for \anon{} and \anonnum{} (\cref{sec:results}).

\noindent \textbf{Implementation.} LoRA rank 8, $\alpha=32$, all linear layers of the language model; learning rate $10^{-4}$, effective batch 16, five epochs; greedy decoding; one RTX 5080 (16 GB). \anonnum{} trains for ten epochs so that the training budgets match (about 9,380 vs.\ 9,375 updates); each model's checkpoint is chosen independently on validation (further details in the supplement). %
At batch size 1 on the same GPU, \anon{} takes 4.8\,s per crop (137 output tokens on average), against 0.9\,s for \anonnum{}; both use about 9\,GB of peak memory.

\noindent \textbf{Baselines.} \emph{Trivial}: no-op and a constant expansion equal to the training mean (T/R/B/L 17/18/16/18\%). \emph{Inward}: GAIC~\cite{zeng2020gaic}. \emph{Outward}: UNIC~\cite{liu2023unic}, raw and with only its outward part. \emph{Open VLMs}: \vlm without fine-tuning, InternVL3-2B without and with nine in-context examples (w/ ICL). \emph{Commercial VLMs}: GPT-4o, GPT-5.5, GPT-6 Astra, Claude Haiku / Sonnet / Opus 5.5, without and with the same nine examples (w/ ICL), one per source and tier from the training set. \emph{\anonnum{}}: our numbers-only variant, the same model and crops trained on numbers-only samples alone. Pure regressors (a CNN from scratch, frozen CLIP and DINOv2 embeddings) are in the supplementary material. %
\begin{table}[t]
\caption{\textbf{Main comparison on the fixed 300-crop test subset.} Canvas mIoU with 95\% bootstrap CI, side MAE in percentage points and direction F1 at the 10\% threshold; VLMs get the same prompt; (w/ ICL) = the same nine in-context examples (In-Context Learning), otherwise no examples. Subj.\ pos.\ / size: error of the subject's position and area share relative to the reference frame (the 265 crops with a detected subject; inward sides clipped to 0, as for MAE, so GAIC equals no-op). Best, second and third per column: \rankkey{rankfirst}{red}, \rankkey{ranksecond}{orange}, \rankkey{rankthird}{yellow}.}
\label{tab:main}
\papertablestyle
\fitpapertable{%
\begin{tabular}{lccccc}
\toprule
Method & \miou$\uparrow$ {\scriptsize(95\% CI)} & Side MAE$\downarrow$ & Dir.\ F1$\uparrow$ & Subj.\ pos.$\downarrow$ & Subj.\ size$\downarrow$ \\
\midrule
\multicolumn{6}{l}{\textit{References: trivial, inward cropping, outward camera-view model}} \\
No-op (canvas = input) & 0.593 {\scriptsize[0.582, 0.603]} & 15.71 & 0.000 & 0.181 & 0.207 \\
Constant (train mean) & 0.691 {\scriptsize[0.682, 0.701]} & 13.11 & 0.672 & 0.135 & \rankB 0.045 \\
GAIC & 0.401 {\scriptsize[0.390, 0.412]} & 15.71 & 0.000 & 0.181 & 0.207 \\
UNIC, raw & 0.596 {\scriptsize[0.584, 0.607]} & 13.72 & 0.213 & 0.152 & 0.150 \\
UNIC, expand-only & 0.643 {\scriptsize[0.631, 0.654]} & 13.72 & 0.213 & 0.152 & 0.150 \\
\midrule
\multicolumn{6}{l}{\textit{Open VLMs}} \\
Qwen3.5-4B & 0.549 {\scriptsize[0.527, 0.571]} & 29.37 & 0.504 & 0.162 & 0.106 \\
InternVL3-2B & 0.613 {\scriptsize[0.602, 0.624]} & 15.22 & 0.293 & 0.171 & 0.176 \\
InternVL3-2B (w/ ICL) & 0.610 {\scriptsize[0.599, 0.621]} & 15.07 & 0.003 & 0.179 & 0.183 \\
\midrule
\multicolumn{6}{l}{\textit{Commercial VLMs (w/ and w/o ICL)}} \\
GPT-4o & 0.672 {\scriptsize[0.660, 0.684]} & 14.54 & 0.568 & 0.139 & 0.066 \\
GPT-4o (w/ ICL) & 0.676 {\scriptsize[0.663, 0.689]} & 13.72 & 0.474 & 0.157 & 0.054 \\
GPT-5.5 & 0.700 {\scriptsize[0.688, 0.711]} & 12.51 & 0.653 & 0.124 & 0.062 \\
GPT-5.5 (w/ ICL) & 0.713 {\scriptsize[0.701, 0.725]} & 11.91 & 0.644 & 0.121 & 0.055 \\
GPT-6 Astra & 0.707 {\scriptsize[0.694, 0.719]} & 12.10 & 0.663 & 0.117 & 0.066 \\
GPT-6 Astra (w/ ICL) & 0.734 {\scriptsize[0.723, 0.746]} & 10.75 & \rankC 0.701 & \rankC 0.103 & 0.057 \\
Claude Haiku 5.5 & 0.690 {\scriptsize[0.678, 0.702]} & 12.14 & 0.435 & 0.135 & 0.082 \\
Claude Haiku 5.5 (w/ ICL) & 0.688 {\scriptsize[0.676, 0.701]} & 12.48 & 0.435 & 0.149 & 0.072 \\
Claude Sonnet 5.5 & 0.727 {\scriptsize[0.715, 0.738]} & 10.83 & 0.605 & 0.117 & 0.051 \\
Claude Sonnet 5.5 (w/ ICL) & 0.730 {\scriptsize[0.718, 0.742]} & 10.90 & 0.645 & 0.121 & \rankC 0.046 \\
Claude Opus 5.5 & 0.724 {\scriptsize[0.712, 0.736]} & 10.77 & 0.585 & 0.114 & 0.064 \\
Claude Opus 5.5 (w/ ICL) & \rankC 0.748 {\scriptsize[0.737, 0.760]} & \rankC 9.96 & 0.686 & 0.108 & 0.047 \\
\midrule
\anonnum{} (ours) & \rankB 0.789 {\scriptsize[0.775, 0.804]} & \rankB 8.58 & \rankB 0.748 & \rankB 0.098 & \rankA 0.022 \\
\anon{} (ours) & \rankA 0.797 {\scriptsize[0.781, 0.814]} & \rankA 8.33 & \rankA 0.766 & \rankA 0.095 & \rankA 0.022 \\
\bottomrule
\end{tabular}}
\vspace{-6pt}
\end{table}

\begin{figure*}[t]
\centering
\includegraphics[width=1.0\textwidth]{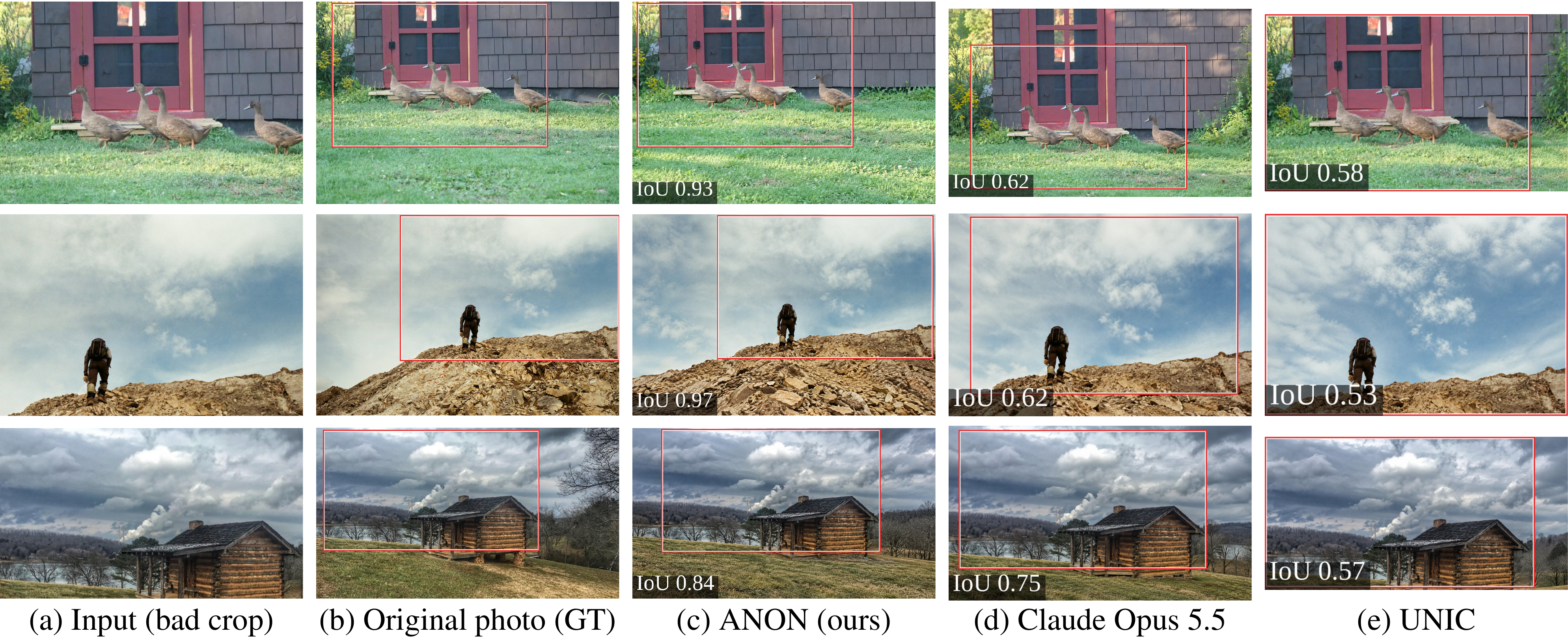} %
\caption{\textbf{Qualitative comparison on crops from the test set.} (a) input, (b) the reference frame (GT) with the input region framed, (c) \anon{}, (d) Claude Opus 5.5 (w/ ICL), (e) UNIC; (c)--(e) share the same outpainter (GPT Image 2.5). The red frame marks the input region and each panel shows its IoU with the reference frame. Examples are chosen where the planners differ most; full-set numbers are in \cref{tab:main}.}
\label{fig:qual}
\vspace{-8pt} %
\end{figure*}

\subsection{Main results}
\label{sec:results}
\Cref{tab:main} compares all methods on the same 300 crops. \anon{} reaches \res{ours_final@test300_seed0}{miou} canvas mIoU, exceeding the strongest commercial VLM, Claude Opus 5.5 (w/ ICL), by $0.049$ ($p<0.001$). The gap also holds when scoring only the added region: IoU is 0.577 for \anon{} against 0.463 for Opus and 0.409 for constant expansion (supplement). \anon{} also places the subject closest to its position and size in the reference frame (\cref{tab:main}). %
Inward cropping cannot add canvas, so GAIC reaches \res{gaic@test300_seed0}{miou}, below no-op (\res{noop@test300_seed0}{miou}); UNIC's outward-only prediction reaches \res{unic_clip0@test300_seed0}{miou} and constant expansion \res{constant@test300_seed0}{miou}; snapping the constant baseline's canvas to the nearest common aspect ratio (1:1, 4:3, 3:2, 16:9 and their portrait versions) changes nothing (0.691), as 91\% of the crops keep the photo's aspect. %
The best commercial VLM reaches \res{claude-opus_icl9@test300_seed0}{miou}; the supplement analyzes how its in-context examples contribute magnitude and direction information.
\anonnum{} reaches \res{direct15k@test300_seed0}{miou}; its paired difference from \anon{} is not significant (95\% CI $-0.006$ to $+0.022$, $p=0.59$). Across three seeds, \anon{} and \anonnum{} reach $0.789 \pm 0.007$ and $0.801 \pm 0.011$ (mean $\pm$ std), with no stable ordering. Thus \anon{} offers comparable geometric accuracy together with a rationale that can be checked and edited (\cref{sec:faithful}).

\noindent \textbf{Manually cropped real images.} %
On 81 manually cropped images from 48 source images (59 camera crops, 7 game screenshots and 15 animation crops), \anon{} reaches 0.684 canvas mIoU against 0.641 for Opus; the paired difference has a source-photo bootstrap 95\% CI of $[+0.014,+0.069]$ (supplement). Both under-expand on this set, which requires larger expansions than the training crops (25.6\% per side vs.\ 17.1\%). This set tests framing only; capture defects in casual photos, such as those in SPAQ~\cite{fang2020spaq}, local motion blur~\cite{chen2026bocchirealisticchallengingbenchmark} and low-light color casts~\cite{10.1007/978-3-032-37032-7_5}, are outside the task and need other pipeline modules.

\subsection{Ablations}
\label{sec:ablation}
\begin{table*}[t]
\caption{\textbf{Per-side decisions.} \anon{}: its predictions rewritten after the fact (\cref{sec:ablation}). Claude Opus 5.5 (w/ ICL): the same rewrite applied to its predictions (stating each rule in its prompt instead: supplement). \emph{Dir.} = per-axis direction kept, \emph{Amount} = amount kept, \emph{Grid} = coarsening. \anonnum{} results are in the supplement. Bold = best, underline = second, over both models.}
\label{tab:abl_geometry}
\papertablestyle
\fitpapertable{%
\begin{tabular}{l|ccc |ccc ccc}
\toprule
 & & & & \multicolumn{3}{c}{\anon{} (rewritten)} & \multicolumn{3}{c}{Opus w/ ICL (rewritten)} \\
\cmidrule(lr){5-7}\cmidrule(lr){8-10}
Rewrite & Dir. & Amount & Grid & \miou$\uparrow$ & MAE$\downarrow$ & F1$\uparrow$ & \miou$\uparrow$ & MAE$\downarrow$ & F1$\uparrow$ \\
\midrule
as predicted & \checkmark & \checkmark & -- & \best{0.797} & \best{8.33} & \best{0.766} & 0.748 & 9.96 & 0.686 \\
symmetrized (area-matched) & -- & -- & -- & 0.713 & 12.27 & 0.666 & 0.704 & 12.26 & 0.606 \\
balanced axes & -- & \checkmark & -- & 0.718 & 11.97 & 0.678 & 0.703 & 12.28 & 0.584 \\
fixed magnitude (train mean) & \checkmark & -- & -- & 0.747 & 10.60 & 0.751 & 0.748 & 10.33 & 0.725 \\
quantized, 25\% grid & \checkmark & \checkmark & 25\% & 0.731 & 11.36 & \second{0.763} & 0.706 & 11.58 & 0.639 \\
quantized, 10\% grid & \checkmark & \checkmark & 10\% & \second{0.778} & \second{9.12} & 0.756 & 0.736 & 10.46 & 0.613 \\
binary, 0 or 30\% & \checkmark & -- & 0/30 & 0.743 & 10.52 & 0.756 & 0.699 & 11.94 & 0.613 \\
\bottomrule
\end{tabular}}
\vspace{-6pt}
\end{table*}

\noindent \textbf{Must each side be decided on its own? (\cref{tab:abl_geometry})} We modify the same predictions to isolate their geometric information. The symmetric control assigns one ratio $s$ to all sides and preserving canvas area, $(1+2s)^2=(1+\hat e_l+\hat e_r)(1+\hat e_t+\hat e_b)$. Balanced axes preserve each axis total but split it equally. Fixed magnitude preserves the predicted split but sets axis totals to their training means. Balancing the axes reduces mIoU from \res{ours_final@test300_seed0}{miou} to \res{ours_final_balanced@test300_seed0}{miou}; fixing magnitudes reduces it to \res{ours_final_fixmag@test300_seed0}{miou}. Both directional allocation and image-dependent magnitude therefore contribute to accuracy. Applied to Claude Opus 5.5 (w/ ICL), the same rewrites cost less (symmetrized $-$0.044 vs.\ $-$0.084), and fixing its magnitude costs nothing (0.748, level with ours under the same rewrite, 0.747): its amounts carry no more than the training mean, so \anon{}'s margin lies in the amount and its placement. %
Symmetric, quantized and binary controls also reduce mIoU significantly; \anonnum{} shows the same pattern. Separately retrained one-ratio and per-axis models reach \res{sym1d15k@test300_seed0}{miou} and \res{axis2d15k@test300_seed0}{miou}, against \res{direct15k@test300_seed0}{miou} for four per-side ratios (supplement).

\noindent \textbf{Rationale vs.\ number samples (supplement).} With the same final data and 9,375 steps, rationale samples alone give \res{abl_final_r1to0@test300_seed0}{miou}; adding one or two numbers-only samples per crop gives \res{ours_final@test300_seed0}{miou} and \res{abl_final_r1to2@test300_seed0}{miou}. The 1:1 mixture has the highest observed mIoU among these ratios. Keeping only 4,780 rationales, with numbers-only samples for all 15,000 crops, gives \res{abl_final_rs4780@test300_seed0}{miou}. All CIs overlap, so these are trends rather than established differences.

\begin{figure*}[t]
\centering
\includegraphics[width=1.0\textwidth]{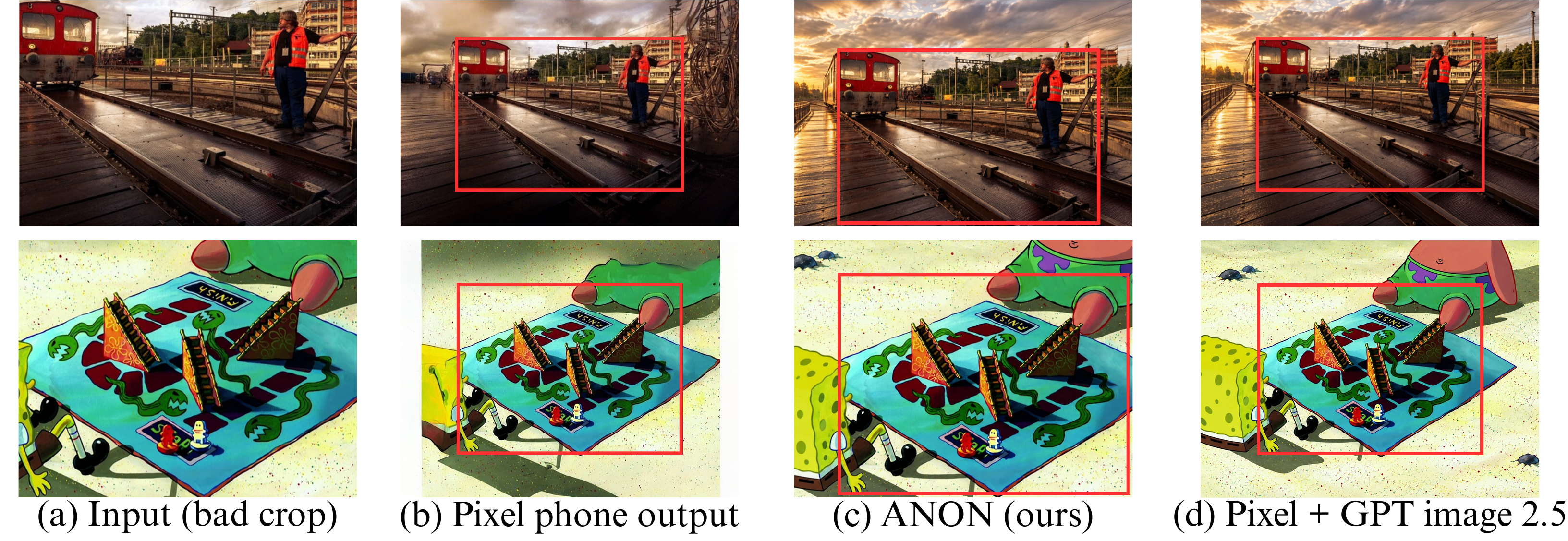} %
\caption{\textbf{A shipped product next to the planner.} (a) input, (b) Pixel-AF output (its own decision and fill), (c) \anon{} and (d) the Pixel-AF decision, both cropped from one shared GPT Image 2.5 fill, so they differ only in the frame; red = input region. Top: a hand-cropped photograph; bottom: a cartoon frame without a reference frame (both from the user study).} %
\label{fig:pixel}
\vspace{-8pt} %
\end{figure*}

\subsection{Rationale consistency and intervention}
\label{sec:faithful}
\begin{table}[t]
\caption{\textbf{Rationale consistency and intervention.} Final model, 300 test crops. Consistency: the sides named in the text are the sides grown. Intervention: the side words in the rationale are swapped and the numbers regenerated; the share whose numbers follow the text (all side words / decision clause only). Rewording: the rationale is paraphrased by GPT-6 Astra with every side word and the decision clause kept (299 crops); the share whose grown sides are unchanged. Claim correctness is judge-based: the share of atomic factual claims a GPT-6 Astra judge marks correct, 2,222 verifiable claims of 2,246, CI by cluster bootstrap over photos; two annotators also labeled a 300-claim sample (mean of the two; they agree with the judge on 85.7\% of claims). %
 Plausibility: 222 ratings from 74 user-study participants, CI by bootstrap over participants.}
\label{tab:faithful}
\papertablestyle
\fitpapertable{%
\begin{tabular}{p{5.6cm}c}
\toprule
Metric & \anon{} (ours) \\
\midrule
Claimed sides = grown sides (consistency) & 98.7\% [97.3, 99.7] \\
Intervention: numbers follow a swapped rationale (all side words / decision clause) & 91.3\% / 82.0\% \\
Intervention: same rationale reworded (numbers keep their sides) & 99.7\% [99.0, 100] \\
Claim correctness, judge-based (GPT-6 Astra, 2,222 verifiable claims) & 90.6\% [89.4, 91.8] \\
Claim correctness, human (two annotators, 300-claim sample) & 87.5\% \\
Rationale plausibility, mean 1--5 (user study) & 4.08 [3.86, 4.27] \\
\bottomrule
\end{tabular}}
\vspace{-6pt}
\end{table}

We evaluate the final model's rationales on 300 test crops (\cref{tab:faithful}). 
\textbf{Consistency.} Named sides match those grown by at least 15\% on 98.7\% of crops, and every rationale names the largest side. 
\textbf{Intervention.} With image, prompt and decoding fixed, we replace the rationale and regenerate only the numbers. After all side words are swapped, the grown sides follow the swapped text on 91.3\% of crops; after only the ``add space on \ldots'' clause is swapped, on 82.0\% (they change at all on 99.0\% and 97.7\%). Mirroring the rationale lowers mIoU from \res{ours_final@test300_seed0}{miou} to 0.605, while meaning-preserving rewording leaves grown sides unchanged on 99.7\% of crops. The expansions therefore follow the rationale's directional content; text-order interventions in the supplement examine this further.

\textbf{Claim correctness.} GPT-6 Astra judges atomic factual claims against the crop as correct, incorrect or unverifiable, excluding decision clauses evaluated above. Of 2,246 claims, 24 are unverifiable; 90.6\% of the remainder are judged correct, and 47\% of crops have no incorrect claim. Orientation is weakest (78.5\%), while subject and scene claims are strongest (96\%). Per-type results and auxiliary-field accuracy are in the supplement.

\textbf{Plausibility.} In the user study (\cref{sec:generation}) participants rated three \anon{} rationales, each shown with input and output, at 4.08 on a 1--5 scale (222 ratings), and chose the expanded photo over the input in 95\% of cases (210 of 222). We use these checks because text-similarity scores are insensitive to direction-word errors; the supplement tests this limitation.
\textbf{Attention.} While the planner writes the subject cell, three attention heads in decoder layer 15 look at the subject (AUC 0.832; 0.617 without LoRA; supplement).

\subsection{Outpainted results}
\label{sec:generation}

\noindent \textbf{Outpainter.} %
With the planner fixed, we compare three outpainters on the same 102 canvases using input-region change, KID, seam ratio and a blind check of 60 outputs (supplement). In the blind check, visible failures occur in 0\% of GPT Image 2.5, about 7\% of FLUX.1 Fill [dev] and about 83\% of SDXL inpainting outputs; we use GPT Image 2.5 for subsequent results. Because the planner outputs only a canvas and a mask, any of these outpainters can serve as the backend, or as a fallback when one declines a request; FLUX.1 [dev], however, is not licensed for commercial use.
\begin{table}[t]
\caption{\textbf{User study} (74 participants, 12 photos). B-T score: Bradley--Terry worth $\pi$ fitted on these four conditions, $P(i \succ j)=\pi_i/(\pi_i+\pi_j)$, normalized so that the four sum to 1, with a 95\% CI. \anon{} wins: share of trials in which \anon{} is chosen over the row, with a 95\% CI in parentheses. Both CIs come from a bootstrap over participants and photos. Photos: photos on which \anon{} is chosen in the majority of trials (the cartoon frame has no reference frame). $p$: sign test over participants, Holm-corrected over the four comparisons. Pixel-AF brings its own fill, so it has no B-T score and is compared with \anon{} as a whole system.} %
\label{tab:userstudy}
\papertablestyle
\fitpapertable{%
\begin{tabular}{lcccc}
\toprule
Condition & B-T score$\uparrow$ & \anon{} wins (CI) & Photos & $p$ \\
\midrule
Ref. (GT) & 0.257 {\scriptsize[0.18, 0.36]} & 53\% (41--65) & 8/11 & 0.08 \\
Constant & 0.160 {\scriptsize[0.11, 0.22]} & 67\% (58--76) & 10/12 & $<$1e-10 \\
Opus (w/ ICL) & 0.257 {\scriptsize[0.19, 0.32]} & 59\% (51--66) & 8/12 & $<$1e-6 \\
Pixel-AF (system) & -- & 79\% (72--86) & 12/12 & $<$1e-17 \\
\midrule \textbf{\anon{} (ours)} & \textbf{0.326} {\scriptsize[0.27, 0.39]} & -- & -- & -- \\
\bottomrule
\end{tabular}}
\vspace{-6pt}
\end{table}

\noindent \textbf{User study.} 74 participants compared 64 pairs each across 12 photos (6 test crops, 2 hand-cropped photographs, 3 game screenshots and 1 cartoon frame without a reference frame), with randomized order and sides and attention checks. We randomly selected test crops from the test set and added game screenshots and cartoon frames from the web and a few photos we took. %
\begin{figure*}[t]
\centering
\includegraphics[width=0.9\textwidth]{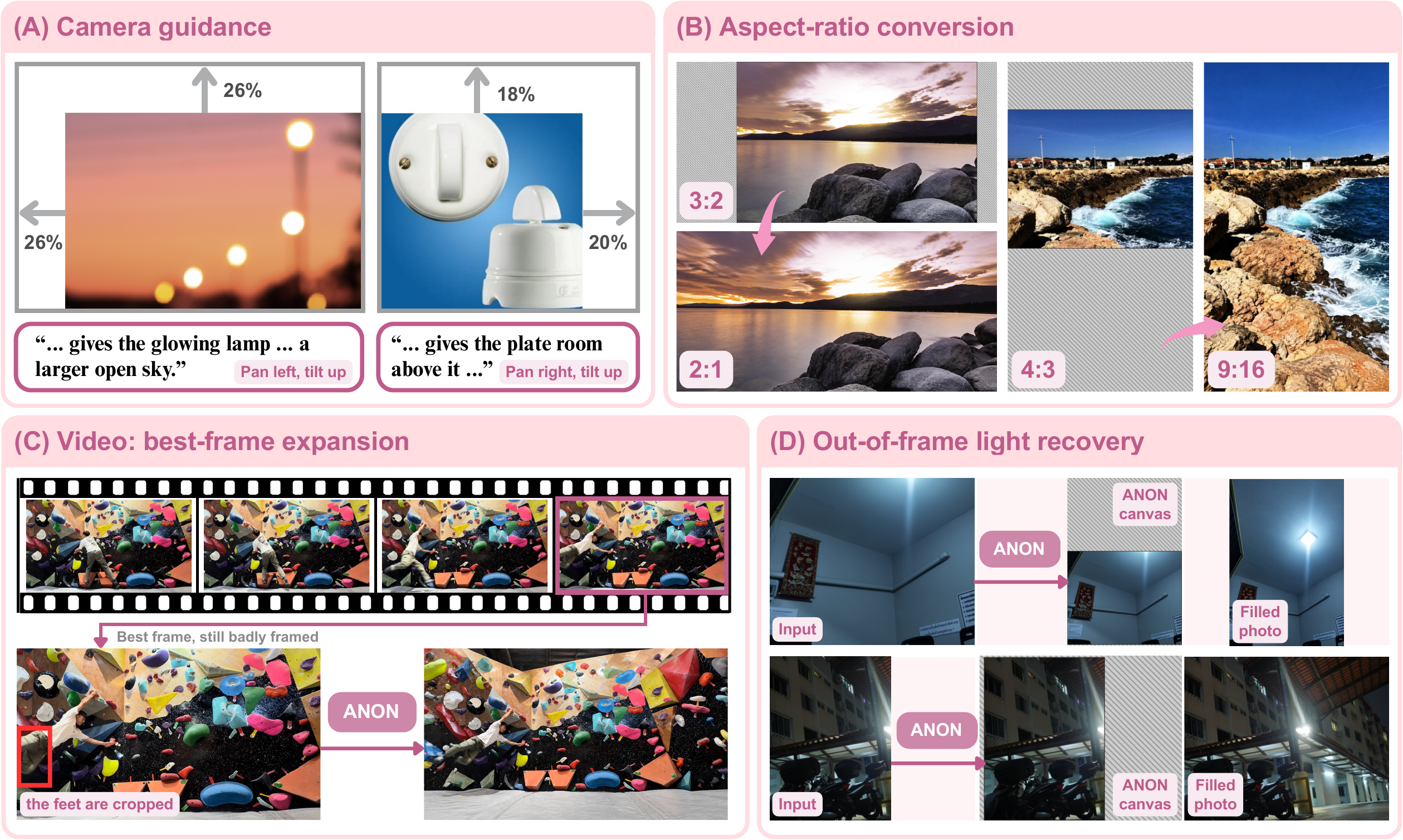} %
\caption{\textbf{Downstream uses of the planner.} (a) Camera guidance: the four percentages become arrows on the preview and a pan / tilt instruction, with the rationale as the explanation. (b) Aspect-ratio conversion (3:2 to 2:1; 4:3 to 9:16): the planner replaces the uniform 17\% expansion with 25\% left and 8\% right, prioritizing the lake over the rocks, and splits the added height between sky and sea (each pair: canvas, then result). (c) Video: the best frame of a clip still cuts the climber's feet; the planner grows the frame (authors' own clip). (d) Out-of-frame light recovery as in LightsOut~\cite{tsai2025lightsout}: a VLM names the side of the hidden light, the planner sets how far to grow, the outpainter restores the source (inputs from its project page).} %
\label{fig:apps}
\vspace{-8pt} %
\end{figure*}
\emph{The expansion decision.} \anon{}, Opus (w/ ICL) and constant expansion use crops from one shared GPT Image 2.5 fill, isolating framing; the reference frame is shown as photographed. \anon{} is preferred to Opus in 59\% of trials and to constant expansion in 67\% (\cref{tab:userstudy}); constant expansion adds more canvas than \anon{} on these photos (1.81 vs.\ 1.68 times the input area on average), so added area alone does not explain the preference over it. %
 \anon{} is preferred over the reference frame in 53\% of trials overall (not significant), 60\% on test crops and 45\% on non-test images ($p=0.30$ for the latter). \emph{The whole system.} Against Pixel-AF's own framing and fill, \anon{} is preferred in 79\% of trials and by 72 of 74 participants. This comparison includes both frame and fill; \cref{fig:pixel}d separately shows the Pixel-AF frame with our fill.

\subsection{Applications}
\label{sec:applications}
Without retraining, the same four percentages and rationale feed camera guidance, aspect-ratio conversion, repair of a selected video frame and light-source recovery, each with a task-specific downstream step (\cref{fig:apps}; further uses in the supplement). For video, existing frame selection, ranking, highlight detection or summarization~\cite{chang2016triage,ren2020bestframe,lei2021qvhighlights,liu2024r2tuning,wang2025timer1,apostolidis2021pglsum} chooses the moment; \anon{} repairs its frame.
\emph{Camera-guidance rule (\cref{fig:apps}a).} The four percentages of the planner become one instruction: all sides $<$8\%, the framing is fine; all sides $\geq$15\%, step back; left and right differ by $\geq$10\%, pan toward the larger; top and bottom differ, tilt; otherwise shift toward the largest side. The rationale is shown unchanged as the explanation. Auto-Filmer~\cite{zhang2023autofilmer} executes a human-given framing target on a drone; the planner could supply that target.
\emph{Storyboard expansion.} Five keyframes of the authors' climbing clip (0.5, 2.0, 3.5, 4.2 and 4.5\,s; the last three are the take-off and the airborne climber leaving the frame at the upper left) are each expanded by the planner (top and left 30\% for the first two, 38--39\% for the last three) and outpainted with the previous filled keyframe as the reference image, which keeps the wall and the lighting consistent across the board.

\subsection{Limitations}
\label{sec:limits}
Composition has no unique optimum and its judgment varies across viewers. Canvas mIoU measures recovery of the photographer's own frame; aesthetic optimality lies outside its scope, and perceived composition is assessed in the user study. As with every generative outpainter and commercial generative-expand tool, the fill is a plausible continuation, not evidence of the actual scene; its quality is bounded by the outpainter and drops as the added area grows.

\FloatBarrier %
\section{Conclusion}
\label{sec:conclusion}
We formulated per-side expansion as an explicit decision and built \anon{}, which predicts four ratios and a rationale. It achieves geometric accuracy comparable to its numbers-only variant and exceeds all tested VLMs; its rationale makes claims about the image that can be checked, and editing its decision clause changes the predicted expansions. The same output works with any of the tested outpainters and in four applications. Data, prompts and code will be released.

\section*{LLM Usage Statement}
Large language models and generative image models are part of the method and of its evaluation, and were also used in writing; all uses are listed here.
\emph{Data (Sec.~\ref{sec:rationale}).} Claude Sonnet 5.5 wrote the training rationales in two isolated sessions (observe, explain) and, in a third isolated session, acted as the judge that flags doubtful observations; the flagged rows were decided and corrected by a person.
\emph{Model.} The planner is a LoRA fine-tune of Qwen3.5-4B, a vision-language model.
\emph{Evaluation (Sec.~\ref{sec:faithful}, \ref{sec:generation}).} GPT-6 Astra judged the factual claims of the model's test rationales and paraphrased them for the rewording test; in light recovery (Fig.~\ref{fig:apps}d), an off-the-shelf VLM names the side of the hidden light; GPT-4o, GPT-5.5, GPT-6 Astra, Claude Haiku / Sonnet / Opus 5.5, InternVL3-2B and Qwen3.5-4B are compared as baselines without and with in-context examples (w/ ICL); GPT Image 2.5, FLUX.1 Fill and SDXL Inpainting are the outpainters. %
\emph{Writing and code.} Claude (Claude Code) and ChatGPT were used to edit the English, to draft and polish paragraphs from the authors' notes, to check the manuscript against the experiment logs, and to write and debug analysis and plotting code. Every number, claim and citation was verified by the authors against the primary results.

{\small
\bibliographystyle{ieeenat_fullname}
\bibliography{references}
}

\clearpage
\makeatletter
\pdfximage{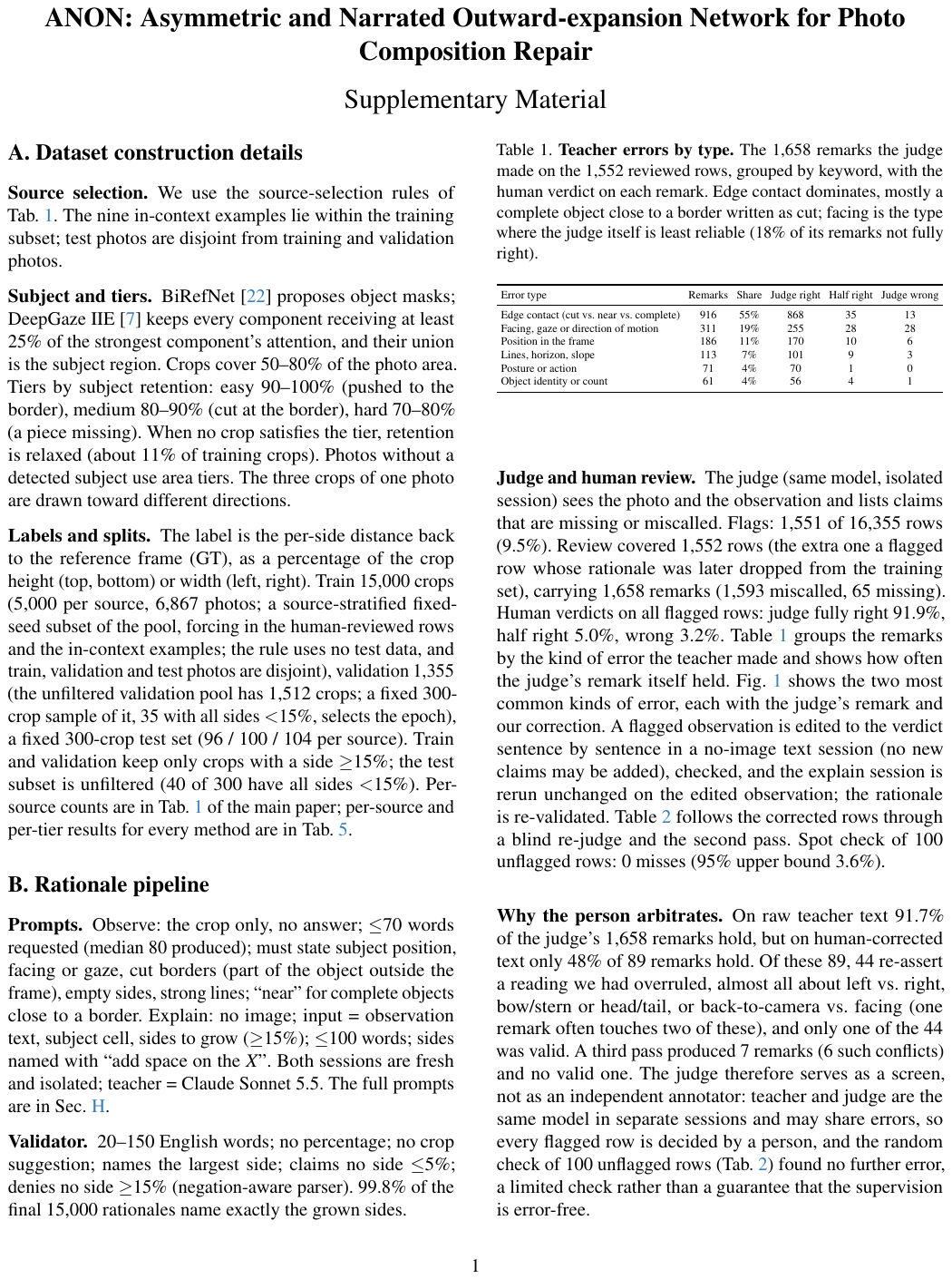}\edef\suppN{\the\pdflastximagepages}
\newcount\suppn \suppn=0
\loop\advance\suppn 1
  \edef\supp@add{\noexpand\AddToShipoutPicture*{\noexpand\AtPageLowerLeft{\noexpand\includegraphics[page=\the\suppn,width=\noexpand\paperwidth,height=\noexpand\paperheight]{ANON_supplementary.pdf}}}}\supp@add
  \null\thispagestyle{empty}\clearpage
\ifnum\suppn<\suppN\repeat
\makeatother
\end{document}